%% file: ms.tex
\documentclass[10pt,twocolumn,letterpaper]{article}

\def\cameraready{1}

\def\showruler{0}
\if\cameraready0
  \def\showruler{1}
\fi
\if\cameraready1
  \usepackage{cvpr}              % To produce the CAMERA-READY version
\else
  \usepackage[review]{cvpr}      % To produce the REVIEW version
\fi

\usepackage{graphicx}
\usepackage{amsmath}
\usepackage{amssymb}
\usepackage{booktabs}

\usepackage{listings}
\usepackage[export]{adjustbox}

\DeclareCaptionFont{white}{ \color{white} }
\DeclareCaptionFormat{listing}{
  \colorbox[cmyk]{0.43, 0.35, 0.35,0.01 }{
    \parbox{\linewidth}{\hspace{15pt}#1#2#3}
  }
}

\usepackage[pagebackref,breaklinks,colorlinks]{hyperref}

\usepackage[accsupp]{axessibility} % Improves PDF readability for those with disabilities.

\usepackage[capitalize]{cleveref}
\crefname{section}{Sec.}{Secs.}
\Crefname{section}{Section}{Sections}
\Crefname{table}{Table}{Tables}
\crefname{table}{Tab.}{Tabs.}

\def\cvprPaperID{3178} % *** Enter the CVPR Paper ID here
\def\confName{CVPR}
\def\confYear{2023}

\newcommand{\fulltitle}{Iterative Next Boundary Detection for  Instance Segmentation of Tree Rings \\ in Microscopy Images of Shrub Cross Sections}

\begin{document}

\title{\fulltitle}

\hypersetup{
    pdftitle={\fulltitle},
}

\author{
  Alexander Gillert$^{1}$ \  Giulia Resente$^{2}$ \ Alba Anadon-Rosell$^{3}$ \\ 
  Martin Wilmking$^{2}$  \ Uwe Freiherr von Lukas$^{1,4}$  \\ 
  \\
  $^{1}$Fraunhofer Institute for Computer Graphics Research (IGD), Rostock \\
  $^{2}$Institute of Botany and Landscape Ecology, Ernst Moritz Arndt University, Greifswald \\
  $^{3}$Centre for Research on Ecology and Forestry Applications (CREAF), Barcelona \\
  $^{4}$Institute for Visual \& Analytic Computing, University of Rostock \\
  {\tt \small \{alexander.gillert, uwe.freiherr.von.lukas\}@igd-r.fraunhofer.de}
}
\newcommand{\affiliationplaceholder}{
  \vspace{0.7cm}\\
  *** Affiliation Placeholder ***
  \vspace{0.70cm}\\
}

\maketitle

\newcommand{\codelink}{
  \href{http://github.com/alexander-g/INBD}{http://github.com/alexander-g/INBD}
}

%bold
\newcommand{\B}[1]{\textbf{#1}}

\input{content/00_abstract.tex}

\input{content/01_intro.tex}

\input{content/02_relatedwork.tex}

\input{content/03_method.tex}

\input{content/04_experimental_setup.tex}

\input{content/05_results.tex}

\input{content/06_conclusion.tex}
\input{content/0x_ack.tex}

%%%%%%%%% REFERENCES
{\small
\bibliographystyle{ieee_fullname}
\bibliography{content/bib.bib}
}

\end{document}

\title{
    Supplementary Materials for \\
    Iterative Next Boundary Detection for  Instance Segmentation of Tree Rings \\ in Microscopy Images of Shrub Cross Sections
}
\author{}
\newcommand{\affiliationplaceholder}{}

\maketitle

\input{content/99_supp.tex}

%%%%%%%%% REFERENCES
{\small
\bibliographystyle{ieee_fullname}
\bibliography{content/bib.bib}
}

%% file: content/00_abstract.tex
\begin{abstract}
    We address the problem of detecting tree rings in microscopy images of shrub cross sections.
    This can be regarded as a special case of the instance segmentation task with several unique challenges such as the concentric circular ring shape of the objects and high precision requirements that result in inadequate performance of existing methods.

    We propose a new iterative method which we term Iterative Next Boundary Detection (INBD).
    It intuitively models the natural growth direction, starting from the center of the shrub cross section and detecting the next ring boundary in each iteration step.
    In our experiments, INBD shows superior performance to generic instance segmentation methods and is the only one with a built-in notion of chronological order.

    \if\cameraready1
        Our dataset and source code are available at \codelink.
    \else
        Our dataset and source code will be published after the review process.
    \fi
 \end{abstract}

%% file: content/01_intro.tex
\section{Introduction}

\begin{figure}[h]
    \centering
    %\fbox{\rule{0pt}{4cm} \rule{.9\linewidth}{0pt}}
    \begin{subfigure}{0.49\linewidth}
      \includegraphics[width=\textwidth]{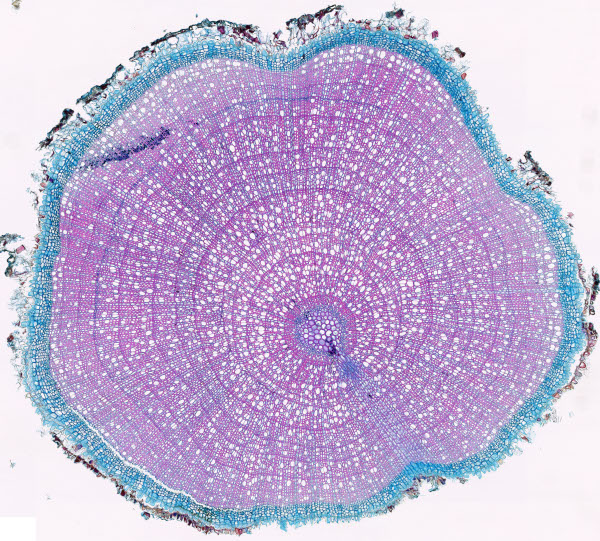}
        %\caption{}
        \label{fig:overview-a}
    \end{subfigure}
    \begin{subfigure}{0.49\linewidth}
      \includegraphics[width=\textwidth]{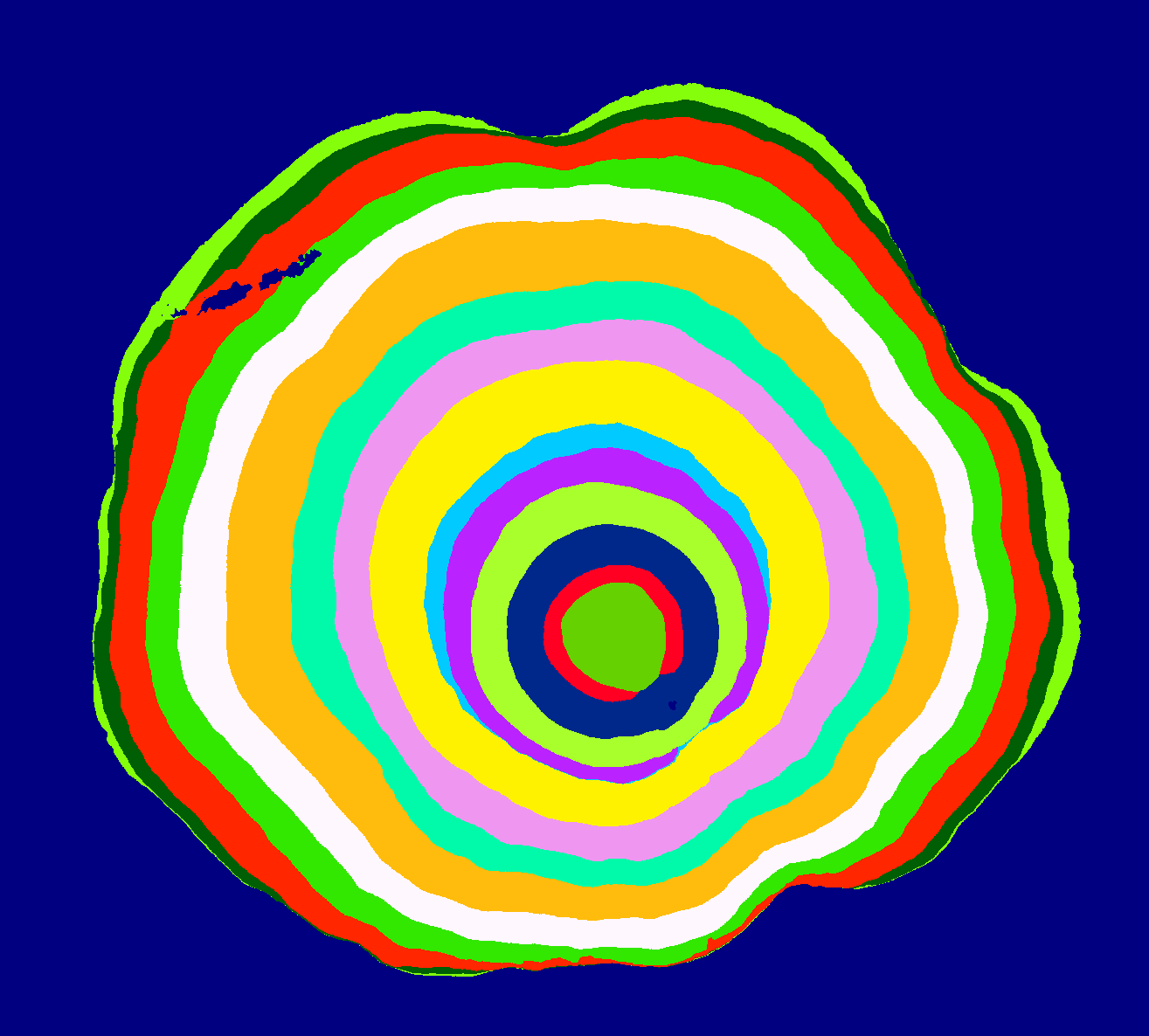}
        %\caption{}
        \label{fig:overview-a}
    \end{subfigure}
    \hfill
    \begin{subfigure}{0.49\linewidth}
      \includegraphics[width=\textwidth]{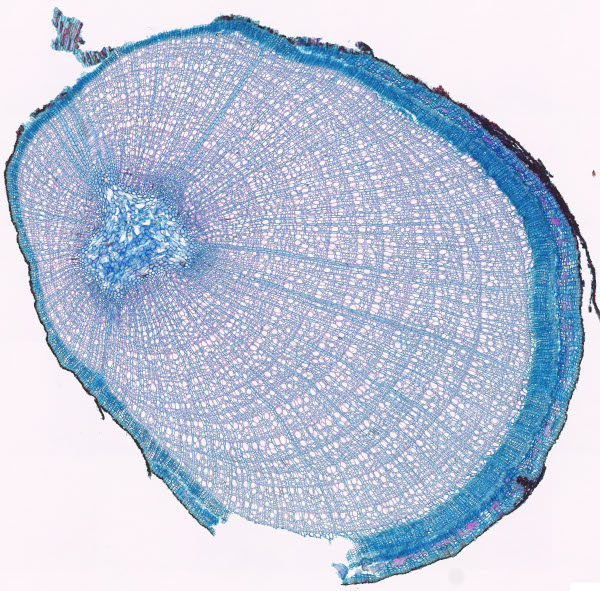}
        %\caption{}
        \label{fig:overview-a}
    \end{subfigure}
    \begin{subfigure}{0.49\linewidth}
      \includegraphics[width=\textwidth]{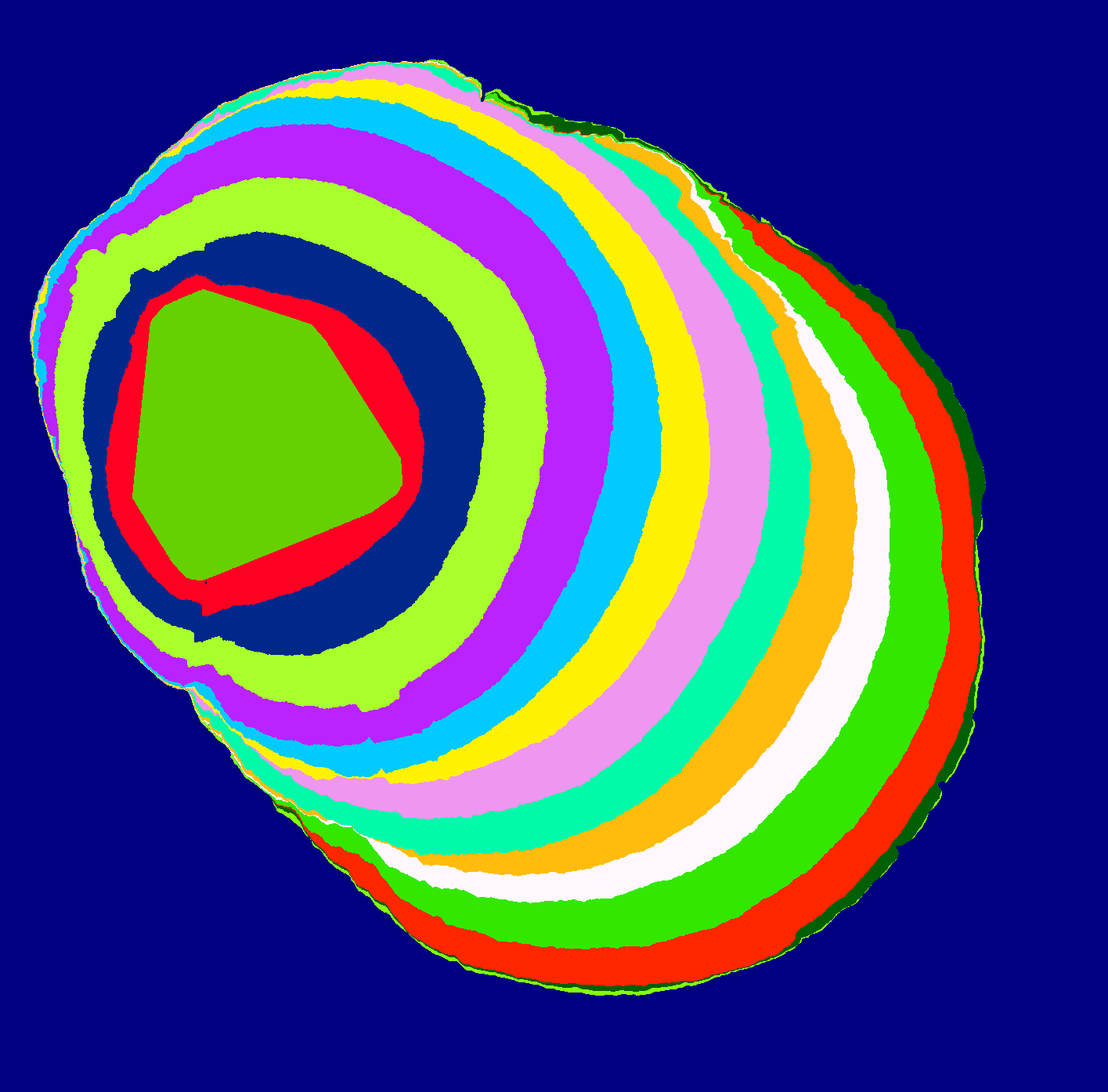}
        %\caption{}
        \label{fig:overview-a}
    \end{subfigure}
    \caption{
      \label{fig:frontpage}
      Example microscopy images (left) of shrub cross sections from our new dataset and the outputs (right) of our proposed method INBD for instance segmentation of tree rings
      \vspace{-0.2cm}
      }
\end{figure}

\begin{figure*}
    \centering
    \begin{subfigure}{0.24\linewidth}
      %\fbox{\rule{0pt}{4cm} \rule{.9\linewidth}{0pt}}
      \includegraphics[width=\textwidth]{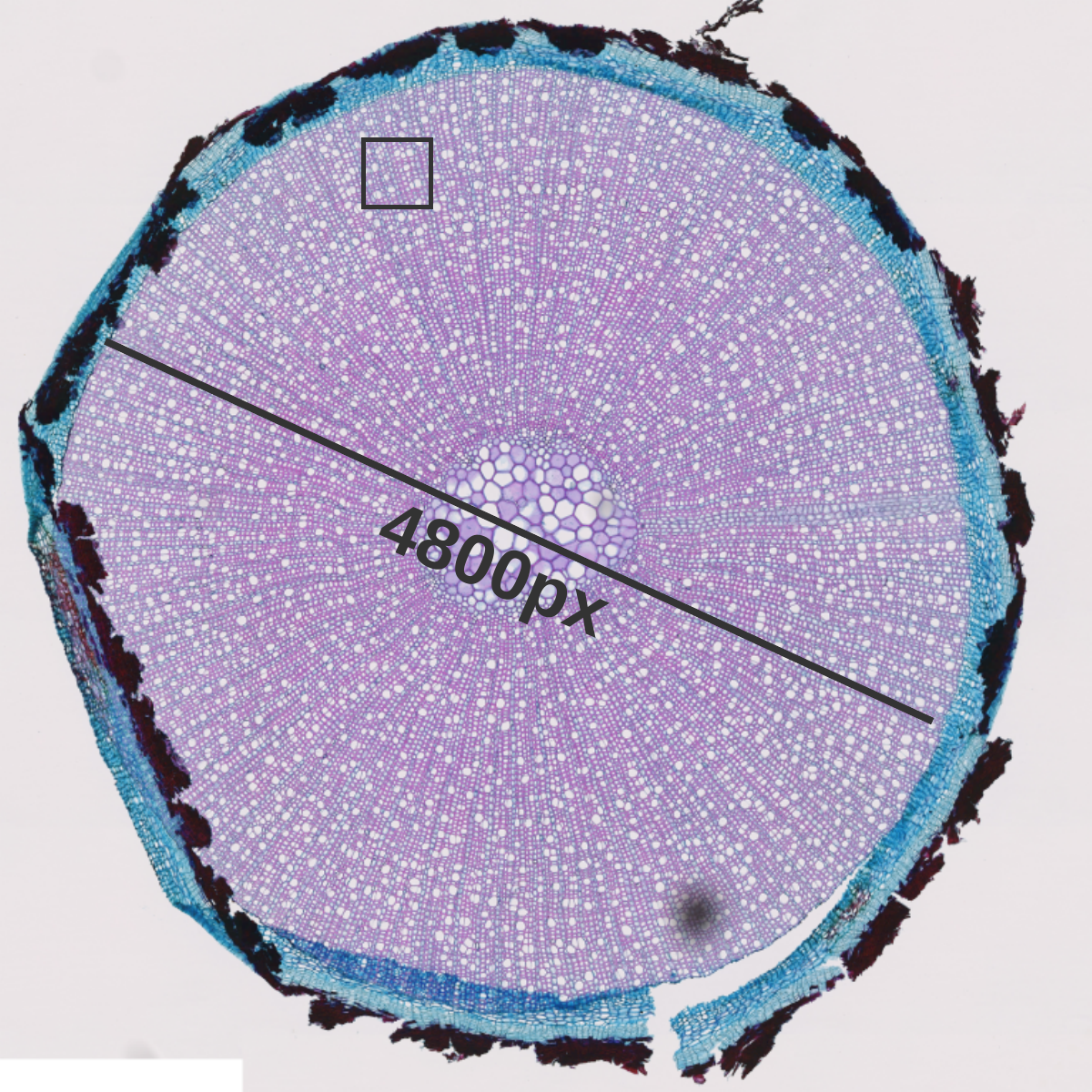}
      %\caption{High resolution and faint boundaries}  %FIXME:line break
      \caption{ \fontsize{7.55pt}{1pt}\selectfont High resolution and faint boundaries }
      \label{fig:challenges-a}
    \end{subfigure}
    \hfill
    \begin{subfigure}{0.24\linewidth}
        %\fbox{\rule{0pt}{4cm} \rule{.9\linewidth}{0pt}}
        \includegraphics[width=\textwidth]{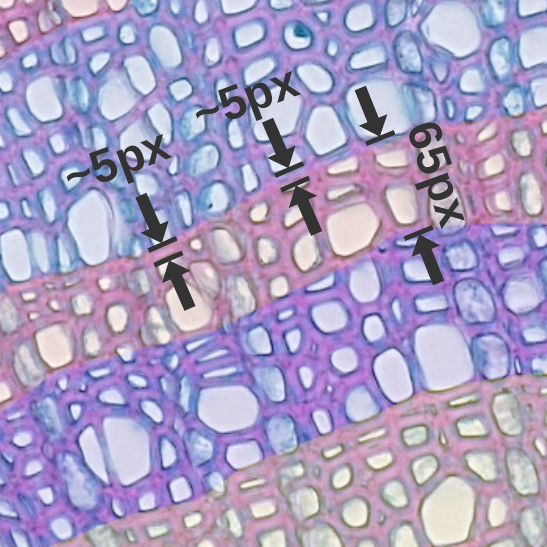}
        %\caption{High precision requirements \\ \phantom{placeholder}}
        \caption{ \fontsize{7.55pt}{1pt}\selectfont High precision requirements }
        \label{fig:challenges-b}
      \end{subfigure}
    \hfill
    \begin{subfigure}{0.24\linewidth}
        %\fbox{\rule{0pt}{4cm} \rule{.9\linewidth}{0pt}}
        \includegraphics[width=\textwidth]{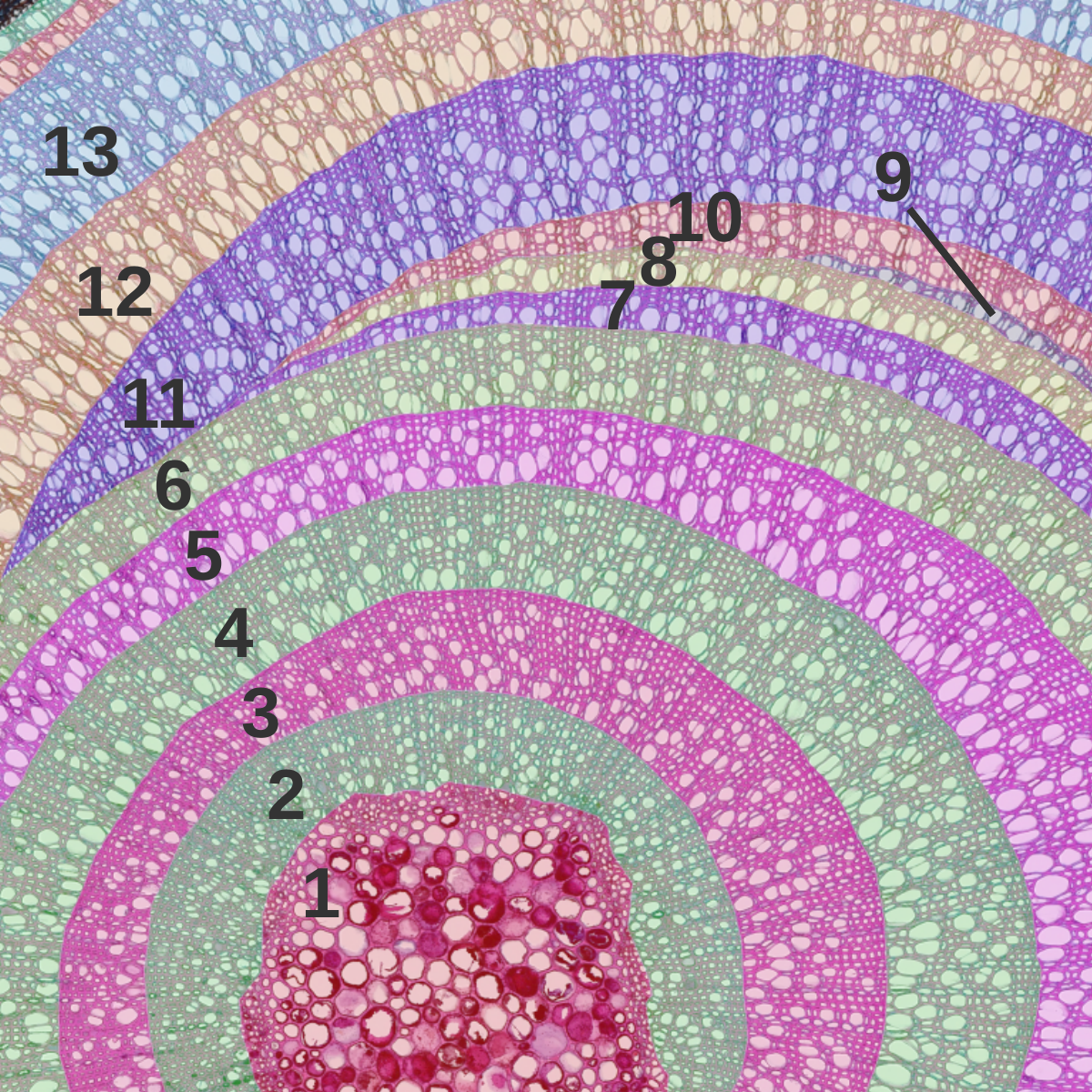}
        %\caption{Wedging rings \\ \phantom{placeholder} }
        \caption{ \fontsize{7.55pt}{1pt}\selectfont Wedging rings }
        \label{fig:challenges-c}
      \end{subfigure}
      \hfill
    \begin{subfigure}{0.24\linewidth}
      %\fbox{\rule{0pt}{4cm} \rule{.9\linewidth}{0pt}}
      \includegraphics[width=\textwidth]{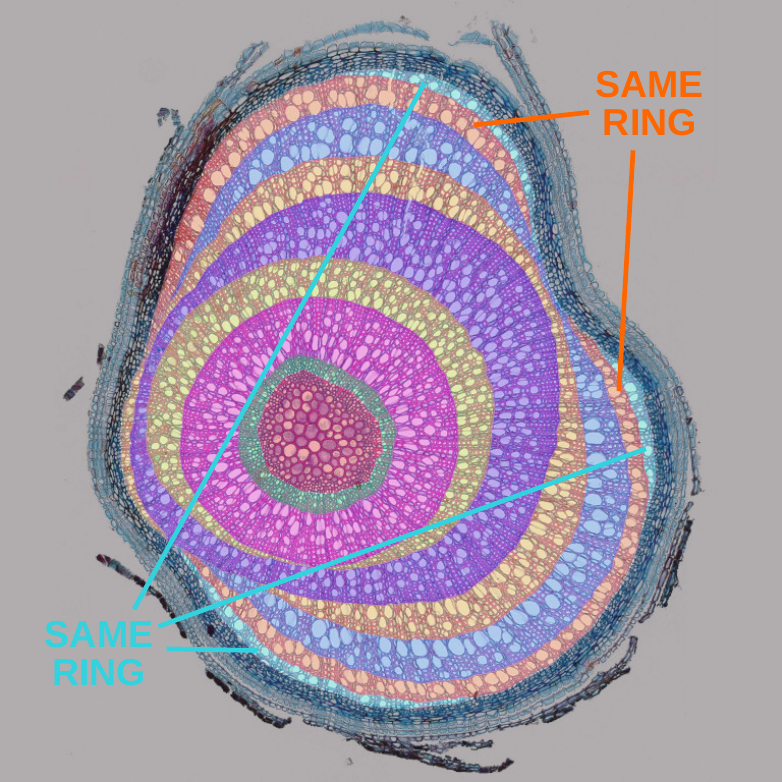}
      %\caption{Disconnected rings \\ \phantom{placeholder}}
      \caption{ \fontsize{7.55pt}{1pt}\selectfont Disconnected rings }
      \label{fig:challenges-d}
    \end{subfigure}
    \caption{
      Some of the challenges encountered in this task:\\ 
      (a) Boundaries inbetween tree rings are often hard to recognize. For example, this cross section contains 14 rings.\\
      (b) Crop of the previous image (indicated by the square) with overlayed annotation. A tree ring is only 65 pixels wide or ca 1.4\% of the full cross section diameter.
          The cell wall that divides late summer cells and the next year's early summer cells is only 5 pixels wide or 0.1\%. \\
      (c) Wedging rings can complicate finding the chronologically correct next year ring. \\
      (d) Rings can grow in multiple disconnected parts from different sides. \\
    }
    \label{fig:challenges}
    %\vspace{-0.2cm}
\end{figure*}

Dendrochronology is the science that provides methodologies to date tree rings \cite{Cook1990MethodsOD}, i.e. measuring and assigning calendar years to the growth rings present in a wood stem.
By analyzing anatomical properties like ring widths or the cell sizes within the rings, dendrochronology can be applied to dating archaeological manufactures, tracking timber sources or reconstructing past climate conditions \cite{lageard2016dendrochronology}.

For climate reconstruction in the Arctic, shrubs constitute the most important source of dendrochronological information, since they are the only woody plants able to thrive there \cite{Weijers2010DendrochronologyIT}. 
As temperature is a limiting factor for shrub growth in the Arctic, it shows a strong relationship with climate, making these plants a reliable proxy to reconstruct past climate events \cite{Wilmking2018InfluenceOL}.
Dendrochronological analyses on shrubs are usually performed on thin cross sections of branches or roots and observed under the microscope with a magnification that allows ring identification at a cellular level.
As of now, ecological studies are limited in size by the amount of manual analysis work due to the lack of automatic tree ring detection methods.

With this paper we want to introduce this problem to the computer vision community and enhance the capabilities for ecological sciences.
We release a new dataset containing high resolution microscopy images of shrub cross sections and propose a specialized method for growth ring identification.
Example images from our dataset and corresponding outputs of our method are shown in Figure \ref{fig:frontpage}.
From a computer vision point of view, this can be regarded as a special case of the instance segmentation task,
however it differs from previous generic datasets in several ways which makes existing methods underperform.

Figure \ref{fig:challenges} illustrates these differences.
For one, the concentric ring shape of the instances can pose a significant obstacle, particularly for top-down methods because the objects have almost identical bounding boxes.
This gets complicated by the fact that year rings can also form incomplete circles (wedging rings) and grow from only one side, or even in multiple disconnected parts from different sides (\ref{fig:challenges-d}).
Depending on the species, plant part and climatic conditions the amount of wedging rings can range from zero to being the majority.
Assigning the correct order to wedging rings can be an issue where rings of more than 2 years touch each other (\ref{fig:challenges-c}).
Bottom-up methods on the other hand struggle with faint ring boundaries (\ref{fig:challenges-a}) as the presence of the boundary pattern is not always constant throughout the whole stem circumference.
They are prone to merging rings where no boundary can be detected or splitting them where the ring width is narrow.
Next, the images are acquired at a high resolution (\ref{fig:challenges-a}) to capture cellular information, yet a high degree of precision is required for the downstream task of assigning individual cells to the correct year.
The thickness of a cell wall that is dividing the cells from one ring to another can be as low as 0.01\% of the whole object (\ref{fig:challenges-b}).
Finally, as the preparation of samples and annotation of the images is very costly, training has to be performed in a low data regime.

We argue that a specialized approach can help to overcome those challenges and propose a new iterative method which we term Iterative Next Boundary Detection (INBD).
In the first step, it performs semantic segmentation to detect basic features such as the background, center and the ring boundary pixels.
From this starting point, it iteratively detects the next year ring's boundaries, following the natural growth of the plant.
This process is augmented with a recurrent wedging ring detection module to counteract issues with incomplete rings.
We compare our method with both top-down and bottom-up generic instance segmentation in our experiments in which it shows better results.
Moreover, it is the first method that automatically assigns a chronological order to the detected objects.

The contributions of this paper can summarized as follows:
\begin{itemize}
  \item Publication of a new challenging dataset for a special case of instance segmentation.
  \item Development of the specialized method INBD for tree ring instance segmentation.
  \item Evaluation of previous generic instance segmentation methods and comparison with INBD
\end{itemize}

%% file: content/02_relatedwork.tex
\begin{figure*}
    \centering
      \framebox{\includegraphics[width=\textwidth]{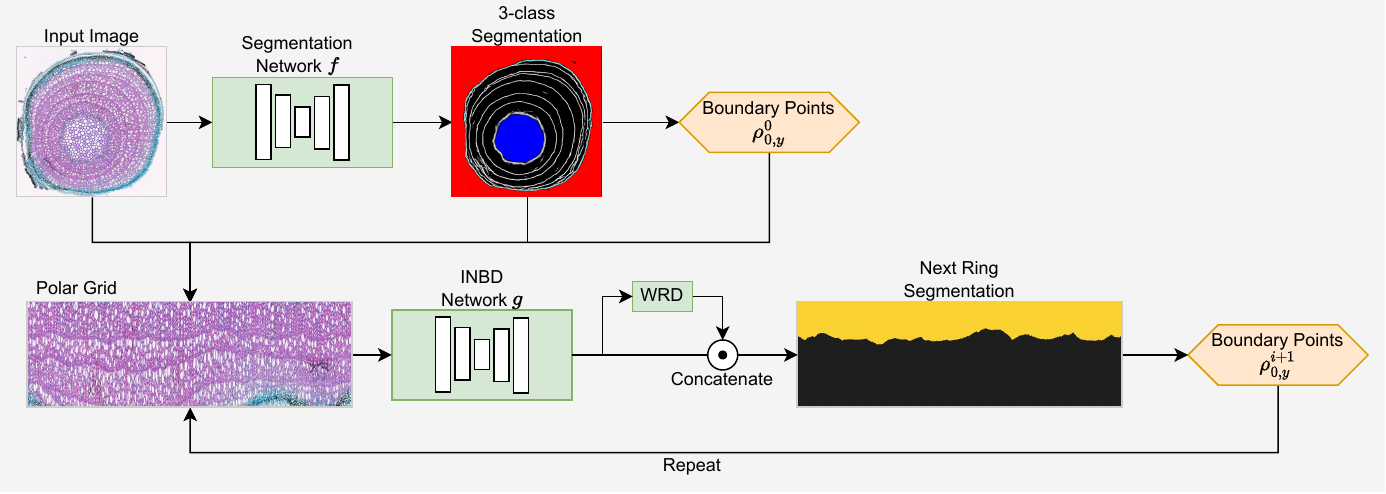}}
    \caption{
      Overview of the INBD pipeline.
      An input image is first passed through a generic semantic segmentation network that detects 3 classes: background, ring boundaries and the center ring.
      A polar grid is sampled starting from the the detected center ring and passed to the main INBD network that detects the next ring.
      This process is repeated until the background is encountered.
    }
    \label{fig:architecture}
  \end{figure*}

\section{Related Work}

Instance segmentation is a widely studied problem in computer vision, commonly benchmarked on a variety of standard generic datasets such as COCO \cite{coco} which contains photographs of everyday objects or the the more specialized CREMI 2016 \cite{cremi2016} challenge for cell segmentation in electron microscopy images.
No publicly available dataset is known to us that contains concentric ring shaped and ordered objects.

Methods can be categorized into \emph{top-down} and \emph{bottom-up} procedures.
Mask-R-CNN \cite{maskrcnn} is the most widely used architecture and belongs to the top-down group.
It relies on an object detector to first detect bounding boxes of objects which are then segmented.
This fails on overlapping or as in our case concentric objects due to non-maximum suppression.
Moreover, it can only generate low resolution masks.
Contour methods such as Deep Snake \cite{deepsnake} or DANCE \cite{dance} can generate masks with higher precision but still require an upstream object detector.

Bottom-up methods for instance segmentation methods work by first computing object boundaries or affinities and then clustering the resulting superpixel graph into whole objects via the multicut objective.
Finding the optimal solution for this is known to be NP-hard \cite{kappes2011globally}, therefore several approximate solvers such as GASP\cite{GASP} have been developed.
These methods perform significantly better on our dataset but still show deficits in cases where object boundaries are hard to recognize and they cannot handle disconnected rings (such as in Fig. \ref{fig:challenges-d}).

None of the above methods has a built-in notion of sequence order of the detected objects that would be needed to assign a tree ring to a year.

Application of deep learning methods to ecological purposes is nowadays an established procedure \cite{christin2019applications} due to the complexity related to ecological investigations and the use of increasingly larger datasets.
Specifically for quantitative wood anatomy (QWA), deep learning research has so far focused mostly on detection and measurement of cells such as in \cite{masktrainrepeat, GarciaPedrero}.
Tree ring detection was subject in \cite{alexis_treerings, deepdendro}, however only on scans or photographs of mature wood core samples rather than full cross sections as in our case.

ROXAS\cite{roxas2014, roxas2005} is the most commonly used analysis tool in QWA, however it is based on traditional image processing methods and not on deep learning which makes it sensitive to sample processing and image quality.
It also contains tree ring detection functionality which works by line-following early summer cells but requires domain knowledge for manual tuning of many species-specific parameters like cell shape and size.

%% file: content/03_method.tex
\begin{figure*}
  \centering
  \begin{subfigure}{0.49\linewidth}
    \includegraphics[width=\textwidth]{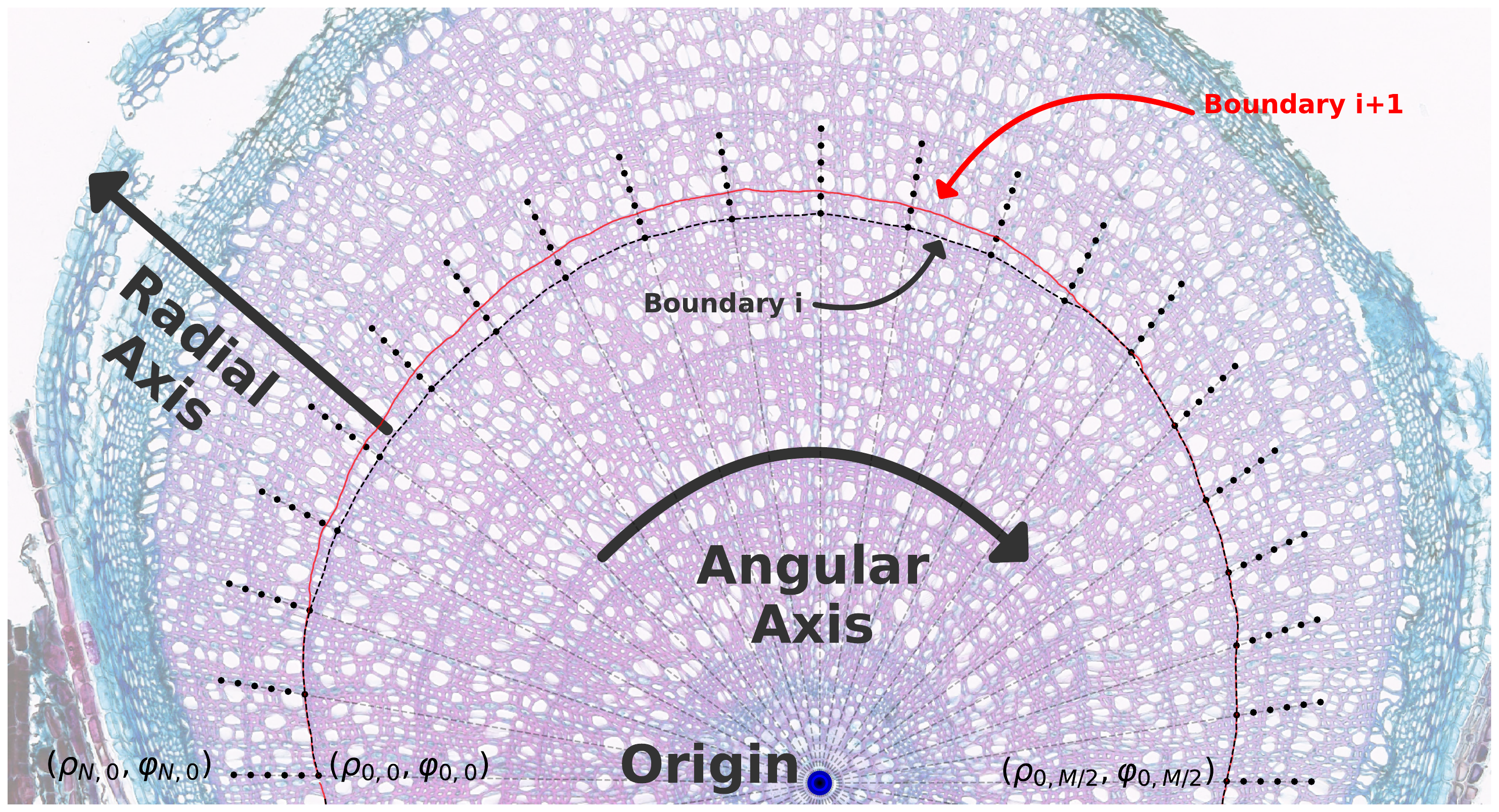}
    \caption{ Polar Grid Sampling }
    \label{fig:polargrid}
  \end{subfigure}
  \hfill
  \begin{subfigure}{0.49\linewidth}
      \includegraphics[width=\textwidth]{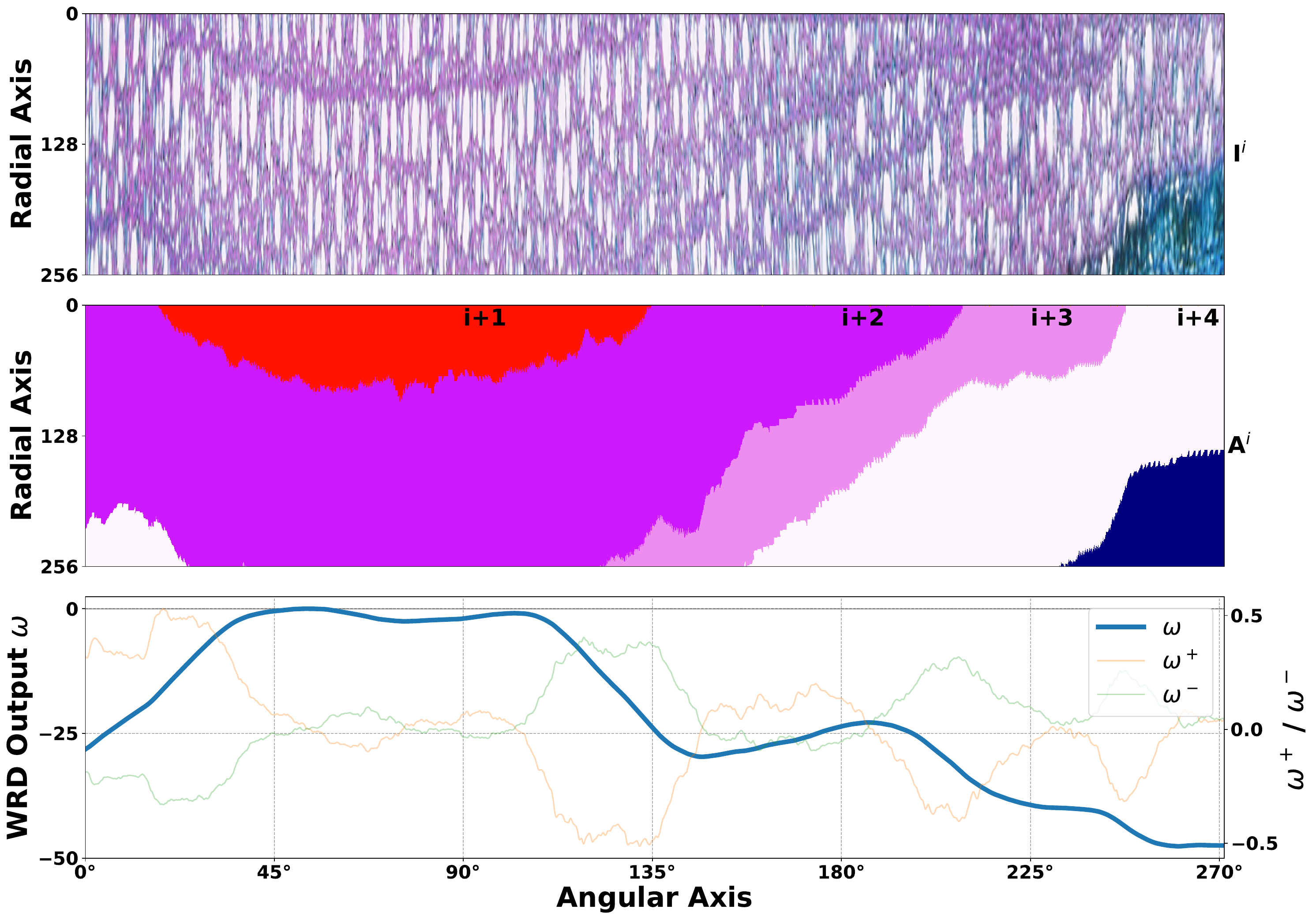}
      \caption{ Wedging Ring Detection}
      \label{fig:WRD}
    \end{subfigure}
  \caption{
    \label{fig:expl}
    Visual explanation of the concepts.
    (a) We sample on a polar grid starting from the previous detected ring boundary. The number of points is  reduced for better visualization.
    (b) shows the resulting input to the network $g$ (top), the corresponding annotation (center) and the output signal $\omega$ of the wedging ring detection module (bottom).
    $\omega$ accumulates along the angular axis, rising on start and falling on end of wedging rings.
  }
  \vspace{-0.3cm}
\end{figure*}

\section{Method}

\label{section:method}

On a high level, INBD simply modifies and extends the various contour based methods like Deep Snake \cite{deepsnake} or PolarMask \cite{polarmask} with an iterative inference procedure.
In reality, this requires several important design choices to make this perform well.
The influence of the individual design choices is analyzed in an ablation study in subsection \ref{sec:ablation}.
An architectural overview of the INBD pipeline can be found in Figure \ref{fig:architecture}.

\input{content/03a_architecture.tex}

\input{content/03b_polargrid.tex}

\input{content/03c_WRD.tex}

\input{content/04a_dataset_table.tex}

\input{content/03d_training.tex}

%% file: content/03a_architecture.tex
\subsection{Network Architecture}

The INBD pipeline is composed of two neural networks.
The first network is a simple semantic segmentation network that is trained to detect three classes: background, ring boundaries and the center ring (or pith).
We denote this network and its output with $ f(I) = (y^{bg}, y^{bd}, y^{ct}) $ when applied on image $I$. %
We select an architecture based on U-Net \cite{unet} with a pretrained backbone.
The three classes are trained with a combination of cross-entropy loss and the Dice loss \cite{vnet}:
\begin{equation}
  L_{f} = \lambda_1 L_{CE}^{background} + \lambda_2 L_{Dice}^{boundaries} + \lambda_3 L_{CE}^{center}
\end{equation}
with $  \mathbf{\lambda} = (0.01,\ 1.0,\ 0.1)$ balancing coefficients to account for class imbalances.
Due to the large size of the images and for a larger field of view, $f$ operates on $\times 0.25$ of the original resolution.

Our main network, which we denote with $g$, is another 2D convolutional segmentation network that classifies each pixel as belonging to the next ring or not.
By choosing a 2D network as opposed to a 1D one, as in many contour methods such as Deep Snake \cite{deepsnake}, we can leverage transfer learning since we are working in a low data regime and in addition to that we can reject and interpolate ambiguous predictions (see below, Eq. \ref{eq:ambiguous2}).
This second network has mostly the same architecture as the first one, except that we replace the normal 2D convolutions with circular convolutions to wrap around the full circle, as also used in Deep Snake \cite{deepsnake}.
The circularity is only applied to  the angular axis (see below).
 %TODO: More details on the used architecture in the supplement.

%% file: content/03b_polargrid.tex
\subsection{Polar Grid}

The network $g$ operates on ``unrolled'' rings $I^{i} \in \mathbb{R}^{[C \times N \times M]}$ sampled on a polar grid $P^{i} \in \mathbb{R}^{[N \times M]}$ with $N = 256$ a fixed resolution in the radial dimension and $M$ an adaptive resolution in the angular dimension.
The polar grid origin is computed from the center of mass of the center ring $y^{ct}$ as detected by $f$.
Polar coordinates impose a prior, ensuring a coherent (quasi-)convex shape, contrary to Cartesian coordinates.

We express the sampling points for ring $i$ as polar coordinates $( \rho^i_{xy} \ , \  \varphi^i_{xy} )$, with $x \in \left[ 0, N \right]$, $y \in \left[ 0, M \right]$ indices within the grid.
The boundary point radii for the second ring $\rho^1_{0,y}$ are inferred directly from the detected center ring $y^{ct}$.

Estimating the extent of the grid in the radial dimension (i.e. $\rho_{N,y}$) is crucial: if too short, the next ring will be cut off, if too long, the next ring might get skipped and not detected at all.
For this, we compute the distances to the closest positive value in $y^{bd}$ for each angle $\varphi$ and set the extent to $1.5 \times$ 95\%-th percentile of these distances, to make sure that most points are included but also to filter outliers.
This was empirically verified to cover all rings in our dataset.
The remaining radial values $\rho$ are then uniformly distributed along this range:
$\rho_{x,y} = \frac{1}{N} (\rho_{N,y} - \rho_{0,y})x + \rho_{0,y}$

The angular resolution $M$ is computed so that the angles $\varphi$ have an approximately uniform euclidean distance to each other across rings: since the outer rings have a larger circumference than the inner ones they should be sampled at a higher angular resolution $M$.
The value $M^i$ for ring $i$ is computed from the previous ring's average radii:

\begin{equation}
  \label{eq:angular_resolution}
        M^{i} = \alpha \ \frac{1}{M^{i-1}} \sum^{M^{i-1}}_y \rho^{i-1}_{0,y}  %
\end{equation}
with $\alpha$ a hyperparameter that controls the general density of $M^{i}$ which we set to $2\pi$ where not otherwise mentioned.
The angles $\varphi$ are spaced uniformly along the full circle: $\varphi_{x,y} = \frac{1}{M} 2\pi y$

The channel dimension $C=7$ is composed of the RGB channels of the input image,
the detected ``background'' and the ``boundaries'' outputs $y^{bg}$ and $y^{bd}$ from $f$, normalized radii $\rho$ and the output of the wedging ring detection module (see below) concatenated together.

The main loss $L^{cls}_{CE}$ for network $g$ is the standard cross-entropy loss to classify each pixel in the polar grid as belonging to the next ring or not, according to the annotation $A^{i}$, sampled on the same polar grid.

\subsection{Inference}

To perform inference of the next ring's boundary points $\rho^{i+1}_{0,y}$, we select last positive point in the output $g(I^{i})$ column-wise, where it is unambiguous:
\begin{align}
  \label{eq:ambiguous}
  \tilde{X}_{y}             &= \{ x, \  \text{where} \ g(I^{i})_{x,y} = 1 \} \\
  \label{eq:ambiguous2}
  \rho^{i+1}_{0,y}  &= \begin{cases}
      \rho^{i}_{\max \tilde{X}_y,y}\ ,        & \text{if} \max \tilde{X}_y = \min \overline{\tilde{X}_y} - 1 \\
      \text{undefined}\ ,         & \text{otherwise} \\
  \end{cases}
\end{align}
Ambiguous values linearly interpolated.
Importantly, the interpolation should be performed on polar coordinates and not on Cartesian ones and wrap around the circle.
This detection process is repeated iteratively with the new predicted ring boundary points $i+1$ as the starting point to detect the ring $i+2$ until the background $y^{bg}$ that was detected by the segmentation network $f$ is reached.

%% file: content/03c_WRD.tex
\subsection{Wedging Ring Detection} 
\label{subsection:wrd}

The method as described so far is able to detect full tree rings sufficiently well but struggles with wedging rings.
More specifically, it is prone to skipping a ring boundary in locations where the wedging ring is far away and outside the field of view, e.g. as in Figure \ref{fig:challenges-c} when trying to detect the next boundary after ring 6.
To counteract this issue we insert a wedging ring detection (WRD) module before the final classification layer.

This module consists of 3 additional convolutional layers with two output channels. %TODO: more details in supplement
The two channels are averaged along the radial axis into 1-dimensional signals $\omega^+$ and $\omega^-$ $\in \mathbb{R}^M$ and combined via a recurrent mechanism:
\begin{align}
  \omega_{0}'       &= \beta \\
  \omega_{\varphi}' &= \sigma( \omega_{\varphi - 1}^+ ) - \sigma( \omega_{\varphi - 1}^- )  \\
  \omega_{\varphi}  &=  \omega_{\varphi}' - \max \omega' \label{eq:wedgienorm}
\end{align}
where $\sigma$ is the sigmoid function and $\beta$ is a starting point constant.
Intuitively, $\omega^+$ is responsible for detecting the start of a wedging ring and increases the output signal $\omega$, whereas $\omega^-$ detects the end and decreases it. 
$\omega$ is then forwarded to the final classification layer by concatenating it to the features along the channel dimension.

During inference, the choice of $\beta$ does not matter because of the normalization by subtracting the maximum (Eq. \ref{eq:wedgienorm}).
This ensures a standardized representation of the signal to the following downstream classification layer, irrespective of the starting point $\beta$. 
High values close to zero indicate valid locations (next ring or $i+1$), whereas low values are invalid locations (next but one or $i+2$).
This functionality is illustrated graphically in Figure \ref{fig:WRD}.

Although in theory the network could derive useful information from this module by itself, we have found that in practice it is highly beneficial to add an explicit training signal.
Again, we use the cross entropy loss, but modified for the single dimension and applied on the unnormalized signal $\omega'$:
\begin{align}
    L^{wrd} &= A^{wrd}_{\varphi} \ log \ \sigma(\omega_{\varphi}') + (1-A^{wrd}_{\varphi}) \ log \ 1-\sigma(\omega_{\varphi}') \\
    A^{wrd}_{\varphi} &= \begin{cases}
        1,         & \text{where} \ A^{i}_{0, \varphi} = i+1 \\
        0,         & \text{otherwise} \\
    \end{cases}
\end{align}

During training $\beta$ is set so that $\sigma(\omega') = 0$  if the ground truth at angle $\varphi = 0$ is low, or so that $\sigma(\omega') = 1$ if it is high,
to avoid incorrect training signals (we choose $\beta = \pm 15$). %

\begin{figure}[h]
  \centering
  \begin{subfigure}{0.49\linewidth}
    \includegraphics[width=\textwidth]{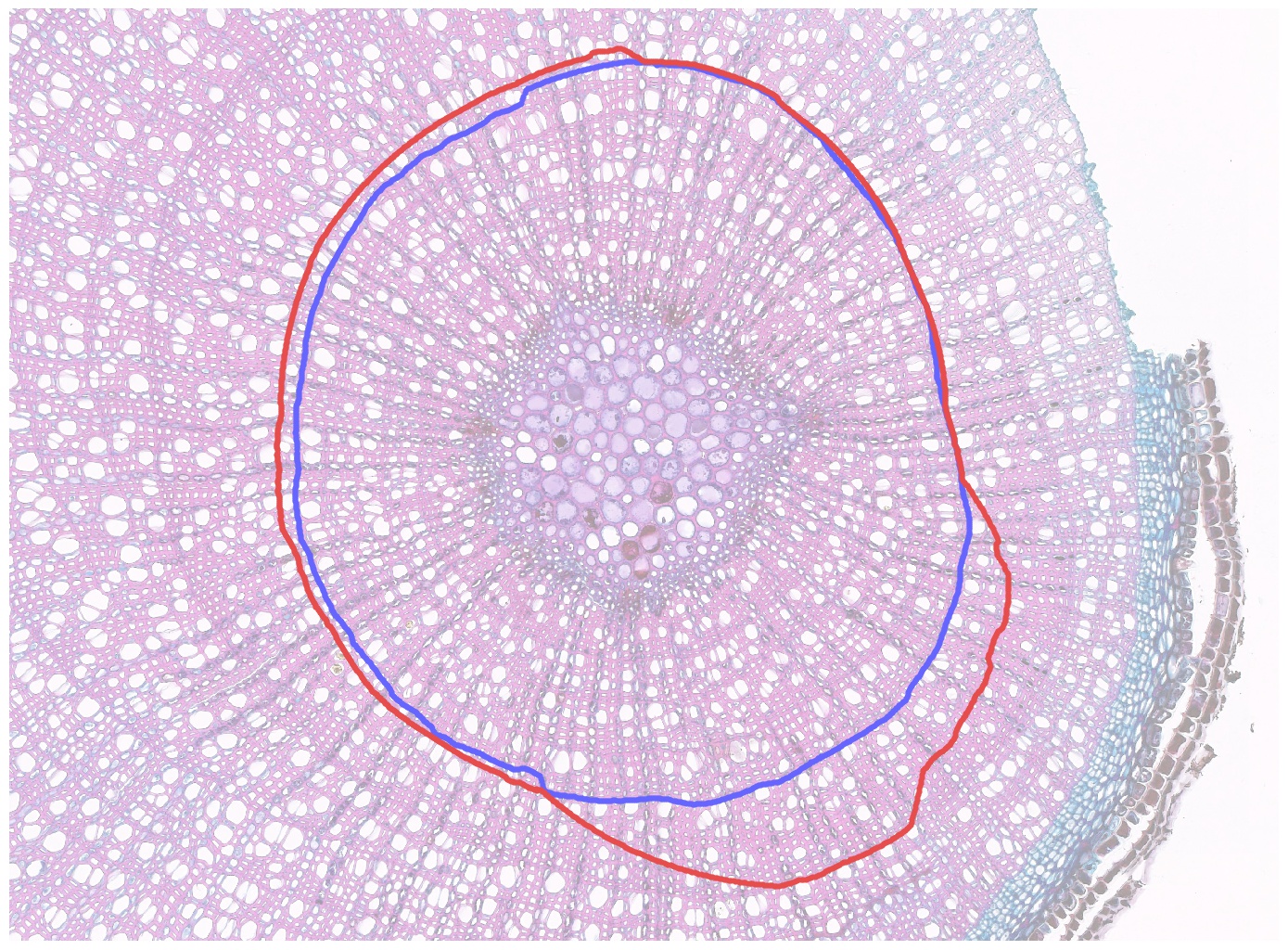}
      %\caption{}
      \label{fig:overview-a}
  \end{subfigure}
  \begin{subfigure}{0.49\linewidth}
    \includegraphics[width=\textwidth]{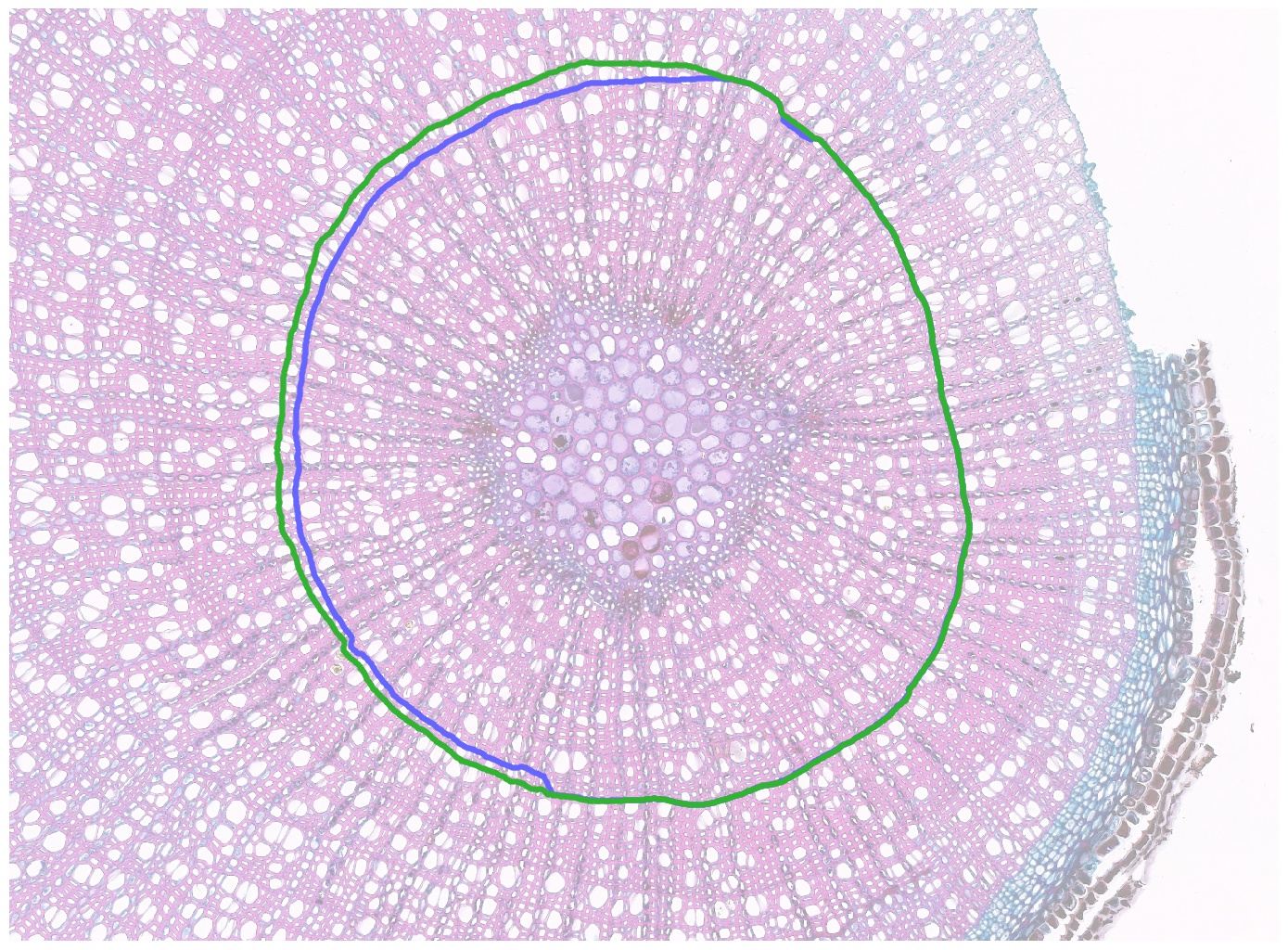}
      %\caption{}
      \label{fig:overview-a}
  \end{subfigure}
  \vspace{-0.2cm}
  \caption{
    \label{fig:wrd_effect}
    INBD can be prone to skipping boundaries.
    In this example, the wedging ring detection module helps to catch mistakes like this. (Left: without WRD, right: with WRD)
    }
  \vspace{0.2cm}
\end{figure}

An example where this module helps to catch an error is shown in Figure \ref{fig:wrd_effect}.

The final loss for network $g$ is defined as $L_g = L_{CE}^{cls} + \lambda_4 L^{wrd}$ with $\lambda_4 = 0.01$. We have found higher values to have a negative impact on the main classification loss.

%% file: content/04a_dataset_table.tex
\begin{table*}[]
    \centering
    \begin{tabular}{llccccll}
     Subset         & 
     Species                 & 
     \begin{tabular}[c]{@{}c@{}}Training \\ Images\end{tabular}  & 
     \begin{tabular}[c]{@{}c@{}}Test     \\ Images\end{tabular}  & 
     \begin{tabular}[c]{@{}c@{}}Number   \\ of rings\end{tabular}  &
     \begin{tabular}[c]{@{}c@{}}Average  \\ diameter\end{tabular}  & %
     Example Images
     \\ %
     %%%%%%%%%%%%%%%%%%%%%%
     \hline
     %%%%%%%%%%%%%%%%%%%%%%
     %
     DO     & Dryas octopetala        &      22 &   42  &   544                  & 3700px                    &  Figs. \ref{fig:frontpage} (bottom), \ref{fig:failurecases} (bottom)  \\
     EH     & Empetrum hermaphroditum &      24 &   58  &   949                  & 3260px                    &  Figs. \ref{fig:frontpage} (top), \ref{fig:challenges-c}, \ref{fig:challenges-d}\\
     VM     & Vaccinium myrtillus     &      22 &   45  &   494                  & 3979px                    &  Figs. \ref{fig:challenges-a}, \ref{fig:challenges-b}, \ref{fig:failurecases} (top) \\
     &  &  &  & 
    \end{tabular}
    
    \caption{
        \label{tab:dataset_stats}
        Overview of our dataset
    }
\end{table*}

%% file: content/03d_training.tex
\subsection{Training Procedure}
\label{subsection:training}

Since INBD is an iterative procedure, errors caused by an earlier ring get easily propagated onto the later rings.
It can however also recover from previous mistakes if trained with an \emph{iterative training} procedure:
%the training loop should incorporate several prediction iterations instead of only taking the boundary points from the annotation as the starting point for the polar grids.
rather than using only the (near-perfect) boundary points from the annotation, the training loop should incorporate previous (possibly faulty) predictions  as the starting point for polar grids.
Listing \ref{lst:training} shows the high-level pseudo code for one training epoch.

\lstdefinestyle{customc}{
    belowcaptionskip=1\baselineskip,
    breaklines=true,
    frame=L,
    showstringspaces=true,
    basicstyle=\footnotesize\ttfamily,
    identifierstyle=\color{black},
    stringstyle=\color{orange},
    backgroundcolor=\color[RGB]{240,240,240},
}

%XXX: OMG!!
\begin{lstlisting}[frame=single, style=customc, caption={Pseudo-code for one training epoch \label{lst:training} \vspace{-0.3cm}}, escapeinside={(*}{*)}] 
  for (image (*$I$*), annotation (*$A$*), ring (*$i$*)) in dataset:
    (*$L_g$*)  = 0
    (*$\rho^{i}_{0,y}$*) = boundary_from_annotation((*$A$*), (*$i$*))
    loop (*$i$*) = (*$i$*)..(*$i$*)+(*$n$*):
      (*$\hat{ \rho}^{i}_{0,y}$*) = augment((*$\rho^{i}_{0,y}$*))
      (*$I^{i}$*)  = sample_polar_grid((*$I$*), (*$(\hat{ \rho}^{i}, \varphi^{i})$*))
      (*$y^{i}$*)  = (*$g$*)((*$I^{i}$*))
      (*$L_g$*) += compute_loss((*$y^{i}$*), (*$A^{i}$*))
      (*$\rho^{i+1}_{0,y}$*) = compute_boundary((*$y^{i}$*))
    backpropagate( (*$L_g$*) / (*$n$*) )
\end{lstlisting}
Where not otherwise mentioned we use $n=3$ iterations per epoch.

Besides the conventional data augmentations such as the pixel-wise color jitter operations we employ additional augmentations specific to polar grids such as varying the boundary points:

\begin{equation}
  \hat{ \rho}_{0,y}  = \rho_{0,y} + \cos ( \varphi_{0,y} + X_0 ) \gamma_0 + X_1 \gamma_1
\end{equation}
with $X \sim \mathcal{U}(-1,1)$ random variables  and $\gamma$ hyperparameters.

Both networks are trained separately with the AdamW \cite{AdamW} optimizer for 100 epochs, 1e-3 base learning rate and cosine annealing \cite{cosine_lr} learning rate schedule.

%% file: content/04_experimental_setup.tex
\section{Experimental Setup}

\subsection{Dataset}

Our dataset consists of overall 213 high-resolution images.
It is split into 3 subsets according to the plant species.
An overview is provided in Table \ref{tab:dataset_stats}.
The amount of images is rather low due to the high cost of sample preparation as well as annotation: a single image containing a large amount of rings can take up to 6 hours to annotate by hand.
\if\cameraready1
    The dataset and annotations are publicly available at \codelink. %{\color{red} {\bf [TODO: LINK] } }.
\else
    The dataset and annotations will be made publicly available after the review process.
\fi

The shrub samples were collected at subalpine, alpine and subarctic sites 
\if\cameraready1
across the Pyrenees, Southern Norway and Northern Sweden. 
\else
at locations revealed after the review process.
\fi
Aboveground shoots (ramets) were clipped at the stem base, above the soil surface.
In the lab, the samples were cut into 15-20 $\mu m$ cross-sections with a rotary microtome, stained with a mixture of 1:1 safranin and astrablue, rinsed with ethanol solutions, embedded in Euparal, dried and finally scanned in a slide scanner to obtain high resolution images.

\subsection{Compared Methods}

As there are no specialized methods for tree ring detection in shrub cross sections yet, we compare our method with generic instance segmentation methods.
From the top-down category we compare with Mask-R-CNN \cite{maskrcnn} and Deep Snake\cite{deepsnake}.
Mask-R-CNN is trained in two modes:
in the \emph{hollow} (h) mode, objects are defined as single calendar years and are donut-shaped (with a hole), whereas in the \emph{filled} (f) mode, objects consist of multiple years (and have no holes).
We use the implementation from the torchvision (v0.11) framework.
The non-maximum suppression is increased to $0.7$ to reduce the filtering of overlapping detections and the images are downscaled to accommodate for GPU memory limits.
For Deep Snake only the filled mode is used because it cannot model hollow objects.

In the bottom-up group we select Multicut\cite{kappes2011globally} and GASP \cite{GASP} for comparison.
We use the implementation from the PlantSeg \cite{plantseg} source code which was developed in part by the original GASP algorithm authors.
For a fair comparison, the detected boundaries from the same segmentation network $f$ as for INBD are used.
We have found bottom-up methods require species-specific tuning of hyperparameters. We have tested several combinations and report only the best ones here. More information can be found in the supplement.

\subsection{Metrics}

Our main evaluation metric is the mean Average Recall (mAR) as defined in \cite{average_recall} averaged at IoU=.50:.05:.95 intervals.
We do not use the mean Average Precision (mAP) that is often used in generic instance segmentation literature, as we regard instance recall as more important than precision: it is easier for the end user to delete false positive objects on manual inspection than adding new ones.

We additionally report the Adapted Rand errors (ARAND) as defined in \cite{ARAND} because this metric is more commonly used in the bottom-up literature.
It can be interpreted as the harmonic mean of the pixelwise precision and recall values.

%% file: content/05_results.tex
\section{Results}

\input{content/05a_main_results.tex}

\input{content/05b_ablation.tex}

\input{content/05c_cross_species.tex}

\input{content/05d_failurecases.tex}

%% file: content/05a_main_results.tex
\subsection{Method Comparison}
\begin{table*}[]
    \centering
    \begin{tabular}{l|ccc|ccc}
                        & \multicolumn{3}{c}{mAR$\uparrow$}   &  \multicolumn{3}{c}{ARAND$\downarrow$} \\
    \hline
     Method             & DO          & EH             & VM         & DO          & EH         & VM   \\
     \hline
     Mask-R-CNN (h)     & .106 (.008) & .144 (.003)    & .185 (.008)& .644 (.007) & .694 (.002)& .532 (.004)   \\
     Mask-R-CNN (f)     & .210 (.006) & .176 (.004)    & .218 (.002)& .441 (.002) & .499 (.001)& .425 (.007)   \\
     Deep Snake (f)     & .061 (.011) & .015 (.001)    & .019 (.008)& .524 (.024) & .620 (.003)& .584 (.027)   \\
     GASP               & .374 (.002) & .667 (.004)    & .576 (.014)& .313 (.003) & .144 (.003)& .168 (.010)   \\
     Multicut           & .387 (.008) & .688 (.005)    & .596 (.006)& .301 (.001) & .132 (.004)& .154 (.005)   \\
     INBD (ours)        & \B{.553} (.011) & \B{.738} (.018)    & \B{.704} (.014)& \B{.196} (.009) & \B{.113} (.010)& \B{.112} (.007)   \\
    \end{tabular}
    
    \caption{
        \label{tab:main_results}
        Method comparison.
        Values are averaged over 3 full training runs with the standard deviation provided in parentheses.
        (h) refers to the hollow mode, (f) to the filled mode. %\\
        $\uparrow$ denotes higher is better, $\downarrow$ lower is better.
    }
    \vspace{-0.2cm}
\end{table*}

%DO snake: mAR: 0.06149712500902976 0.011003115669792361 ARAND: 0.52439256354645 0.02431703618246265
%EH snake: mAR: 0.015493969278865724 0.0008703820799639251 ARAND: 0.62024806853527 0.003472869971163303
%VM snake: mAR: 0.01940173201284312 0.008422757845140834 ARAND: 0.583921210480206 0.029611112530739604

The main results of the compared methods are presented in Table \ref{tab:main_results}.
For all metrics we observe consistently better performance of INBD over the compared methods.

Top-down methods show very unsatisfactory performance.
The filled mode gives a small performance boost but the results are still too inaccurate to be useful, particularly missing many thin rings.
Deep Snake struggles remarkably, often detecting only one or two rings at most.
We attribute this to its base detector CenterNet \cite{centernet} which inherently fails with concentric objects.

The bottom-up methods can compete with INBD on EH thanks to relatively well recognizable ring boundaries in this subset.
The VM and especially DO subsets on the other hand have much less pronounced and sometimes ambiguous boundaries which often cannot be detected at all.
This is particularly a problem for the bottom-up methods which are then prone to incorrectly merging two rings.
INBD on the other hand can interpolate ambiguous locations (Eq. \ref{eq:ambiguous}).
The results of GASP and Multicut are very similar to each other, as also noted in \cite{plantseg}.

\begin{figure}[h]
    \centering
      \includegraphics[width=\linewidth]{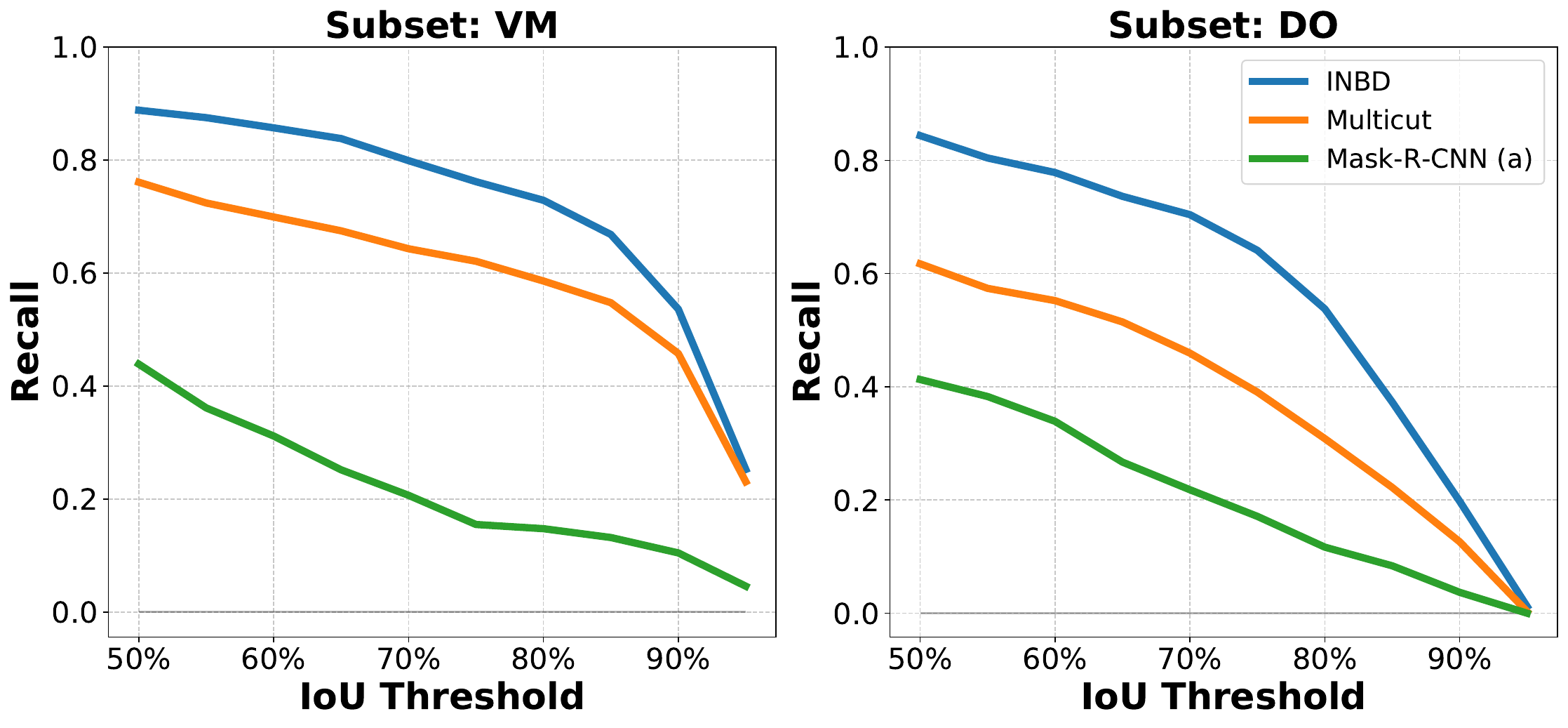}
    \caption{
      \label{fig:recal_over_iou}
      Recall over IoU for the compared methods
      }
\end{figure}

In general, we observe that INBD is better at detecting difficult rings.
This observation is confirmed in the more fine-grained analysis in Figure \ref{fig:recal_over_iou} which shows the recall values for the individual IoU thresholds.
INBD scores only slightly better on the high threshold recalls such as AR90 or AR95 which are usually the easily recognizable rings.
The real benefits come from detecting harder examples.

%% file: content/05b_ablation.tex
\subsection{Ablation Study}
\label{sec:ablation}

In Table \ref{tab:ablation} we show how the individual design choices as proposed in section \ref{section:method} affect the detection performance of INBD.
Two baselines of INBD without the adaptations are evaluated, one with Cartesian and another with polar coordinates.
We note that those two implementations are not fully comparable, more details in the supplement.
%``Baseline'' refers to a naive implementation of INBD without the adaptations listed below.

Our experiments show that increasing the angular resolution (Eq. \ref{eq:angular_resolution}) in order to keep the Cartesian resolution roughly constant across rings yields almost a 3 mAR percentage points improvement.
Interpolating ambiguous boundary points (Eq. \ref{eq:ambiguous2}) is highly important and improves the mAR by more than 6 points.
Iterative training (subsection \ref{subsection:training}), i.e. training with previously predicted boundary points (as apposed to only using the annotation) gives an additional performance boost of more than 3 mAR.
Finally, the WRD module (subsection \ref{subsection:wrd}) helps with wedging rings.
As wedging rings are comparatively few in numbers, the performance gain is relatively moderate but consistent among training runs.

\begin{table}[h]
    \centering
    \begin{adjustbox}{width=\linewidth}
    \begin{tabular}{l|cc}
        Configuration                   &   mAR$\uparrow$           & ARAND$\downarrow$       \\
        %%%%%%%%%%%%%%%%%%%%%%
        \hline
        %%%%%%%%%%%%%%%%%%%%%%
        
        Cartesian coordinates baseline  &  .498                    & .237                 \\
        %\hline
        Polar coordinates baseline      &  .601                    & .218                 \\
        + adaptive angular resolution M &  .629                    & .190                 \\
        + ambiguous boundary interpolation&  .691                  & .146                 \\
        + iterative training            &  .722                    & .126                 \\
        + wedging ring detection        &  .738                    & .113                 \\
    \end{tabular}
    \end{adjustbox}
    
    \caption{
        \label{tab:ablation}
        Influence of design choices on the performance. All values refer to the EH subset.
    }
\end{table}

Additional evaluations on the effect of hyperparameters on the detection performance can be found in the supplementary materials.

%% file: content/05c_cross_species.tex
\input{content/05d_failfigure.tex}

\subsection{Cross-species Performance}

Dendro-ecological studies are rarely limited to the three plant species from our dataset, end users might want to analyze new species, for which trained models are not yet available.
Therefore we test how well the compared methods generalize to unseen species. The results are presented in Table \ref{tab:cross_species}.

\if0

\begin{figure}[h]
    \centering
    \begin{subfigure}{0.49\linewidth}
      \includegraphics[width=\textwidth]{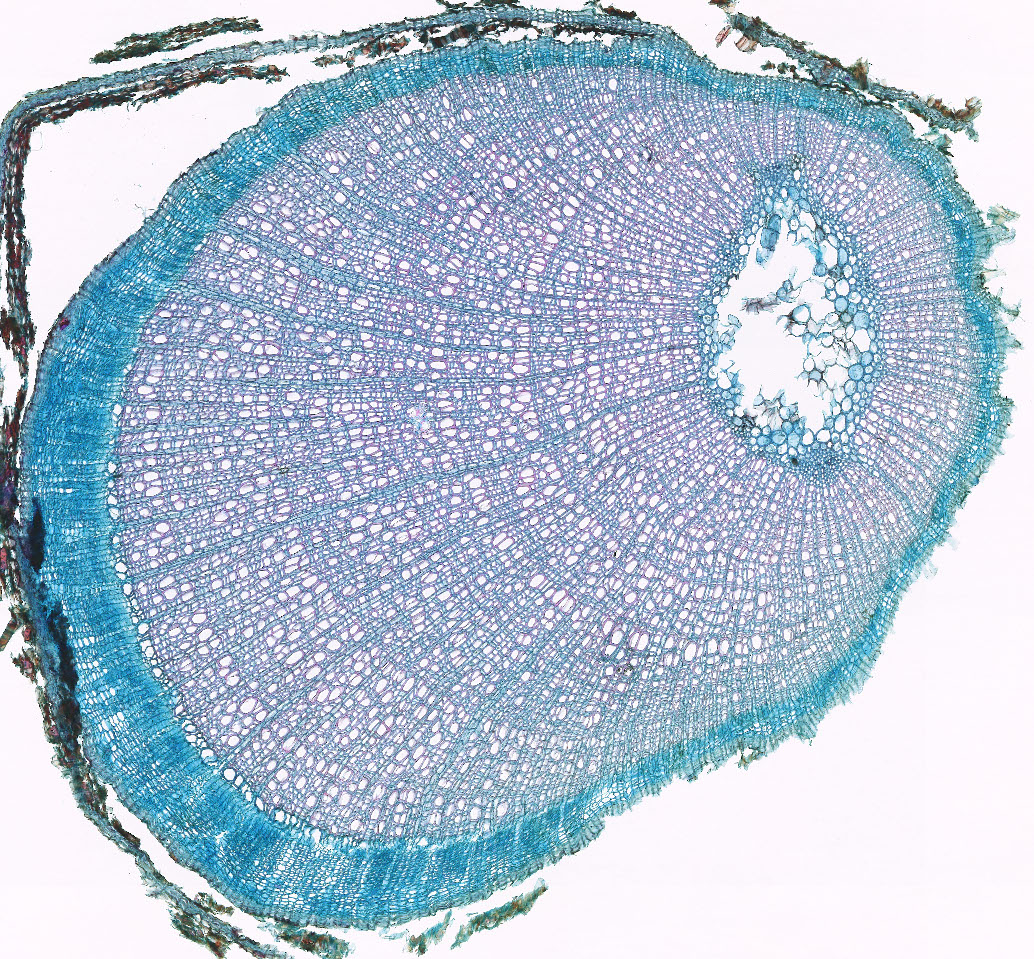}
        \caption{Input}
    \end{subfigure}
    \begin{subfigure}{0.49\linewidth}
      \includegraphics[width=\textwidth]{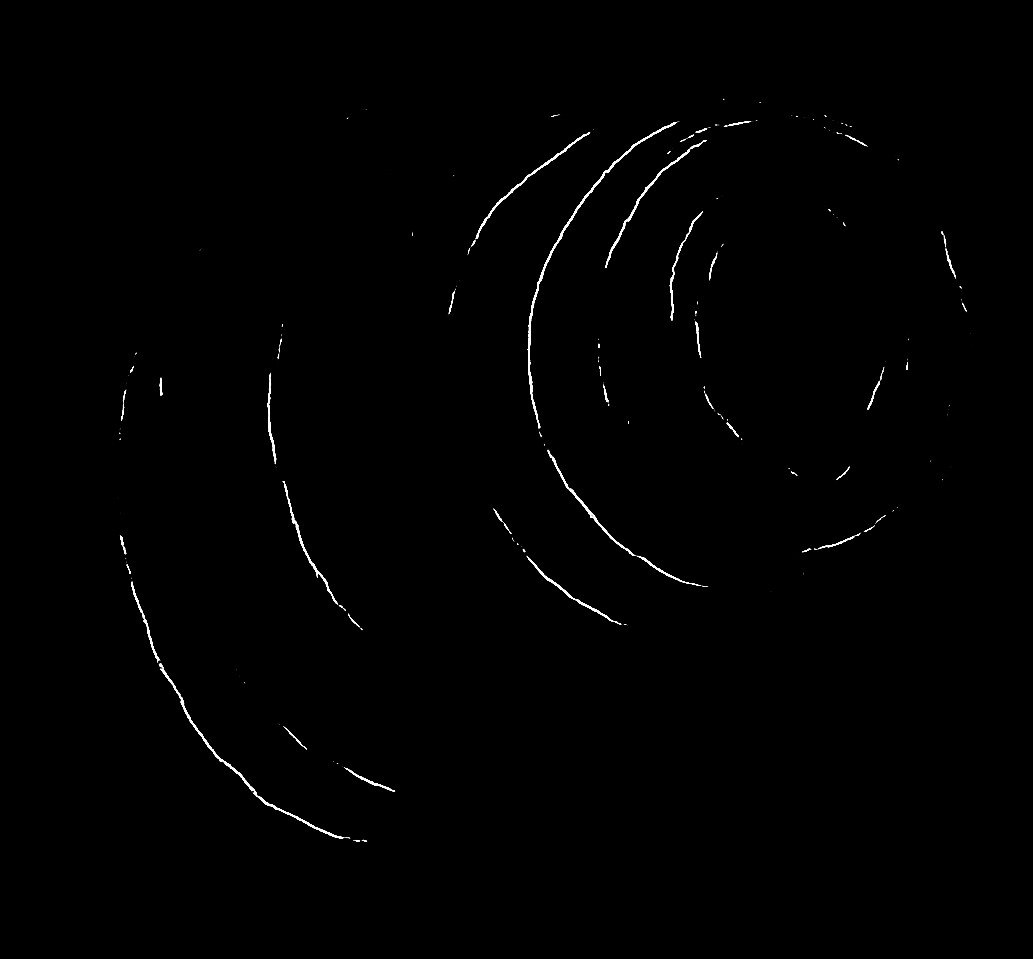}
        \caption{Detected Boundaries}
        %\label{fig:overview-a}
    \end{subfigure}

    \begin{subfigure}{0.49\linewidth}
        \includegraphics[width=\textwidth]{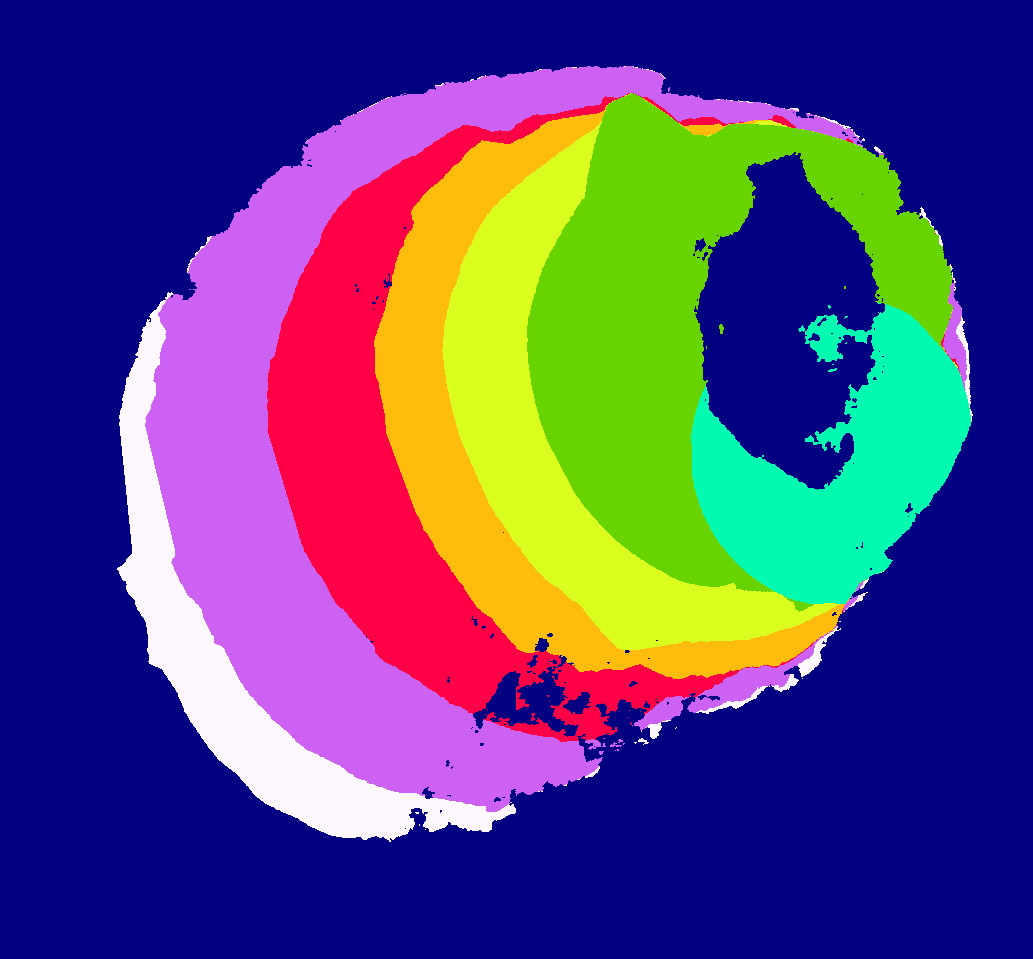}
          \caption{INBD}
      \end{subfigure}
      \begin{subfigure}{0.49\linewidth}
        \includegraphics[width=\textwidth]{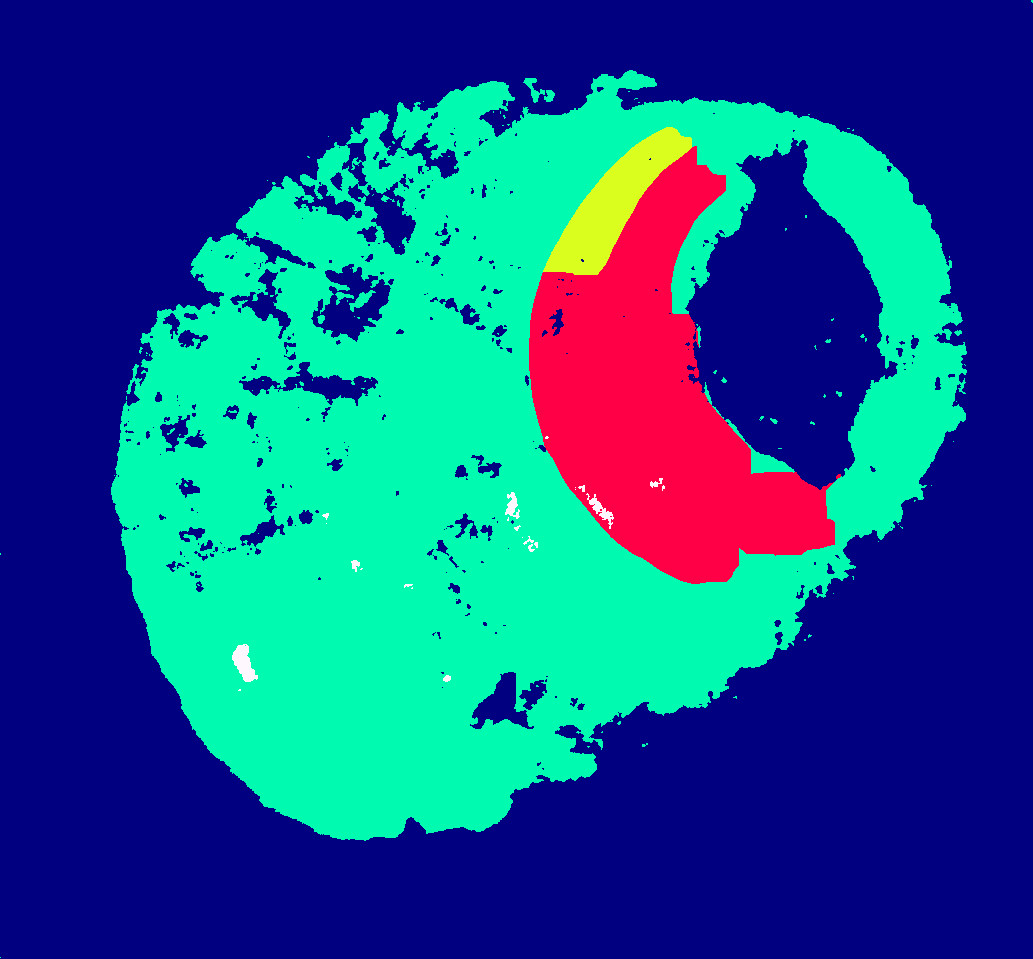}
          \caption{Multicut}
    \end{subfigure}

    \caption{
      \label{fig:cross_species}
      Inference on unseen species.
      Models trained on the EH subset and applied on DO.
      Only few boundaries get detected because the two subsets are visually dissimilar.
      Multicut cannot deal with this, whereas INBD is less reliant on this.
      }
\end{figure}

\fi

\begin{table}[h]
    \centering
    \begin{tabular}{l|ll|cc}
        Method      &   Training set                & Test set         &   mAR$\uparrow$           & ARAND$\downarrow$       \\
        %%%%%%%%%%%%%%%%%%%%%%
        \hline
        %%%%%%%%%%%%%%%%%%%%%%
        INBD        &   EH                          & VM               &  .588                    & .194                 \\
        Multicut    &   EH                          & VM               &  .580                    & .166                 \\
        \hline
        INBD        &   VM                          & EH               &  .472                    & .262                 \\ 
        Multicut    &   VM                          & EH               &  .393                    & .287                 \\ 
        \hline
        INBD        &   EH                          & DO               &  .106                    & .561                 \\
        Multicut    &   EH                          & DO               &  .116                    & .552                 \\
        \hline
        INBD        &   DO                          & EH               &  .219                    & .435                  \\
        Multicut    &   DO                          & EH               &  .169                    & .478                  \\
    \end{tabular}
    
    \caption{
        \label{tab:cross_species}
        Cross species ring detection performance
    }
\end{table}

EH and VM show some level of similarity to each other and methods trained on one set can be used to a limited degree on the other one.
These results might be insufficient for downstream tasks but could be used to generate new annotations for retraining, faster than creating them manually from scratch.
DO on the other hand  is visually dissimilar and requires networks specially trained on it.

Among the methods we observe no clear winner, though INBD is scoring on average slightly better.
The results show that more research needs to be done into this direction.

%% file: content/05d_failfigure.tex
\begin{figure*}[t]
    \centering
    \begin{subfigure}{0.19\linewidth}
      \includegraphics[width=\textwidth]{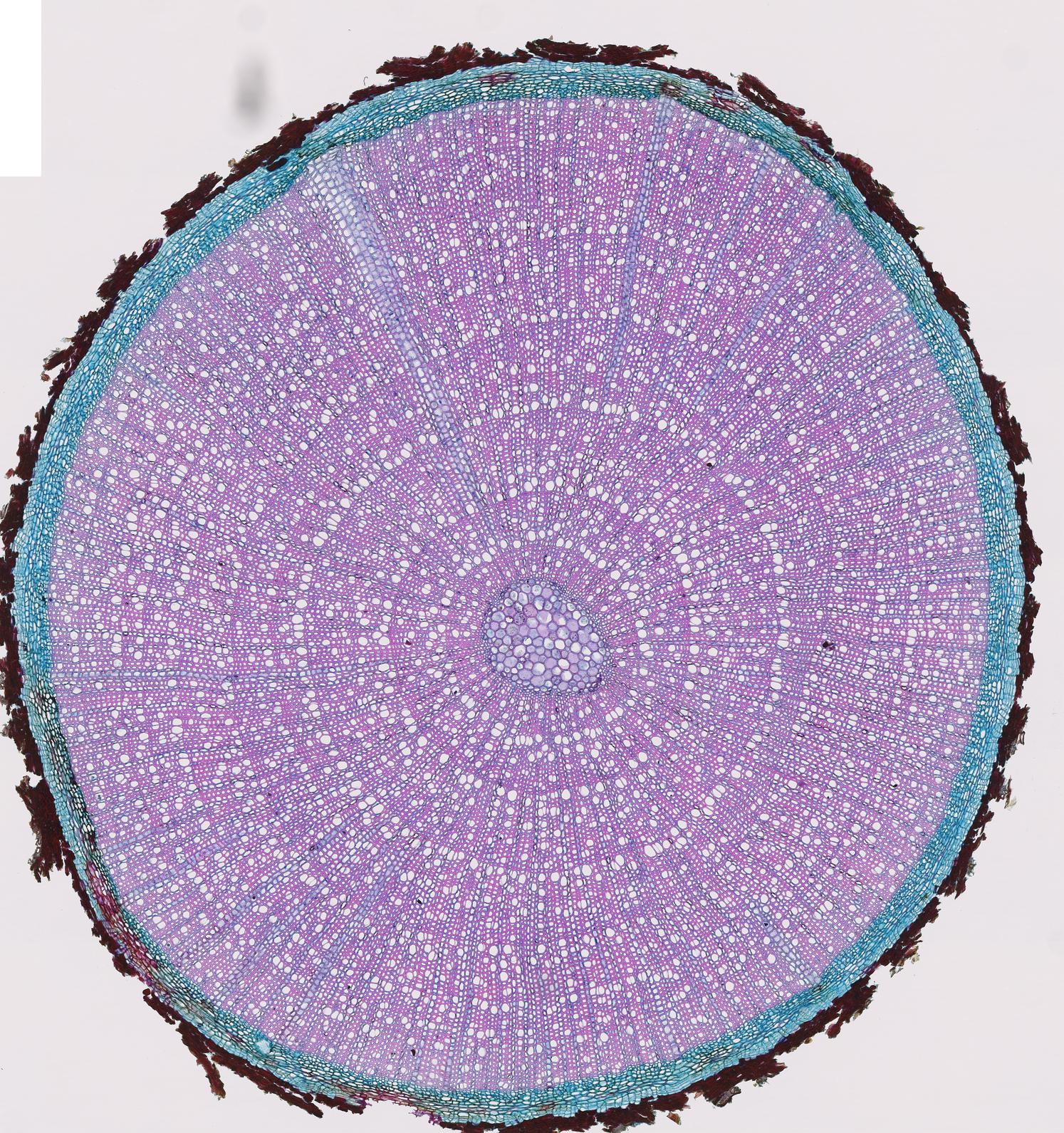}
    \end{subfigure}
    \begin{subfigure}{0.19\linewidth}
        \includegraphics[width=\textwidth]{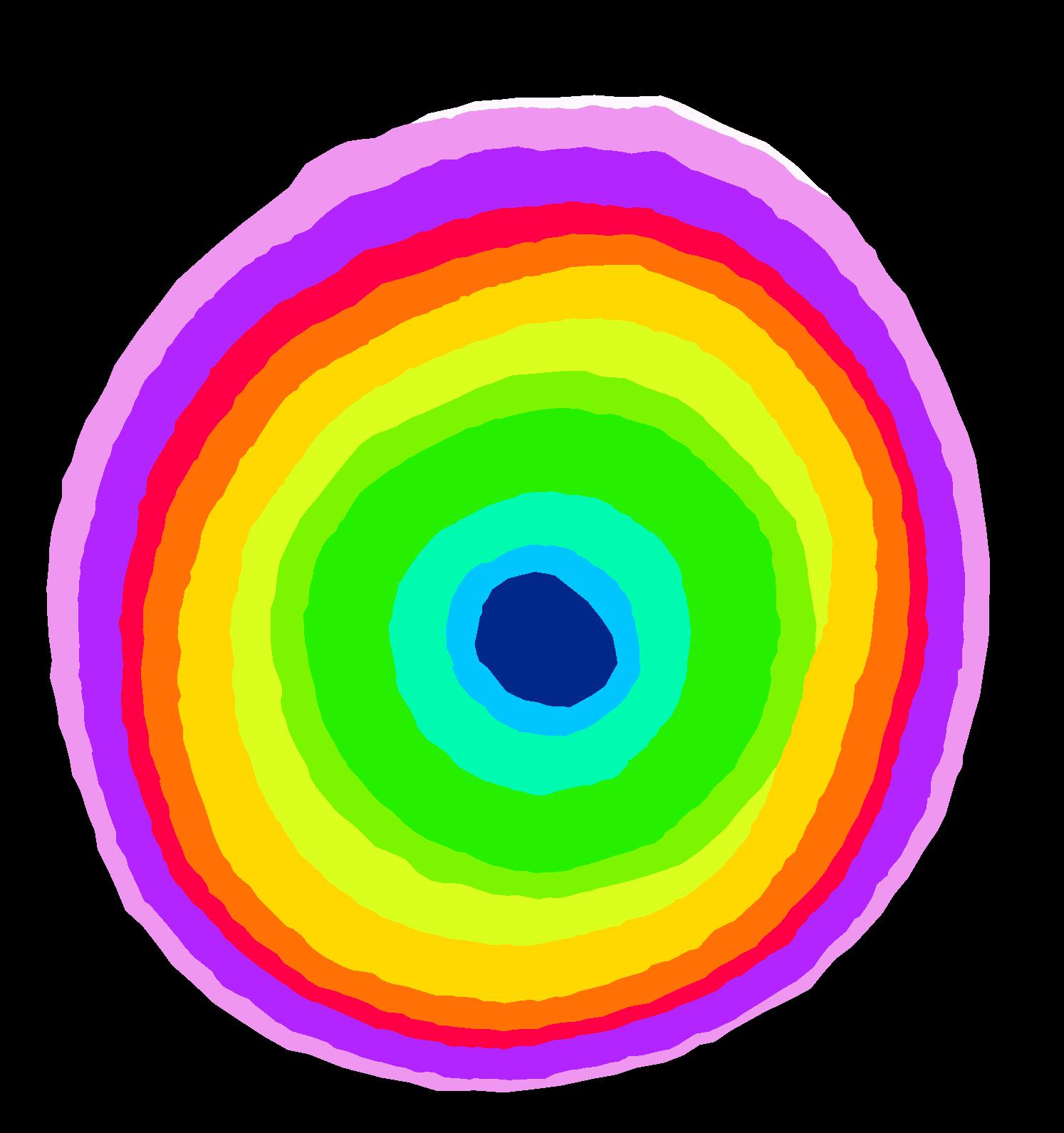} 
    \end{subfigure}
    \begin{subfigure}{0.19\linewidth}
        \includegraphics[width=\textwidth]{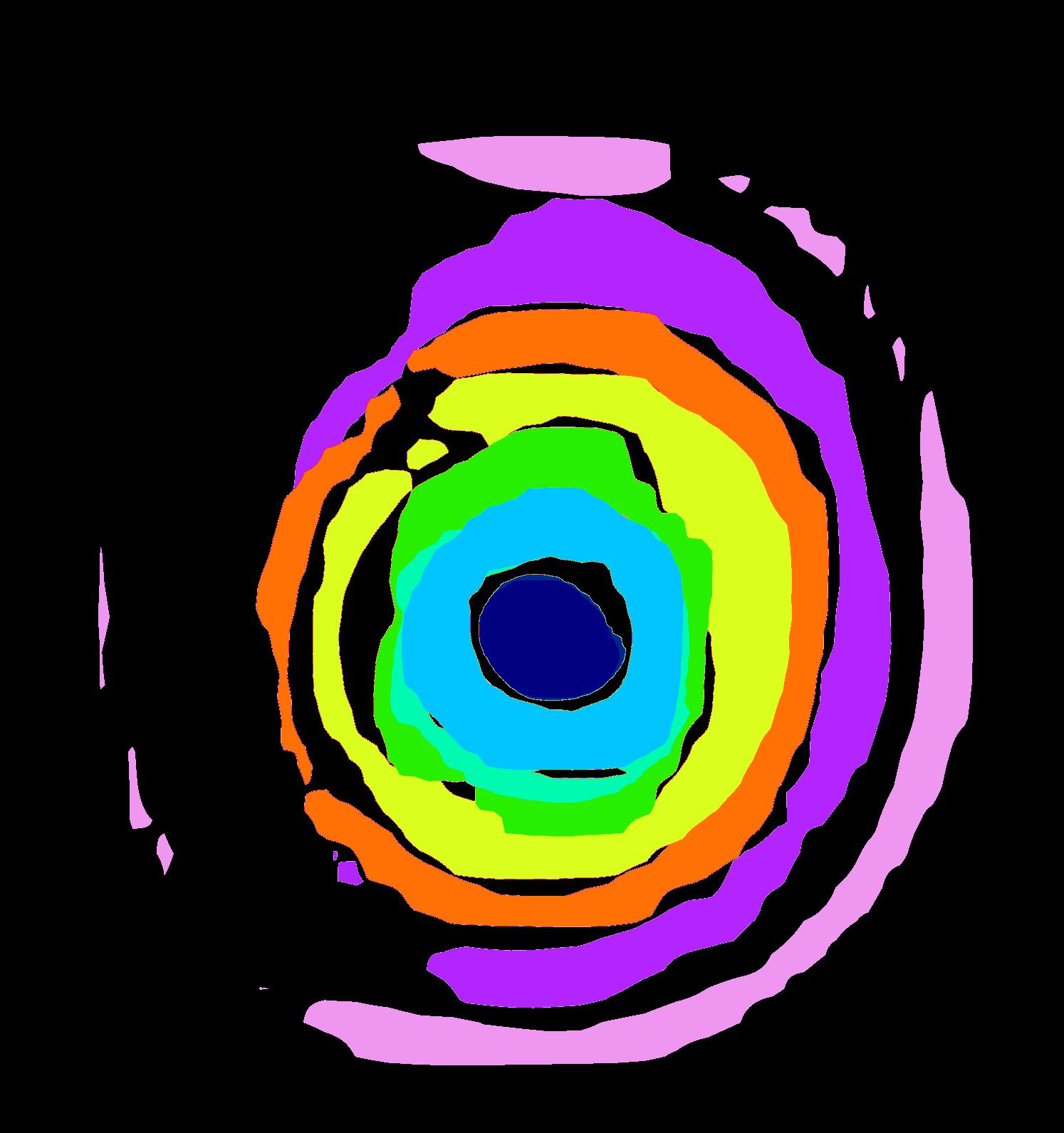}
    \end{subfigure}
    \begin{subfigure}{0.19\linewidth}
        \includegraphics[width=\textwidth]{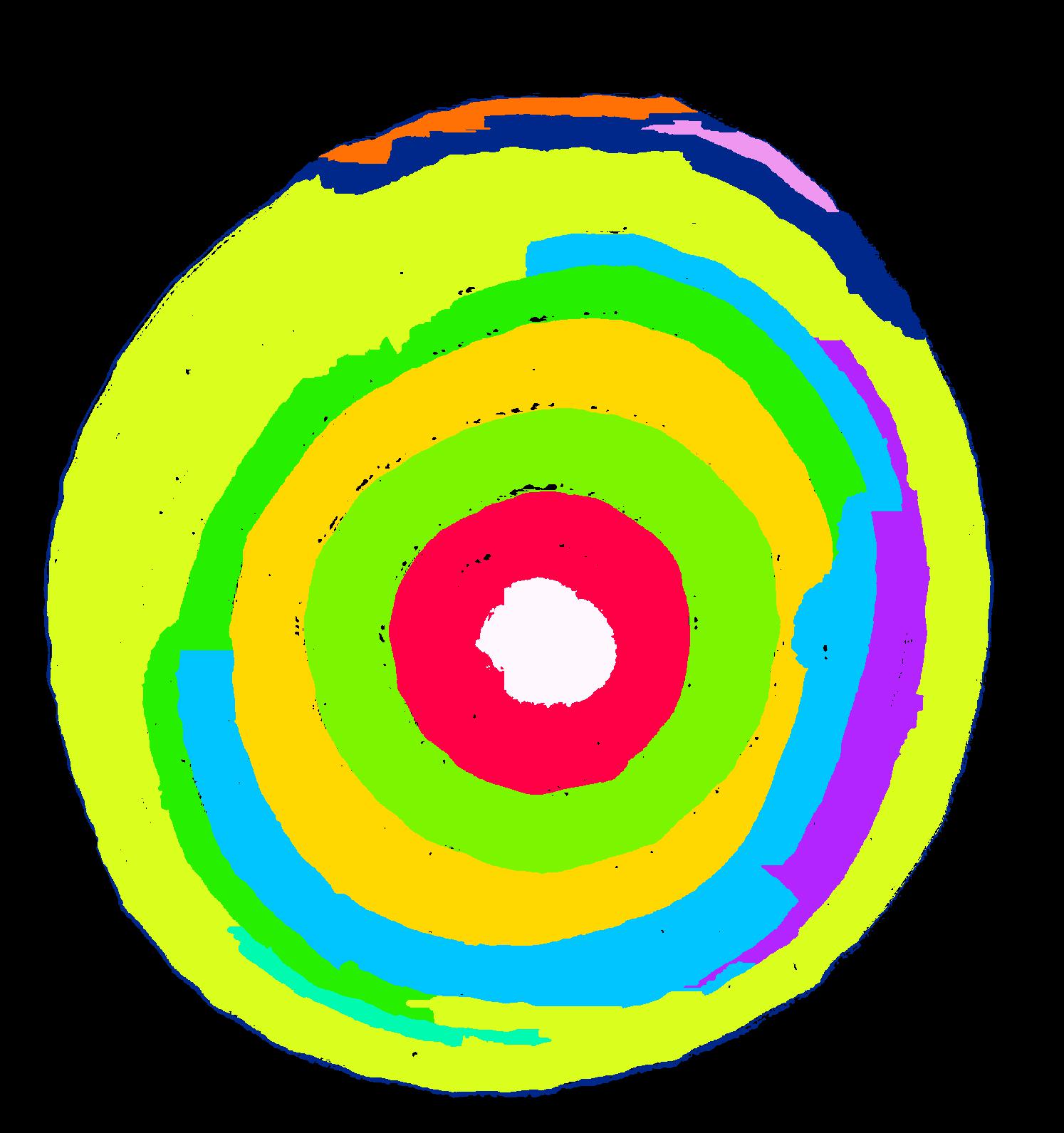}
    \end{subfigure}
    \begin{subfigure}{0.19\linewidth}
        \includegraphics[width=\textwidth]{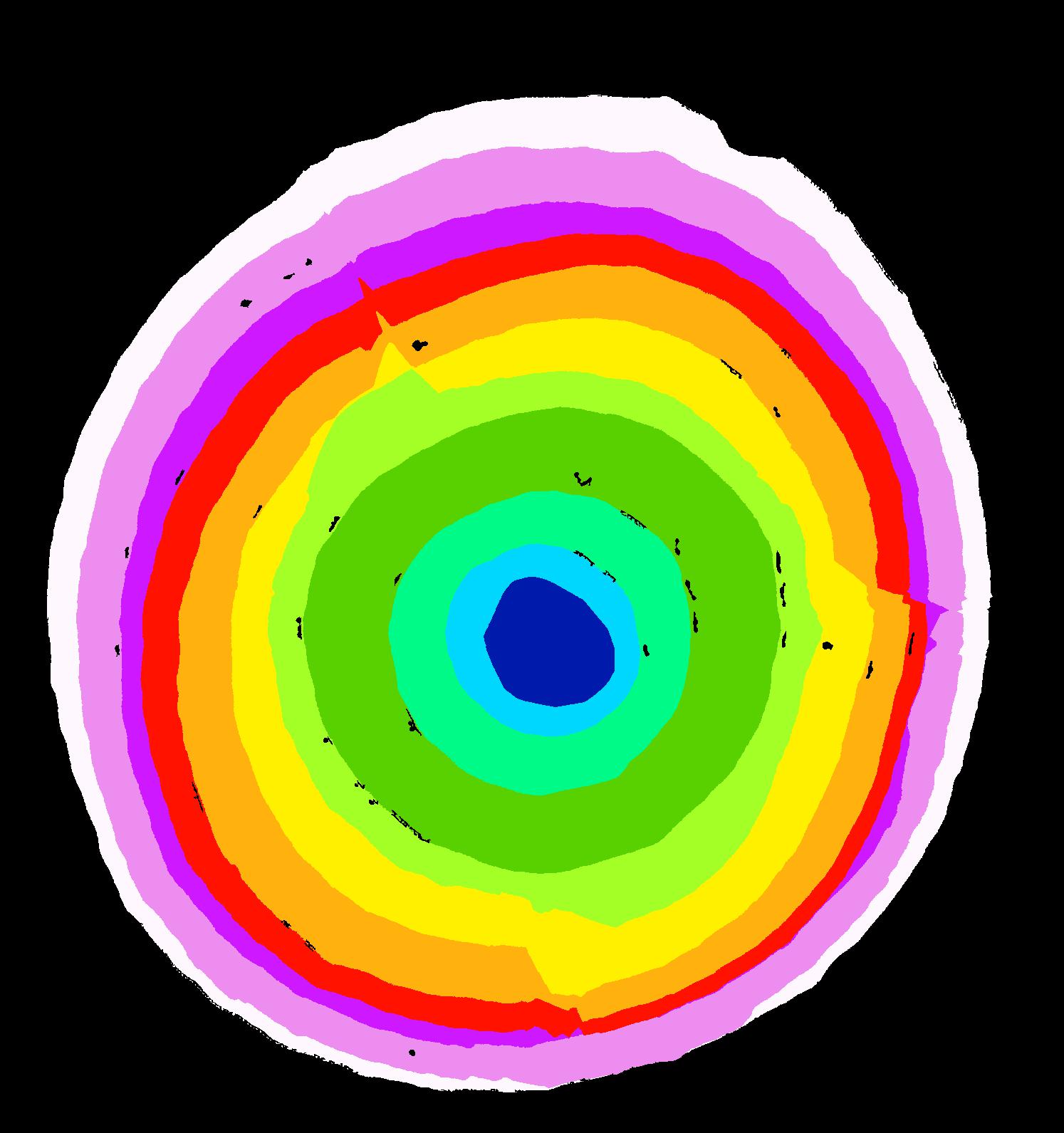} 
    \end{subfigure}
    \begin{subfigure}{0.19\linewidth}
        \includegraphics[width=\textwidth]{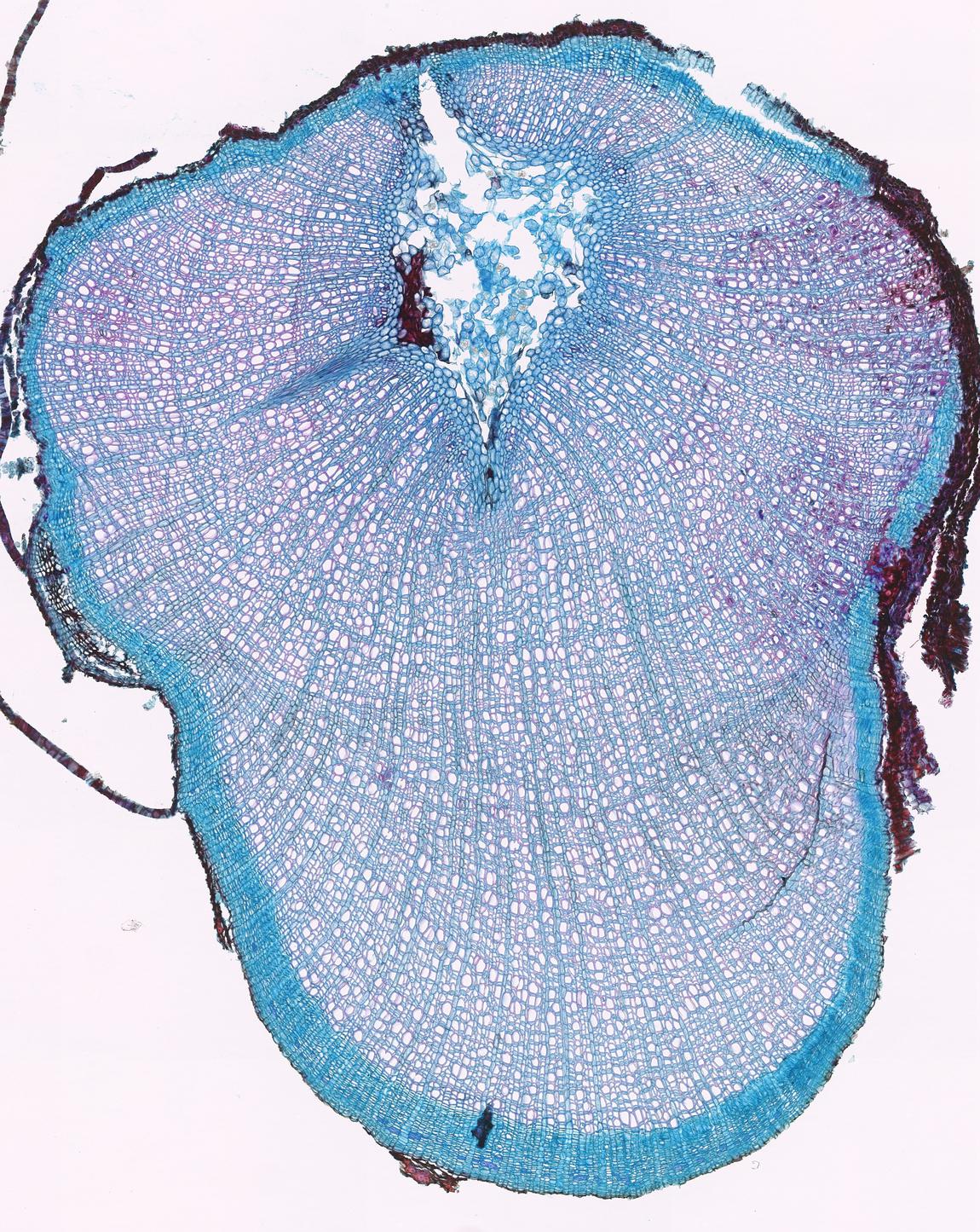}
          \caption{Input}
      \end{subfigure}
      \begin{subfigure}{0.19\linewidth}
          \includegraphics[width=\textwidth]{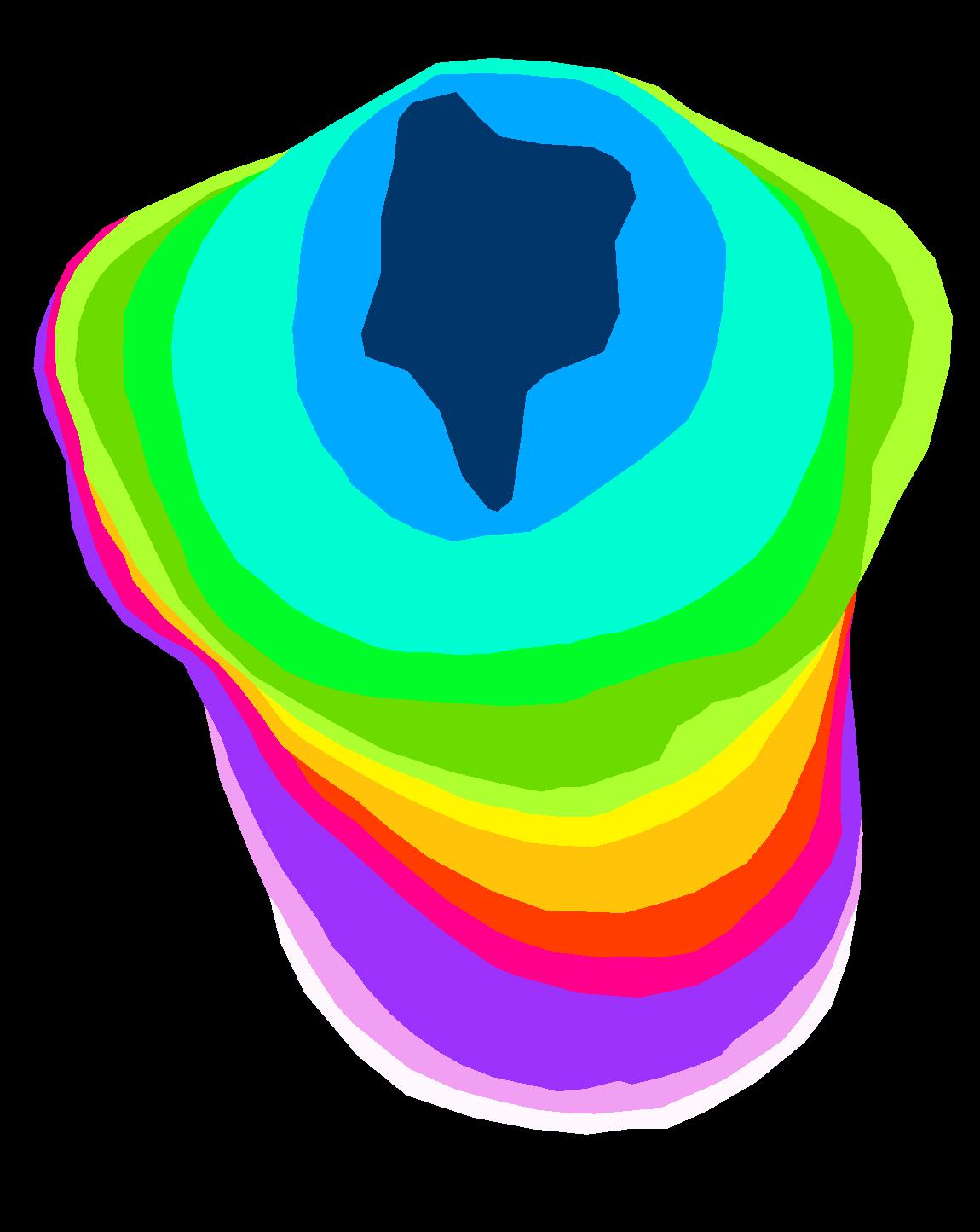}
          \caption{Annotation}
      \end{subfigure}
      \begin{subfigure}{0.19\linewidth}
          \includegraphics[width=\textwidth]{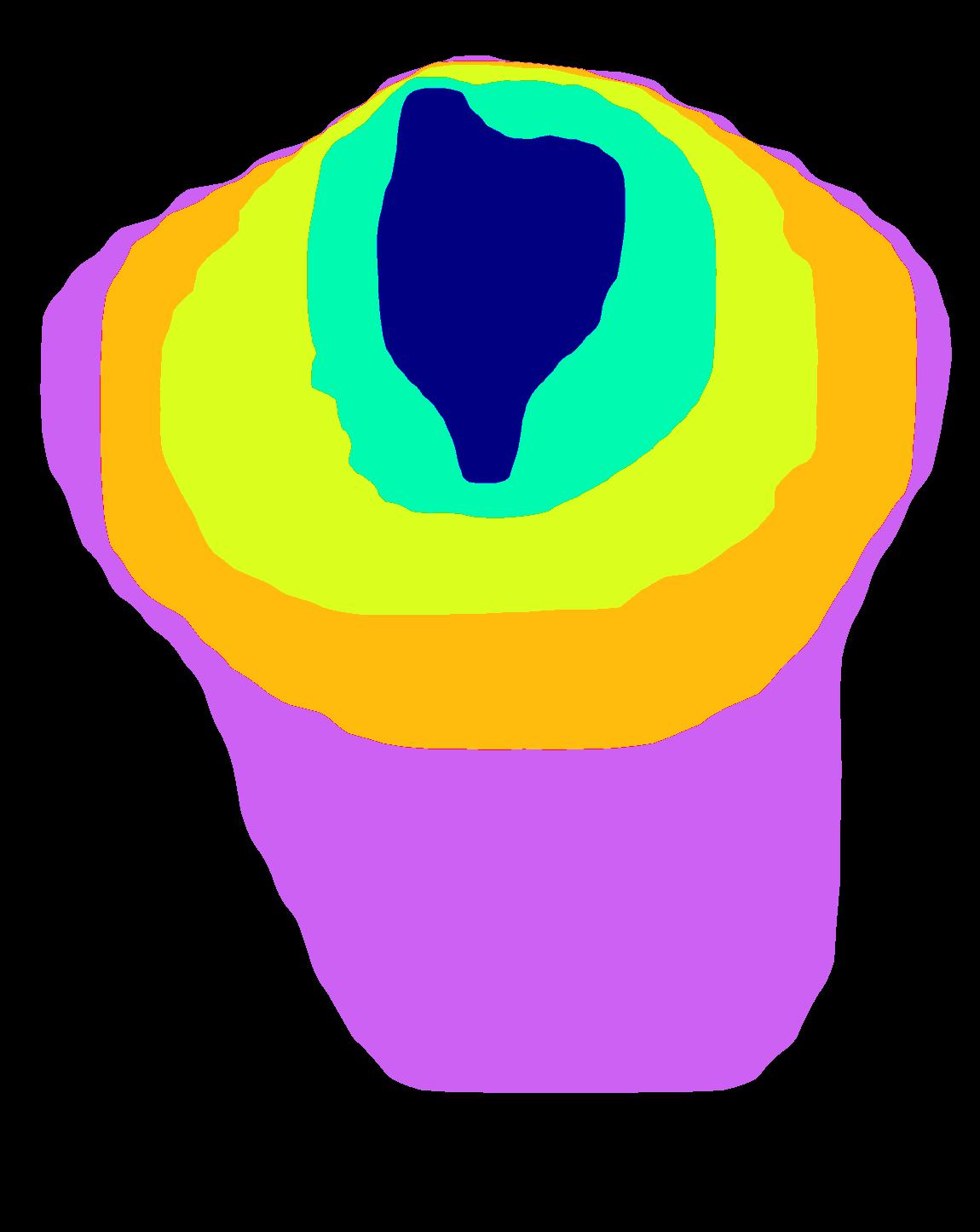}
          \caption{Mask-R-CNN (h)/(f)}
      \end{subfigure}
      \begin{subfigure}{0.19\linewidth}
          \includegraphics[width=\textwidth]{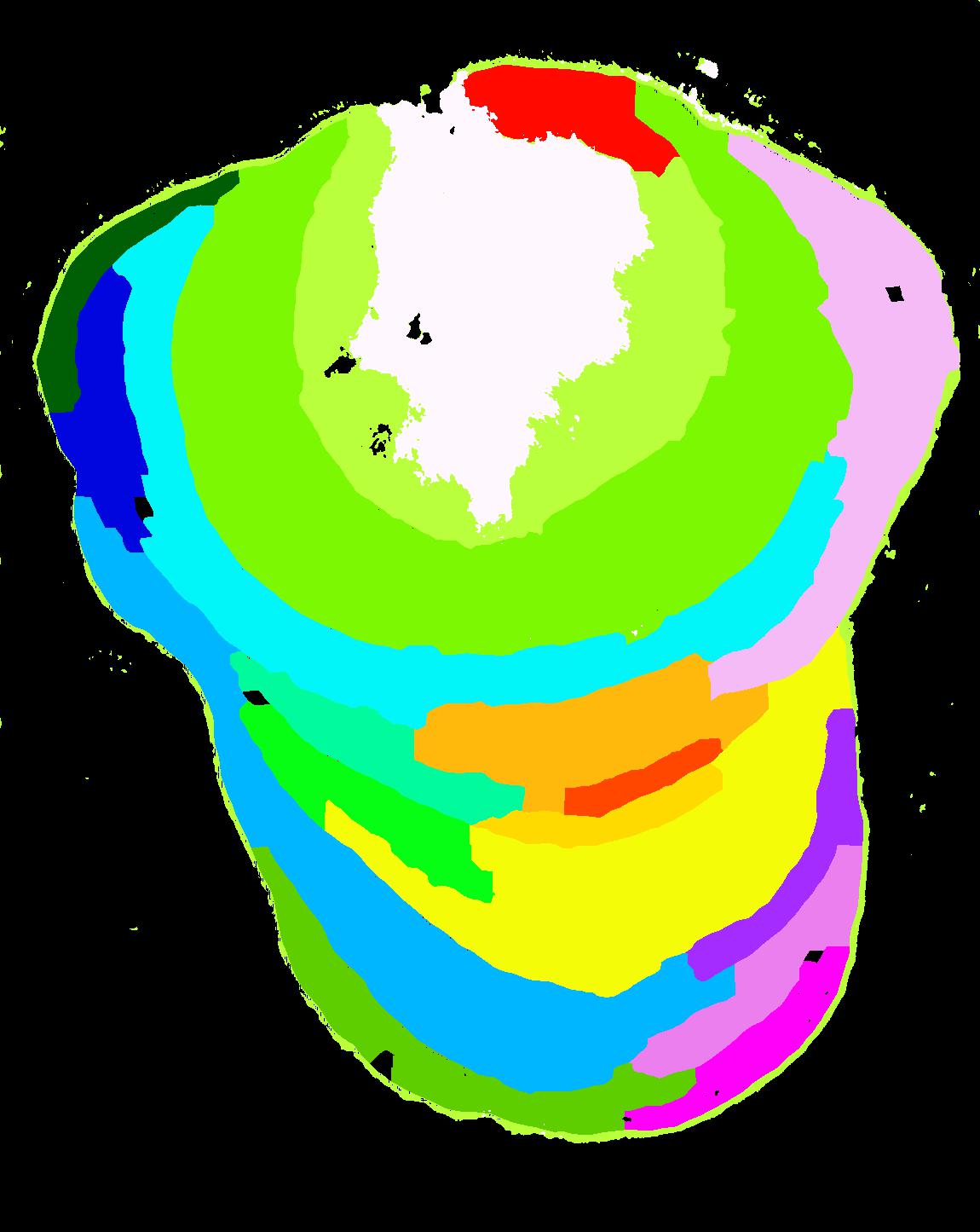}
          \caption{Multicut}
      \end{subfigure}
      \begin{subfigure}{0.19\linewidth}
          \includegraphics[width=\textwidth]{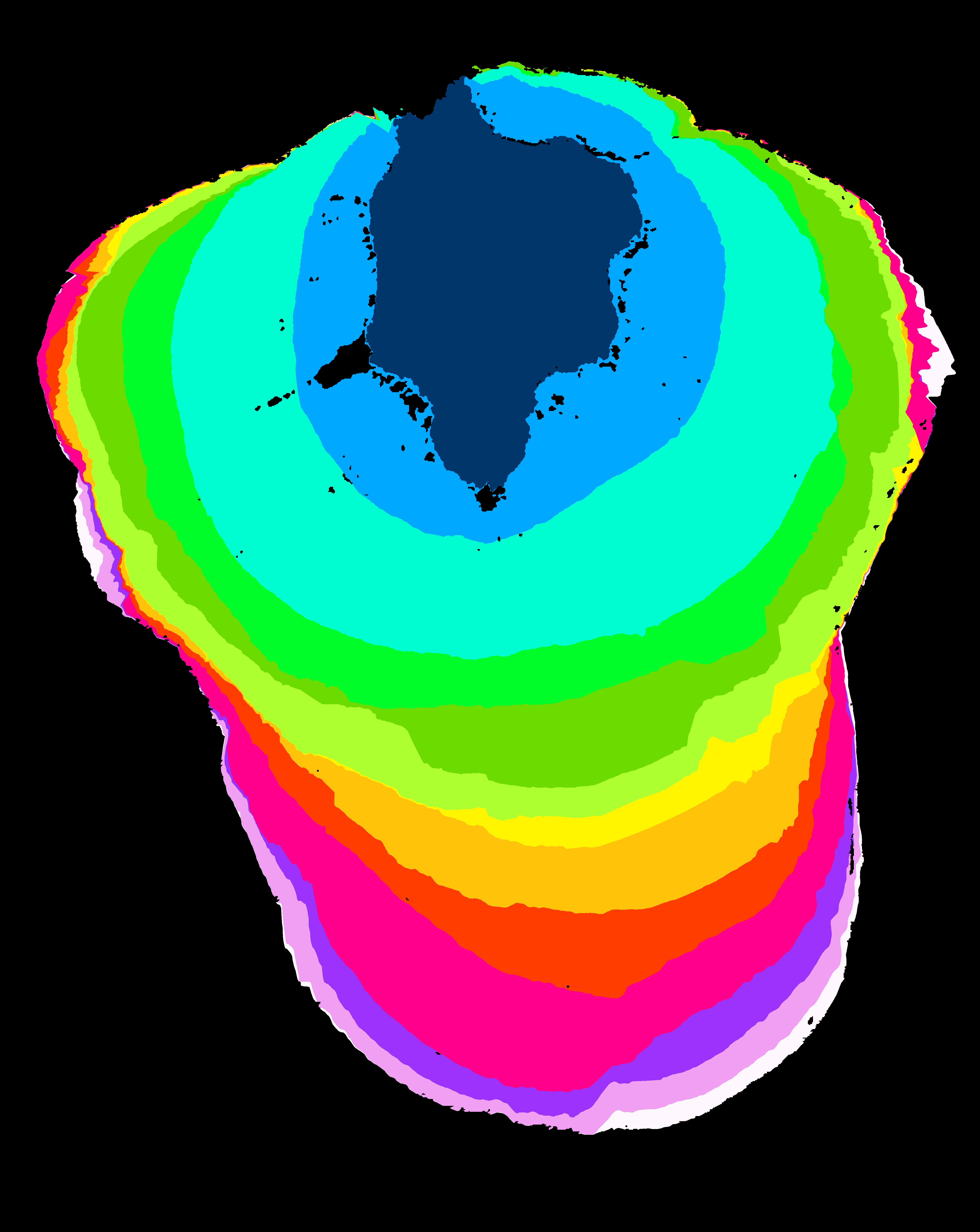}
          \caption{INBD (ours)}
      \end{subfigure}
    
    \caption{
      \label{fig:failurecases}
      Qualitative comparison and examples of typical mistakes made by the compared methods
      }
\end{figure*}

%% file: content/05d_failurecases.tex
\subsection{Qualitative Results}

Figure \ref{fig:failurecases} shows typical mistakes caused by our method as well as the compared top-down and bottom-up procedures.

INBD tends to skip boundaries and this mistake often gets propagated onto the following rings since it is an iterative procedure.
However, thanks to its iterative training procedure and boundary augmentations it can still recover from this.

As expected, the detector-based Mask-R-CNN struggles with the large overlap and fails to detect many rings, and the ones that get detected are very inaccurate.
Bottom-up methods such as Multicut are prone to merging rings where boundaries are difficult to recognize and to splitting them on false positive boundary detections.

More qualitative results can be found in the supplement.

%% file: content/06_conclusion.tex
\section{Concluding Remarks}

Our dataset contains only images for which annotators were confident that they are annotated correctly.
In real-world ecological studies, shrub samples, especially those from harsh climatic conditions, often contain many irregularities in their anatomical structures and may be extremely difficult to fully annotate, even for experts.
In addition, fully annotating images with a large number of rings is very time consuming and costly.
Therefore, future work could focus on weakly supervised training from partially annotated images and on developing methods that provide a confidence estimate for each detected ring or parts of it.

Moreover, as cross section images can vary widely depending on a variety of factors such as plant species, climatic conditions or sample preparation it is not unlikely that a single method trained on a single dataset will not suffice to cover all scenarios.
Further research could be performed on cross-species training for better out-of-distribution generalization.

%% file: content/0x_ack.tex
\if\cameraready1

\section*{ACKNOWLEDGEMENTS}                                                                                                    
This work has been supported by the European Social Fund (ESF) and the Ministry of Education, Science and Culture of Mecklenburg-Vorpommern, Germany  under the project ''DigIT!'' (ESF/14-BM-A55-0015/19).

AAR was funded by a Postdoctoral Research Fellowship from the Alexander von Humboldt Foundation (Germany) and a Juan de la Cierva-Incorporación Grant by the Government of Spain.

\else

\vspace{2cm}
***Acknowledgements placeholder***
\vspace{2.2cm}

\fi

%% file: content/99_supp.tex
\section{Influence of Hyperparameters}

\begin{figure}[h]
    \centering
      \includegraphics[width=0.60\linewidth]{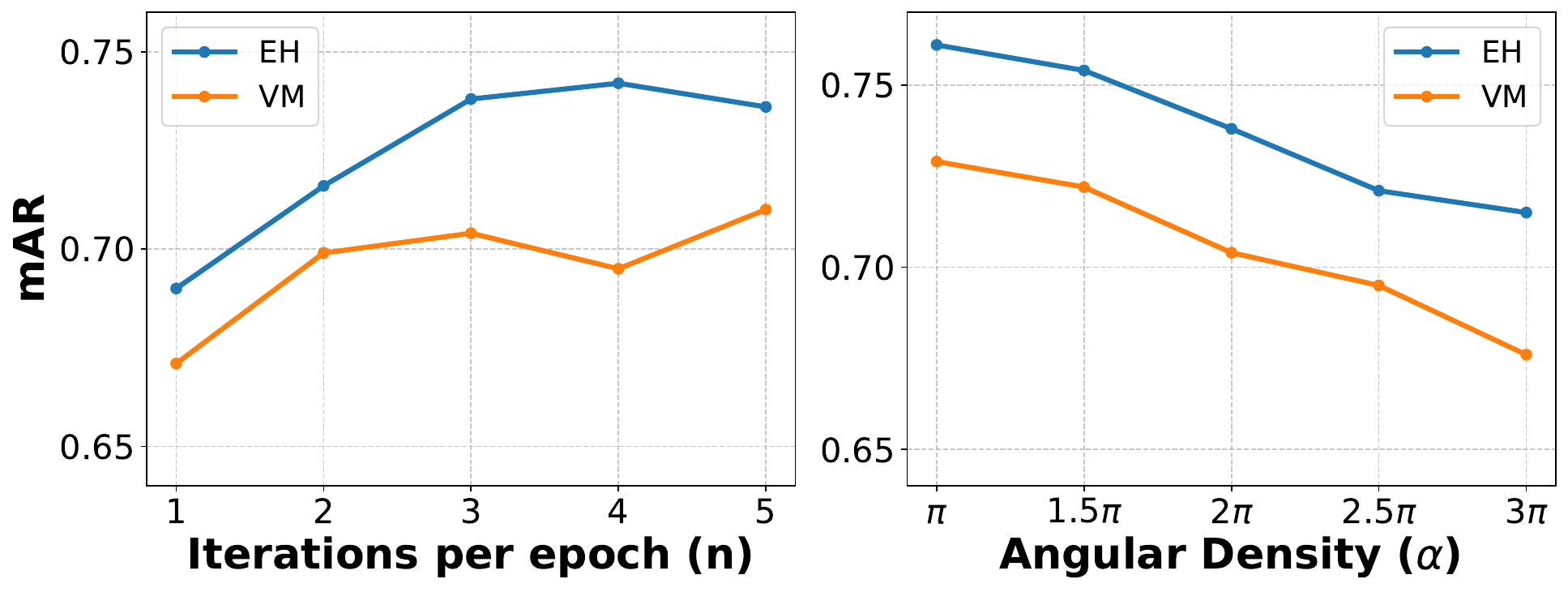}
    \caption{
      \label{fig:inbd_hyperparams}
      Influence of INBD hyperparameters on the detection performance.
      }
\end{figure}

We additionally evaluate the role of the hyperparameters for INBD and Multicut.
Important hyperparameters for INBD are the angular density $\alpha$ that controls the angular resolution $M$ and the number of iterations in one training epoch $n$.
The results of our evaluations are presented in Figure \ref{fig:inbd_hyperparams}.

The performance boost of iterative training diminishes and might even have a detrimental effect after 3 iterations.
Contrary to our expectations and in contrast to other computer vision tasks like image classification, increasing the angular resolution has a negative effect on the detection performance, we attribute to a lower field of view.

\begin{figure}[h]
    \centering
      \includegraphics[width=\linewidth]{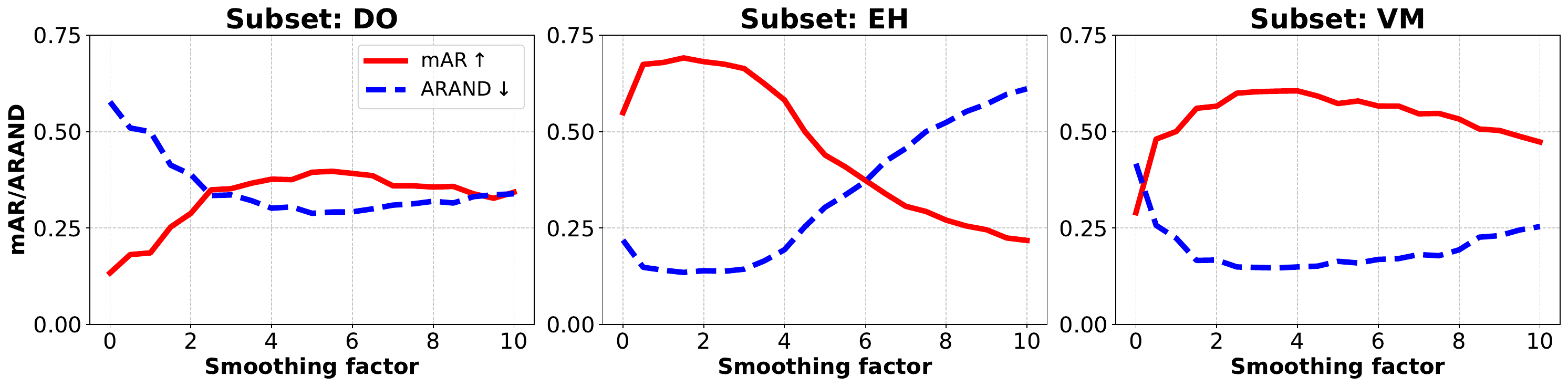}
    \caption{
      \label{fig:multicut_smoothing}
      Influence of the Multicut watershed seed map smoothing factor on the detection performance.
      }
\end{figure}

For Multicut we have found that the smoothing factor for the watershed seed map (referred to as \texttt{sigma\_seeds} in the PlantSeg source code) can be crucial and has to be tuned specifically to the plant species as shown in Figure \ref{fig:multicut_smoothing}.
The results in the main paper show only the best values for each subset.

\section{Network Architecture}

\begin{figure*}[]
    \centering
    \begin{subfigure}{0.37\linewidth}
        \includegraphics[width=\linewidth]{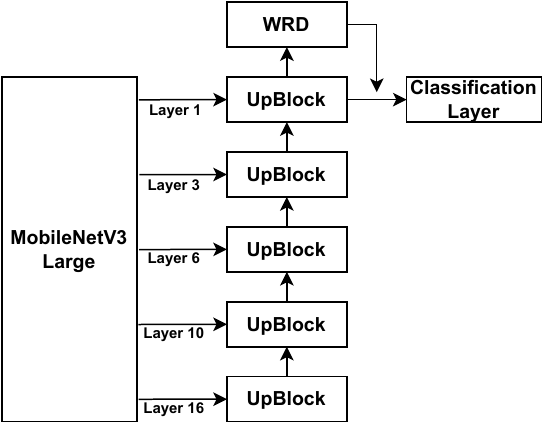}
        \caption{High level overview}
    \end{subfigure}
    \hspace{0.3cm}
    \begin{subfigure}{0.30\linewidth}
        \begin{adjustbox}{width=\linewidth}
        \begin{tabular}{|l|c|c|}
            \hline
            Operator                & Kernel        &   \# of Channels  \\
            \hline
            \texttt{Upsample (x2)}  &               &   x         \\
            \texttt{Concatenation}  &               &   x+y       \\
            \texttt{Conv2D}         & (1,1)         &   y         \\
            \texttt{CirularConv2D}  & (3,3)         &   y       \\
            \texttt{InstanceNorm}   &               &           \\
            \texttt{ReLU}           &               &           \\
            \hline
        \end{tabular}
        \end{adjustbox}
        \caption{INBD network UpBlock (data flow from top to bottom)}
    \end{subfigure}
    \hfill
    \begin{subfigure}{0.28\linewidth}  %
        \begin{adjustbox}{width=\linewidth}
        \begin{tabular}{|l|c|}
            \hline
            Operator                & Kernel/Stride \\
            \hline
            \texttt{MaxPool2D}      & (2,1)         \\
            \texttt{Conv2D}         & (1,1)         \\
            \texttt{InstanceNorm}   &               \\
            \texttt{ReLU}           &               \\
            \hline
            \texttt{MaxPool2D}      & (2,1)         \\
            \texttt{Conv2D}         & (1,1)         \\
            \texttt{InstanceNorm}   &               \\
            \texttt{ReLU}           &               \\
            \hline
            \texttt{Conv2D}         & (1,1)         \\
            \hline
        \end{tabular}
        \end{adjustbox}
        \caption{Wedging Ring Detection (WRD) Module}
    \end{subfigure}
\caption{INBD network architecture \label{fig:network_architecture}}
\end{figure*}

For better reproducibility, we report more details on the used network architectures in Figure \ref{fig:network_architecture}, however we note that our method is not dependent on this specific architecture, other segmentation networks should work as well.
For all our experiments we have used a network architecture based on U-Net with a pretrained MobileNetV3-Large\cite{mobilenetv3} backbone as implemented in torchvision (v0.11).
This backbone was chosen as a compromise between prediction performance and speed: the high image resolution puts limits on the network size for both training and inference on an end user's device.
Circular convolutions \cite{deepsnake} are also used in the backbone and the circularity only applies to the angular axis, not to the radial one.

\section{INBD with Cartesian Coordinates}

\begin{figure}
  \centering
  \begin{subfigure}{0.24\linewidth}
    \includegraphics[width=\textwidth, height=40mm]{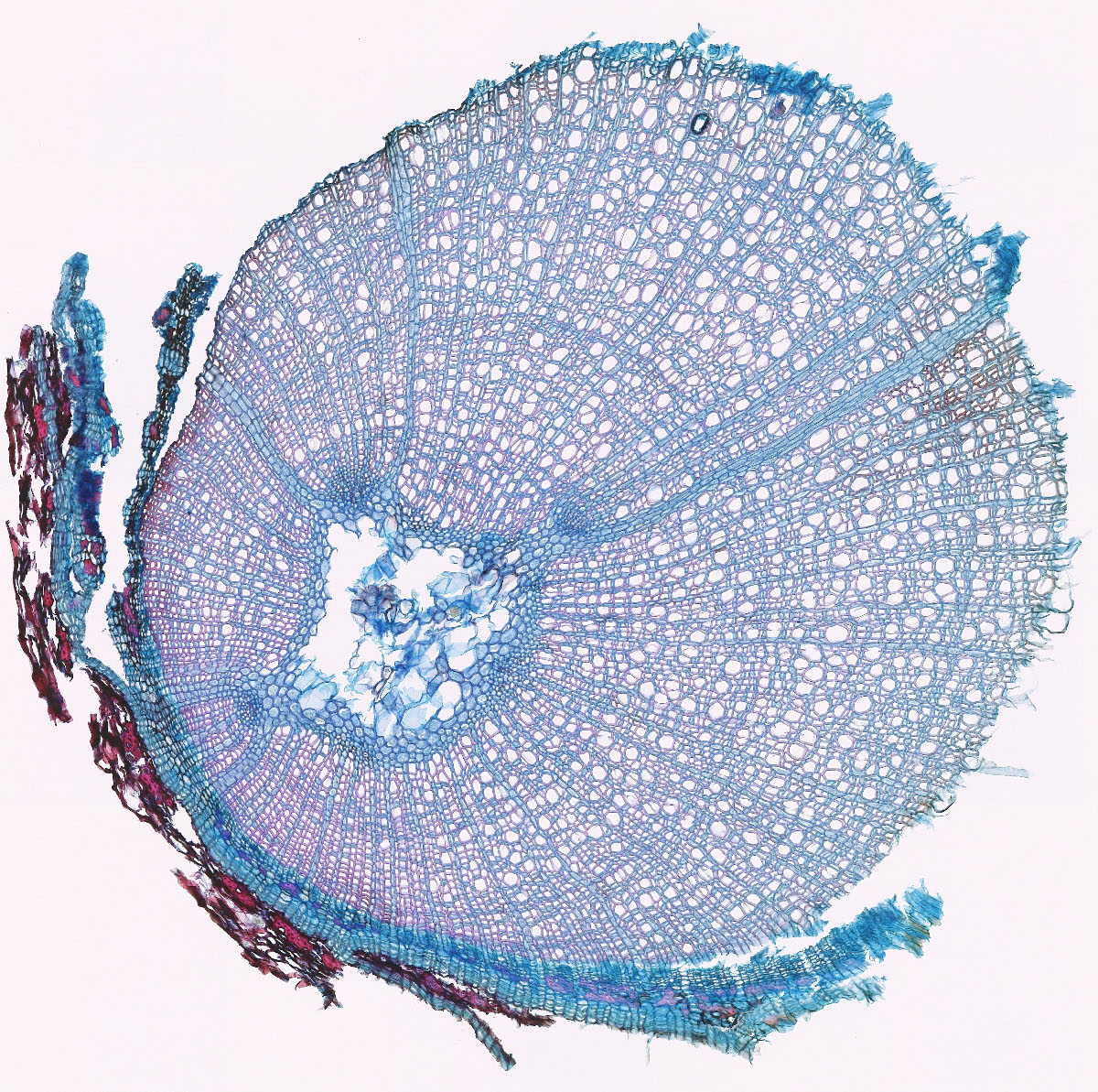}
    \caption{Input}
  \end{subfigure}
  \begin{subfigure}{0.24\linewidth}
      \includegraphics[width=\textwidth, height=40mm]{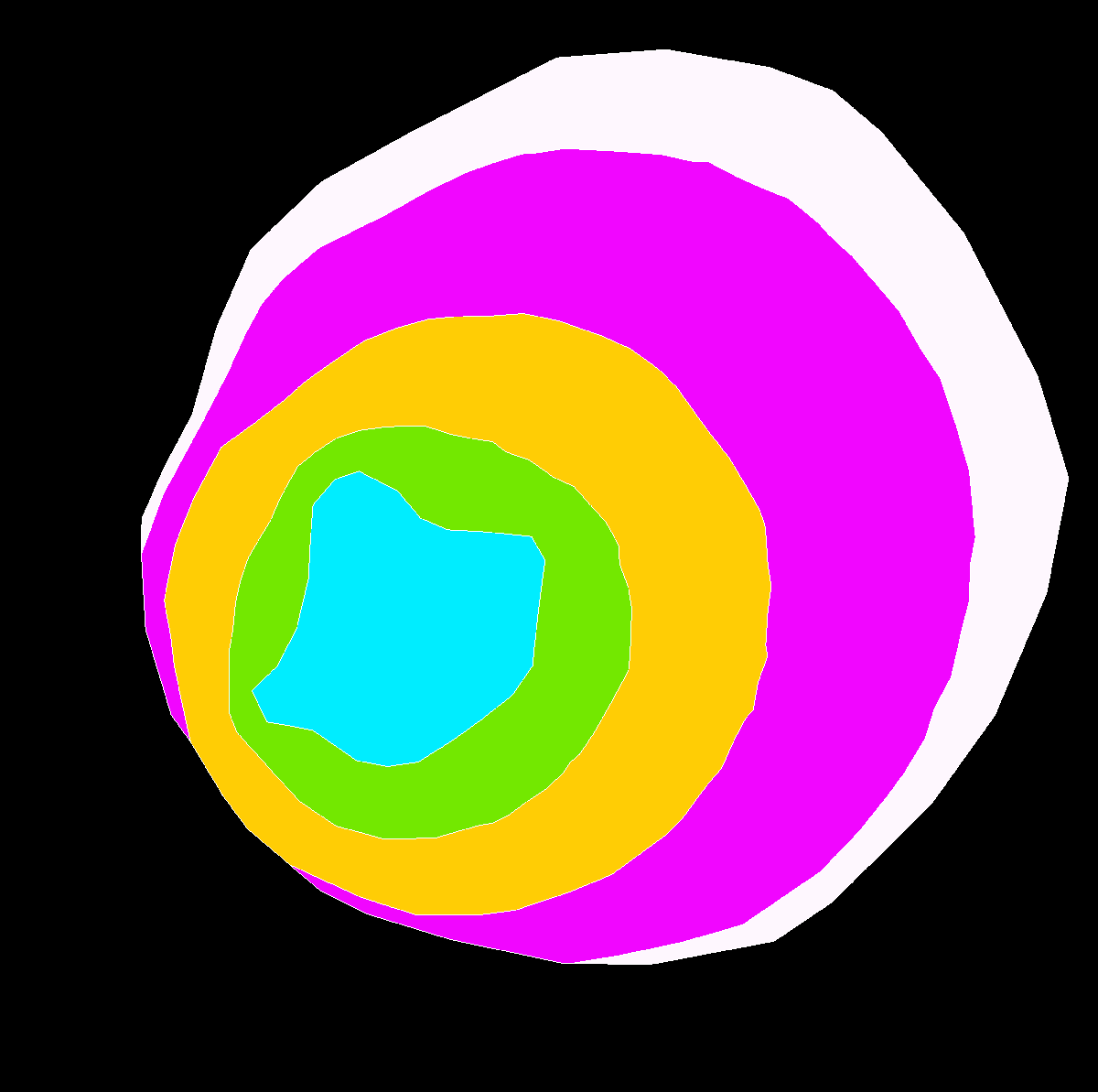}
      \caption{Annotation}
    \end{subfigure}
  \begin{subfigure}{0.24\linewidth}
      \includegraphics[width=\textwidth, height=40mm]{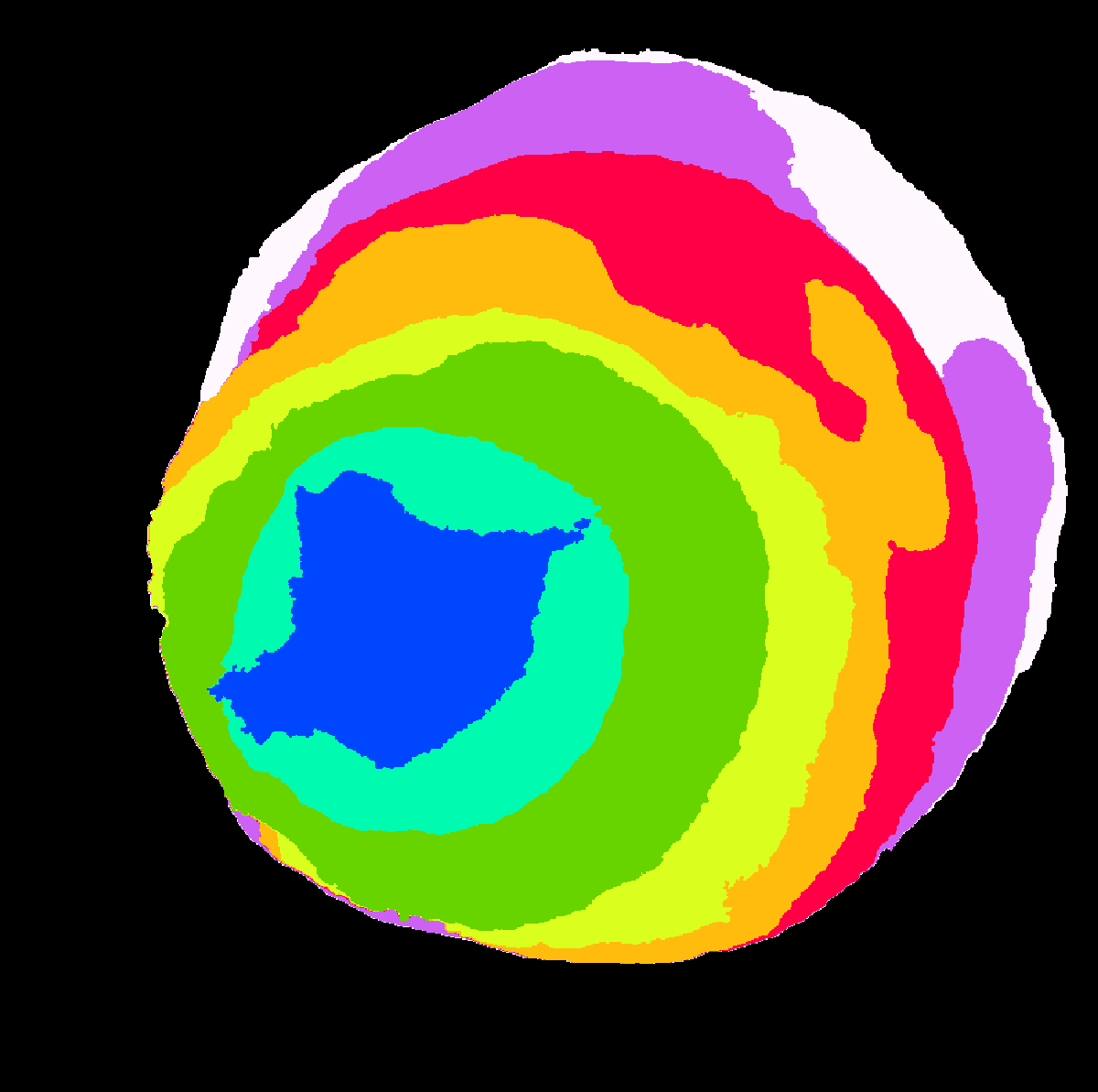}
      \caption{INBD with Cartesian coordinates \label{subfig:nonconvex}}
  \end{subfigure}
  \begin{subfigure}{0.24\linewidth}
      \includegraphics[width=\textwidth, height=40mm]{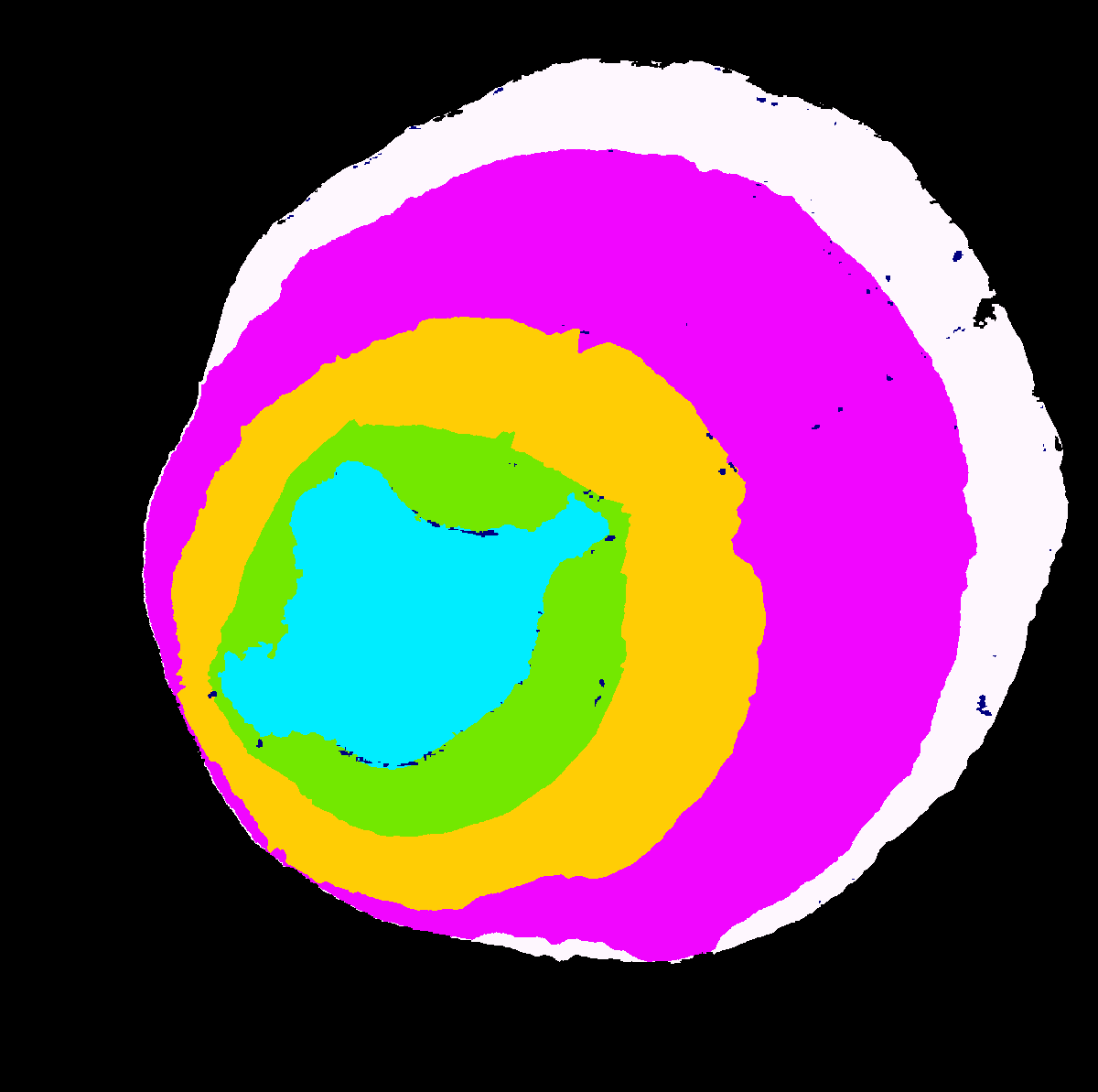}
      \caption{INBD with polar coordinates}
  \end{subfigure}

  \caption{
    \label{fig:nonconvex}
    An image from the DO subset and comparison of INBD with Cartesian and polar coordinates.
    Note the typical nonconvex artifact on the orange ring in \ref{subfig:nonconvex}.
    }
\end{figure}

INBD can in theory work with Cartesian coordinates as well, with the advantage that it is significantly easier to implement.
We also evaluate how well this alternative performs.
For this, we use the same architecture except with standard convolutions and without WRD.
In each iteration step $i$ this network receives the outputs of the 3-class segmentation network $f$ as well as all previously detected rings and it is trained to segment the next ring $i+1$, akin to the our main method, but working on full images and not on polar grids.
A basic result and comparison with polar coordinates can be found in Table 3 of the main paper.
We observe that this alternative is prone to nonconvexities, an example is shown in Figure \ref{subfig:nonconvex}. Polar coordinates on the other hand impose a prior on the shape, ensuring that it is coherent and (quasi-)convex.

\begin{figure}[h]
  \centering
    \includegraphics[width=0.60\linewidth]{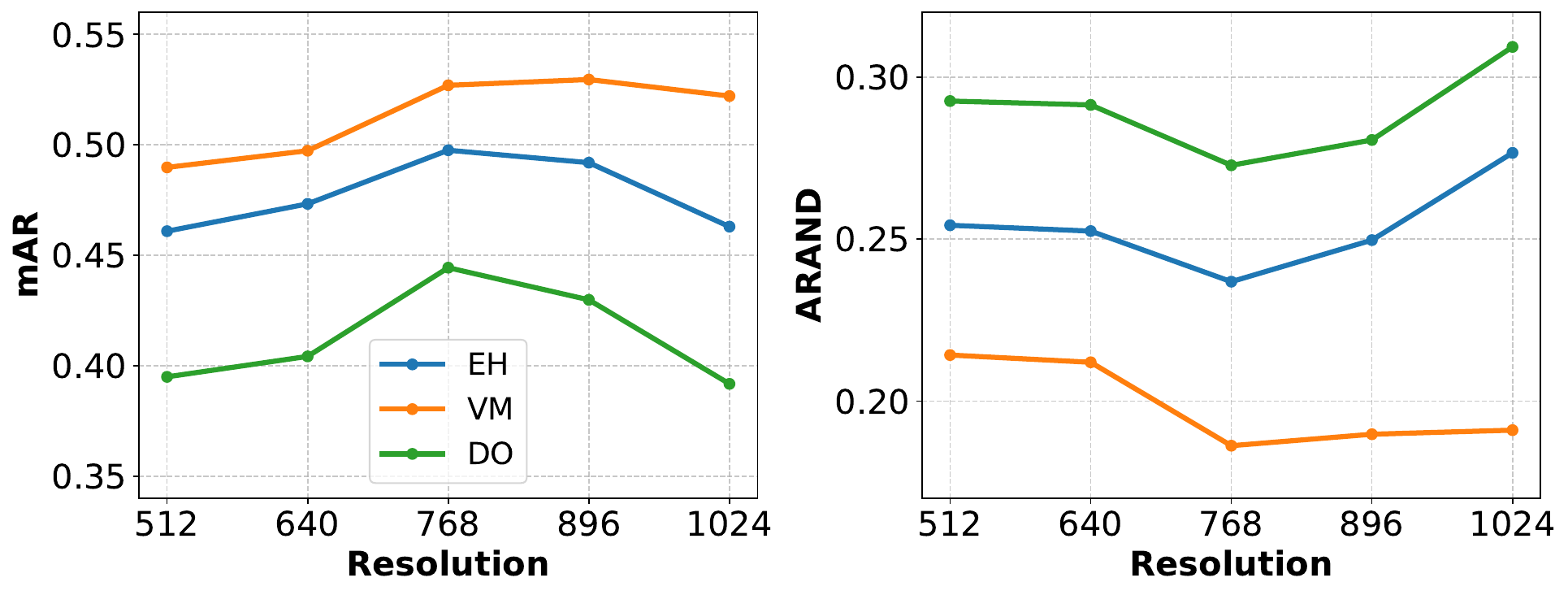}
  \caption{
    \label{fig:cartesians}
    Influence of the image resolution on the performance of the Cartesian baseline
    }
\end{figure}

We note that the image resolution has some influence on the overall detection performance:
a high resolution allows for recognition of very indistinct boundaries but comes at the cost of a lower field of view which is needed for long-range dependencies and a consistent ring segmentation.
An evaluation of the influence is shown in Figure \ref{fig:cartesians}.
For our dataset, the optimal resolution lies around 768$\times$768 pixels.

\section{Additional Images}

\begin{figure}
  \centering
  \begin{subfigure}{0.24\linewidth}
    \includegraphics[width=\textwidth, height=40mm]{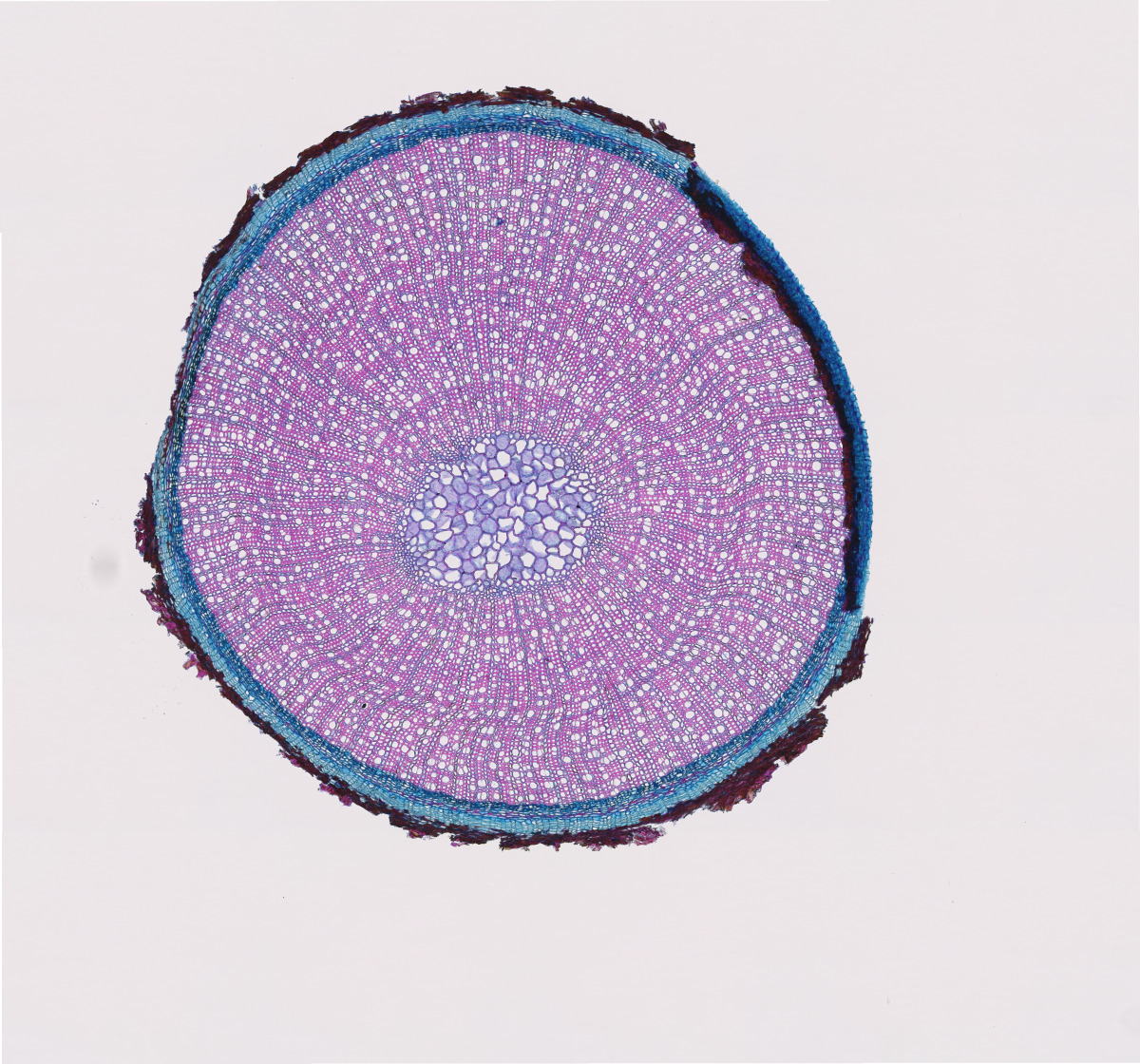}
    \caption{Input}
  \end{subfigure}
  \begin{subfigure}{0.24\linewidth}
      \includegraphics[width=\textwidth, height=40mm]{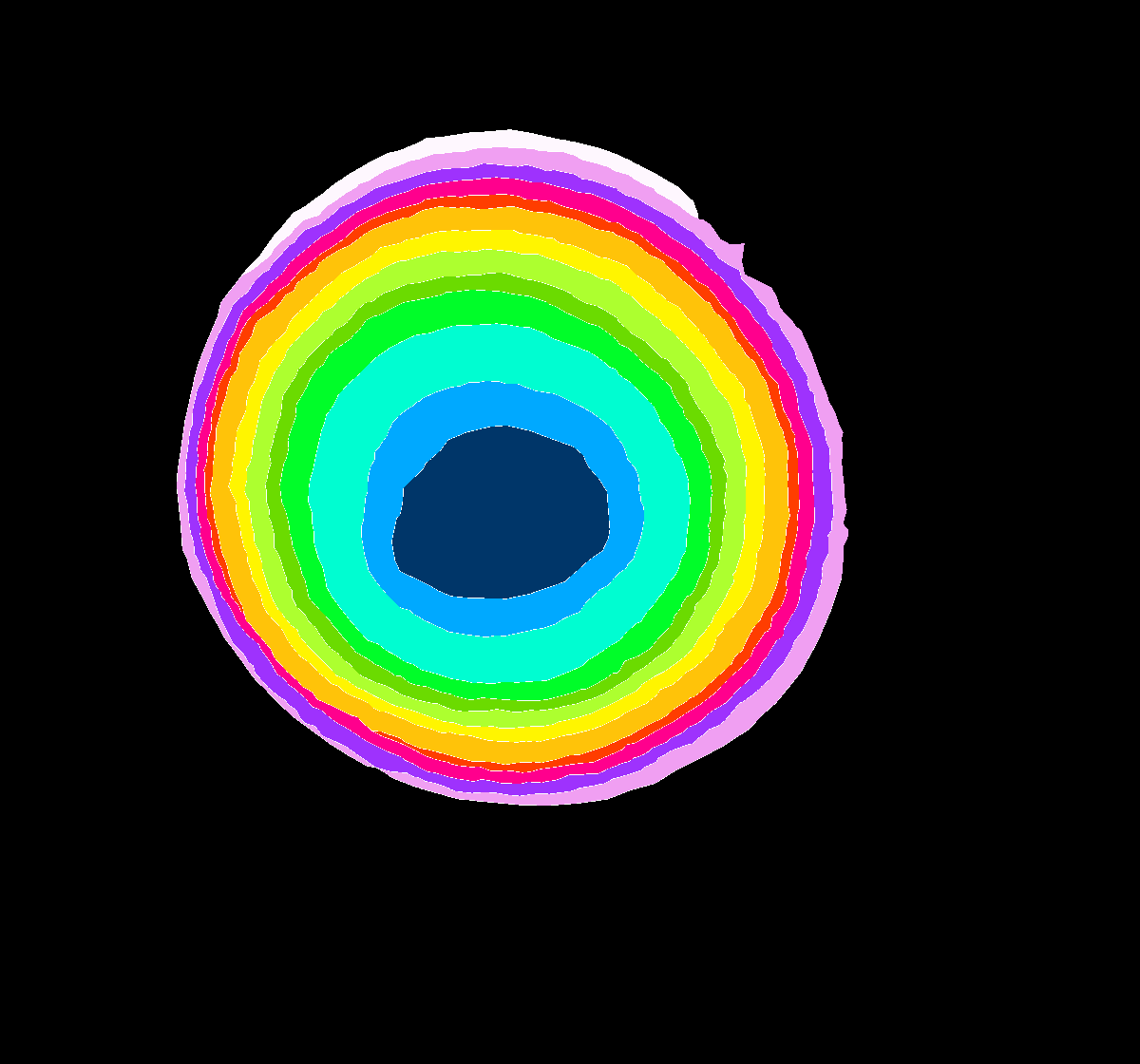}
      \caption{Annotation}
    \end{subfigure}
  \begin{subfigure}{0.24\linewidth}
      \includegraphics[width=\textwidth, height=40mm]{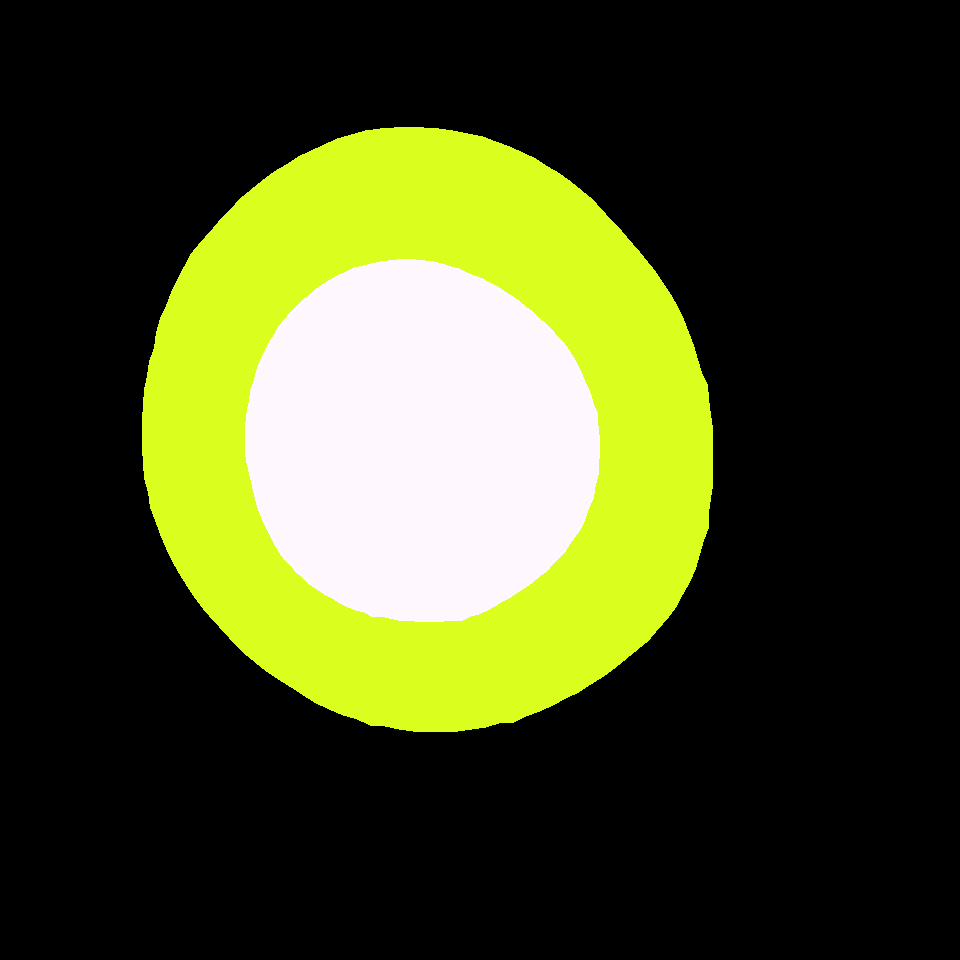}
      \caption{Deep Snake}
  \end{subfigure}
  \begin{subfigure}{0.24\linewidth}
      \includegraphics[width=\textwidth, height=40mm]{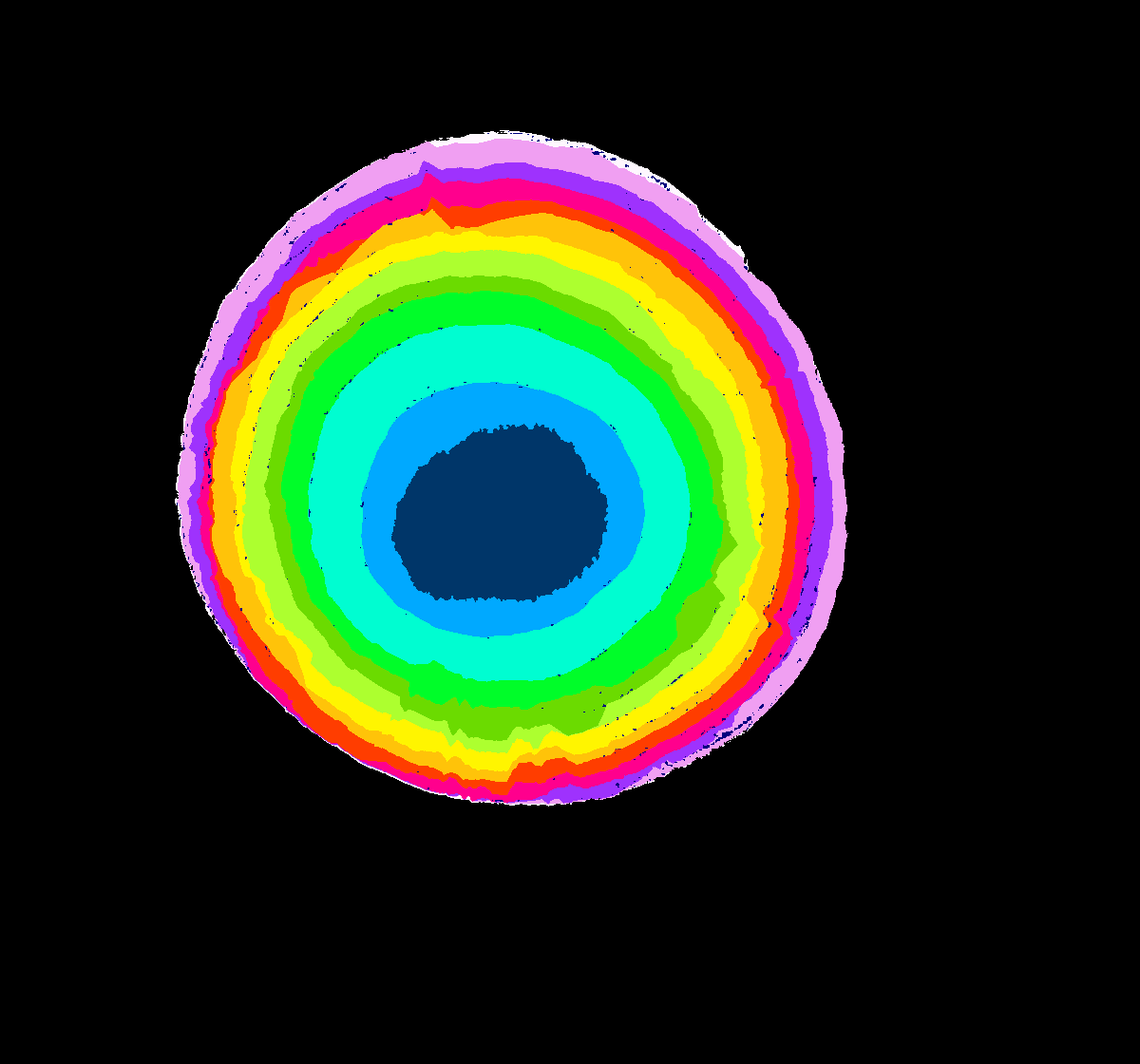}
      \caption{INBD}
  \end{subfigure}
  \caption{
    \label{fig:deepsnake}
    An image from the VM subset and comparison of INBD with Deep Snake.
    Deep Snake inherently struggles detecting concentric objects.
    The result of INBD is better, correctly estimating the number of rings but still too inaccurate for further processing.
  }
\end{figure}

\begin{figure}
    \centering
    \begin{subfigure}{0.24\linewidth}
      \includegraphics[width=\textwidth, height=40mm]{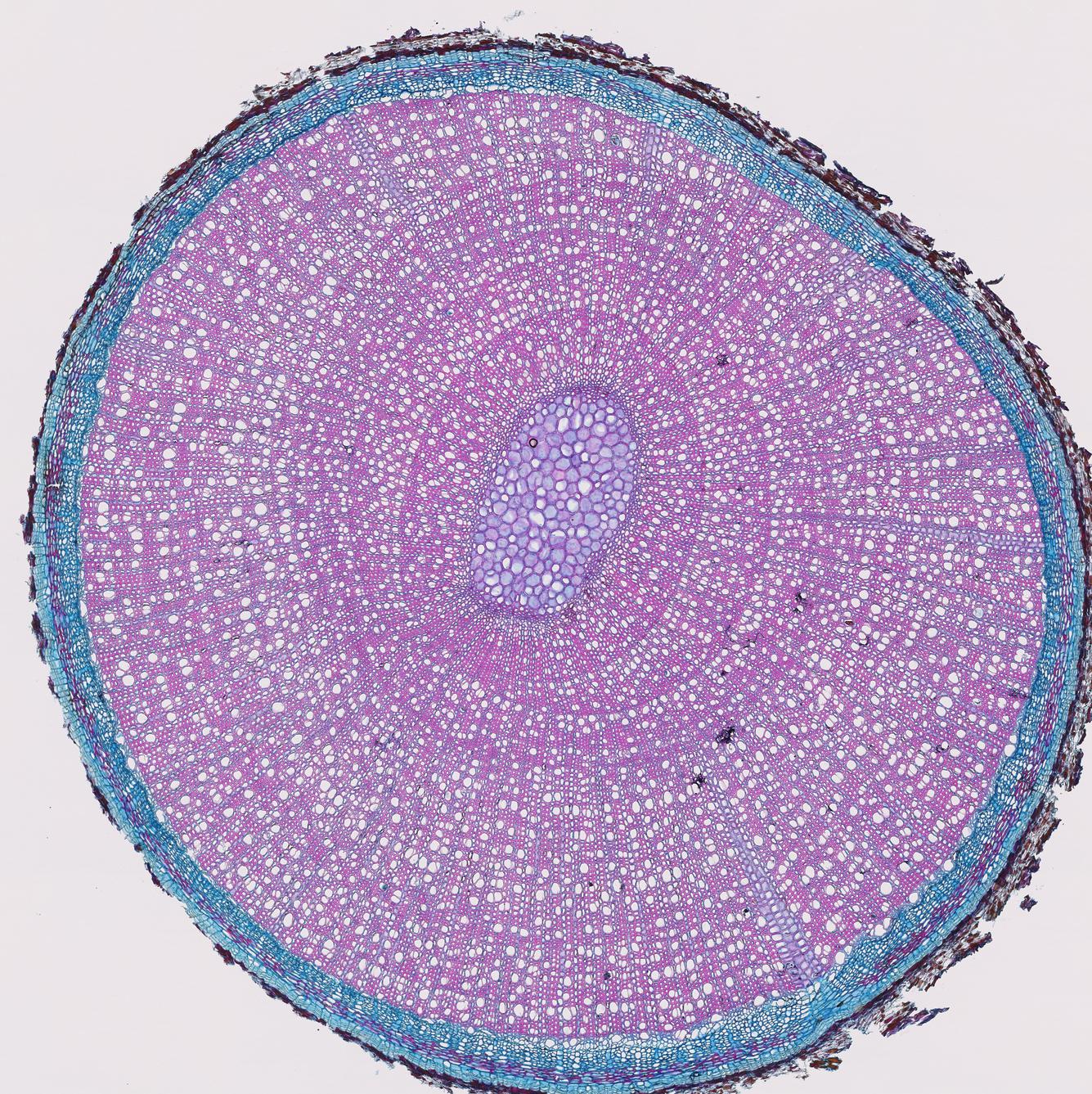}
    \end{subfigure}
    \begin{subfigure}{0.24\linewidth}
        \includegraphics[width=\textwidth, height=40mm]{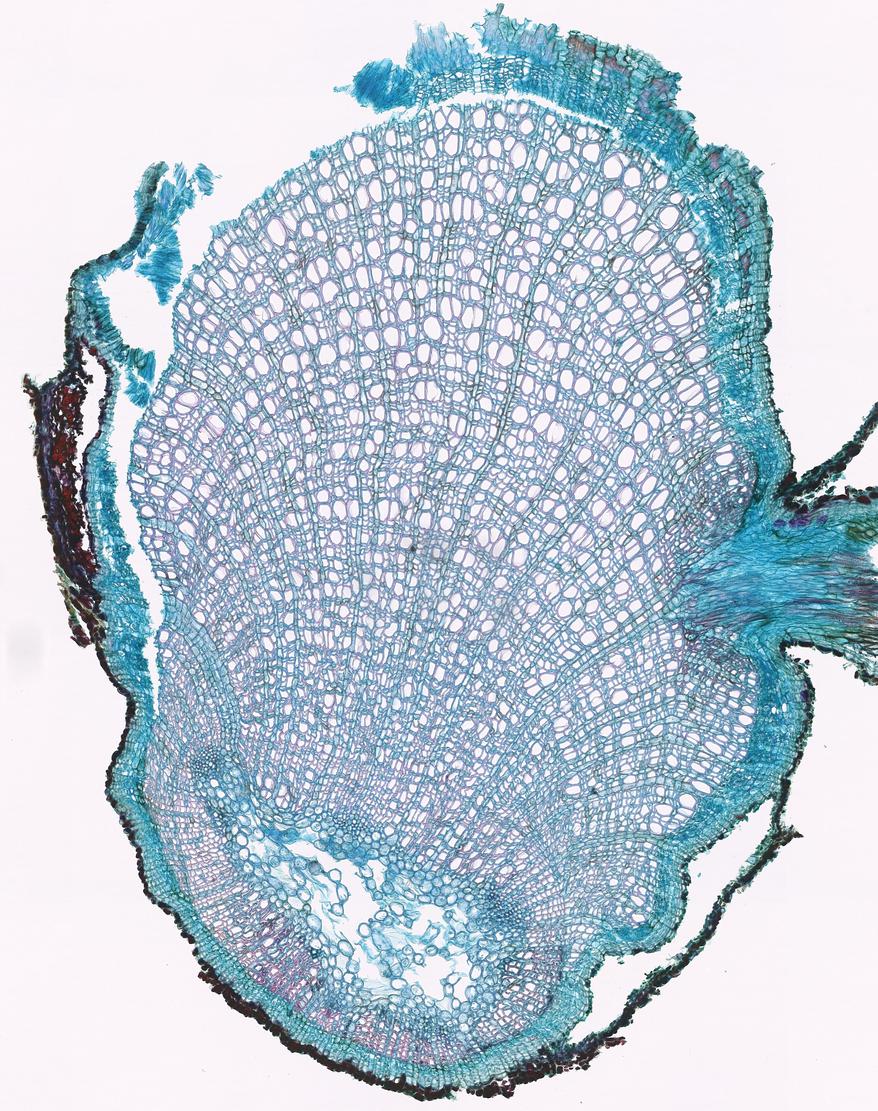}
      \end{subfigure}
    \begin{subfigure}{0.24\linewidth}
        \includegraphics[width=\textwidth, height=40mm]{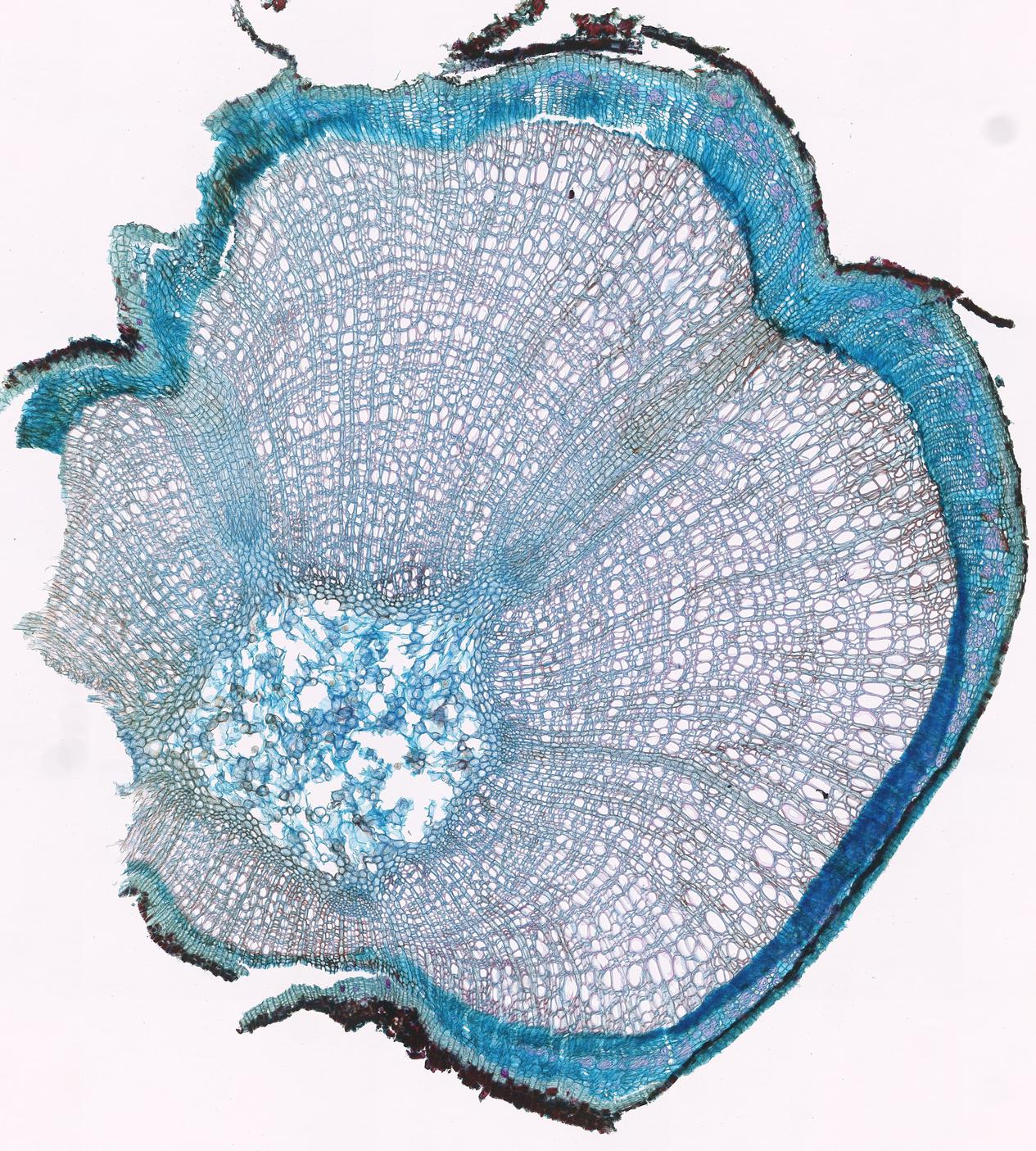}
    \end{subfigure}
    \begin{subfigure}{0.24\linewidth}
        \includegraphics[width=\textwidth, height=40mm]{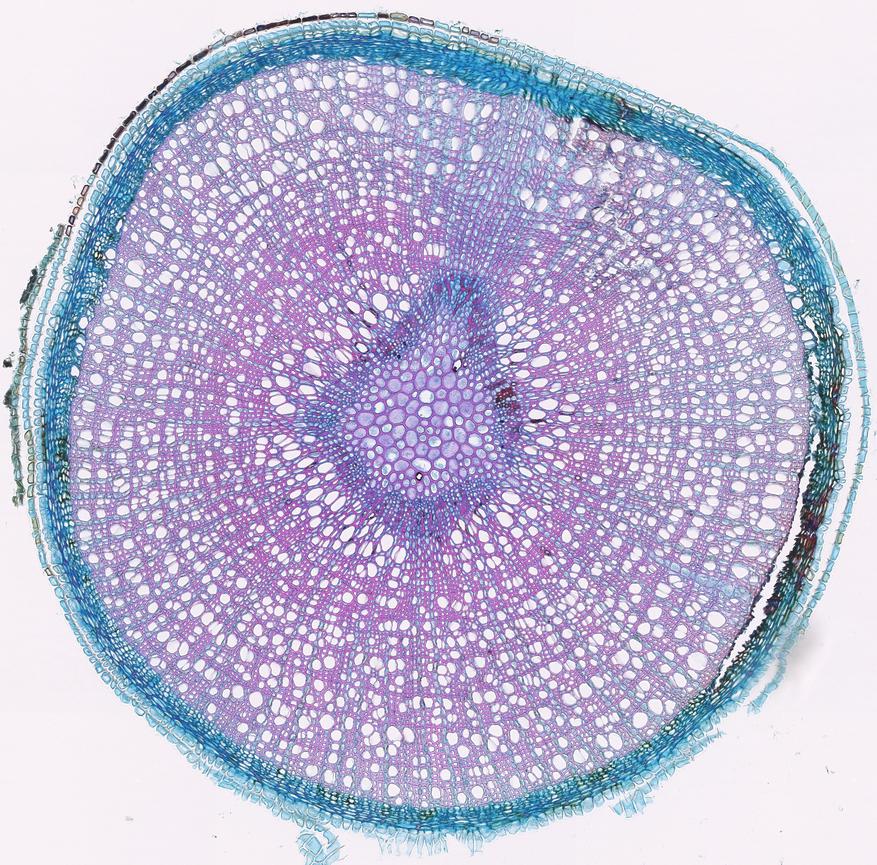}
    \end{subfigure}
    \begin{subfigure}{0.24\linewidth}
        \includegraphics[width=\textwidth, height=40mm]{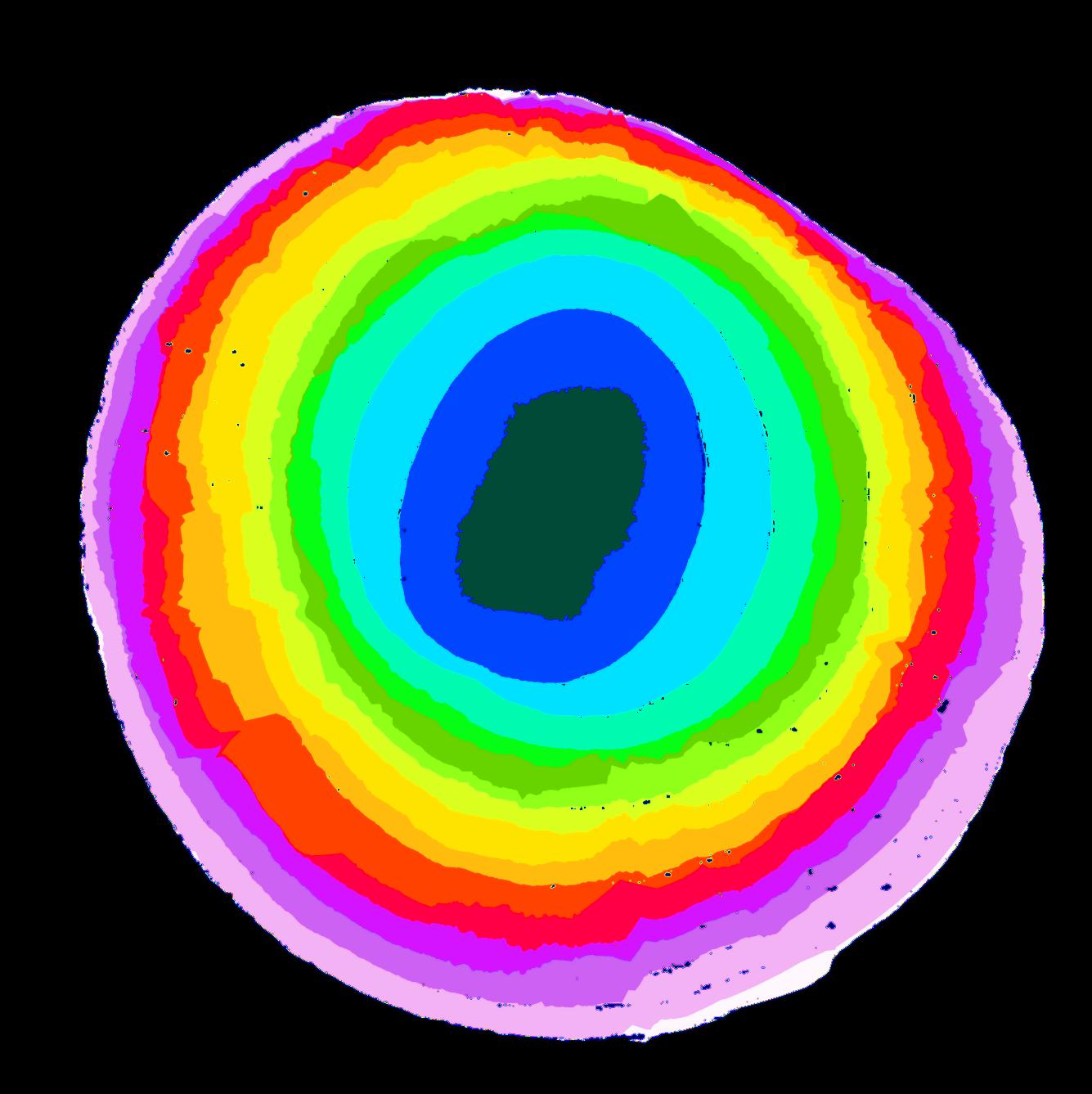}
      \end{subfigure}
      \begin{subfigure}{0.24\linewidth}
          \includegraphics[width=\textwidth, height=40mm]{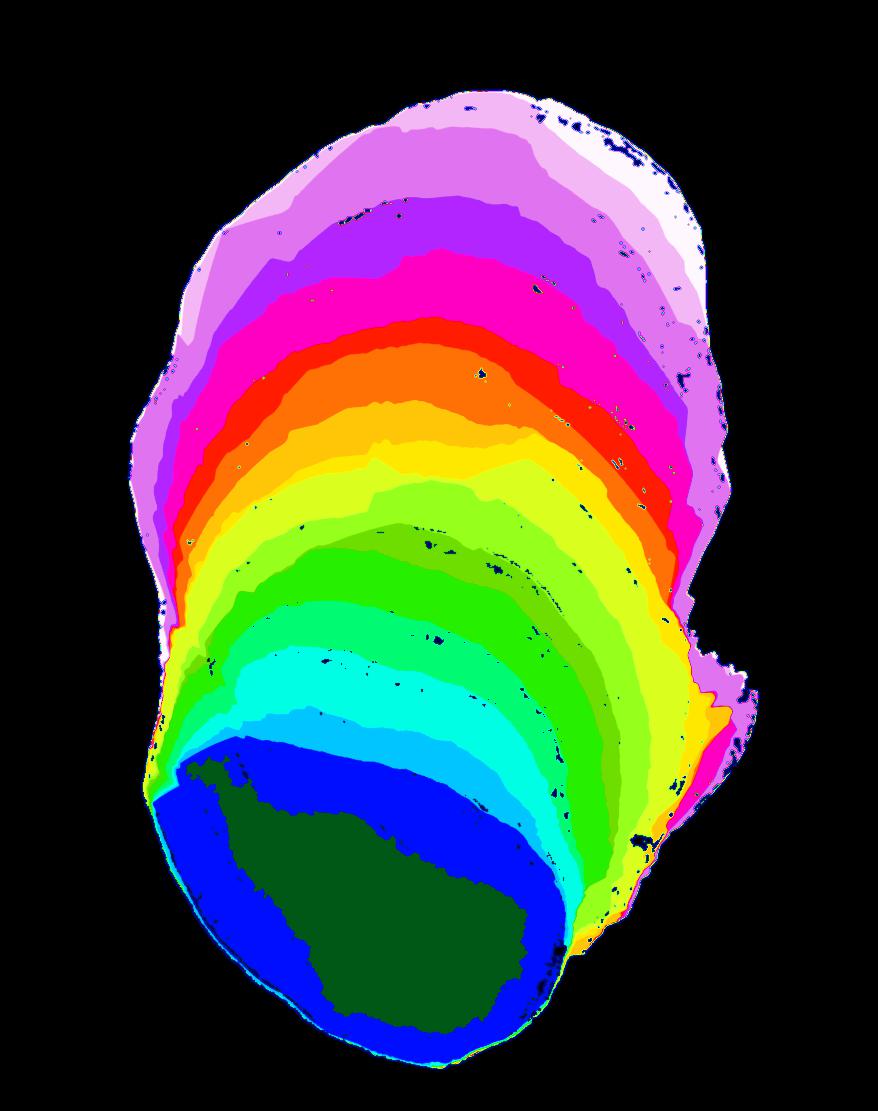}
        \end{subfigure}
      \begin{subfigure}{0.24\linewidth}
          \includegraphics[width=\textwidth, height=40mm]{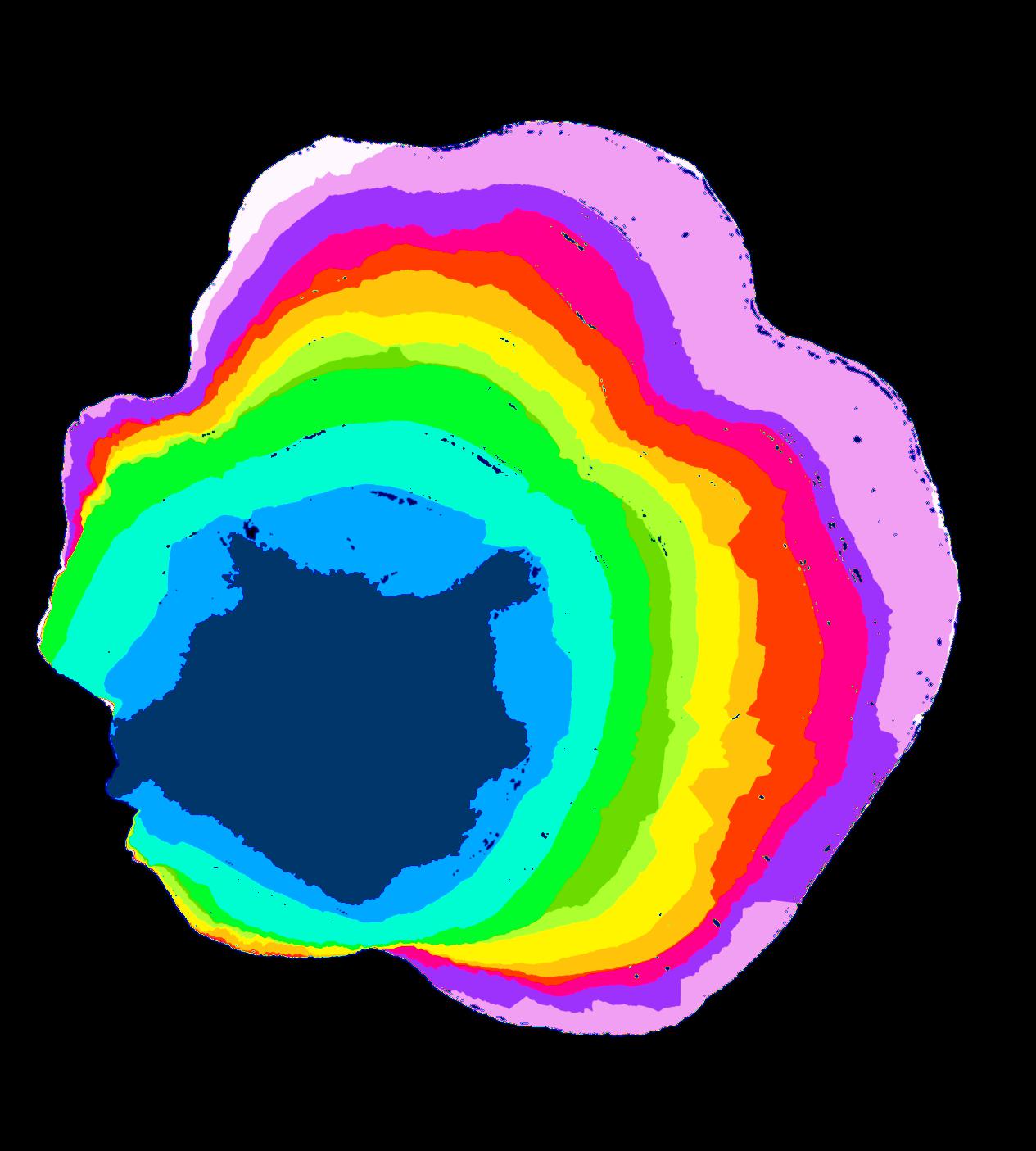}
      \end{subfigure}
      \begin{subfigure}{0.24\linewidth}
          \includegraphics[width=\textwidth, height=40mm]{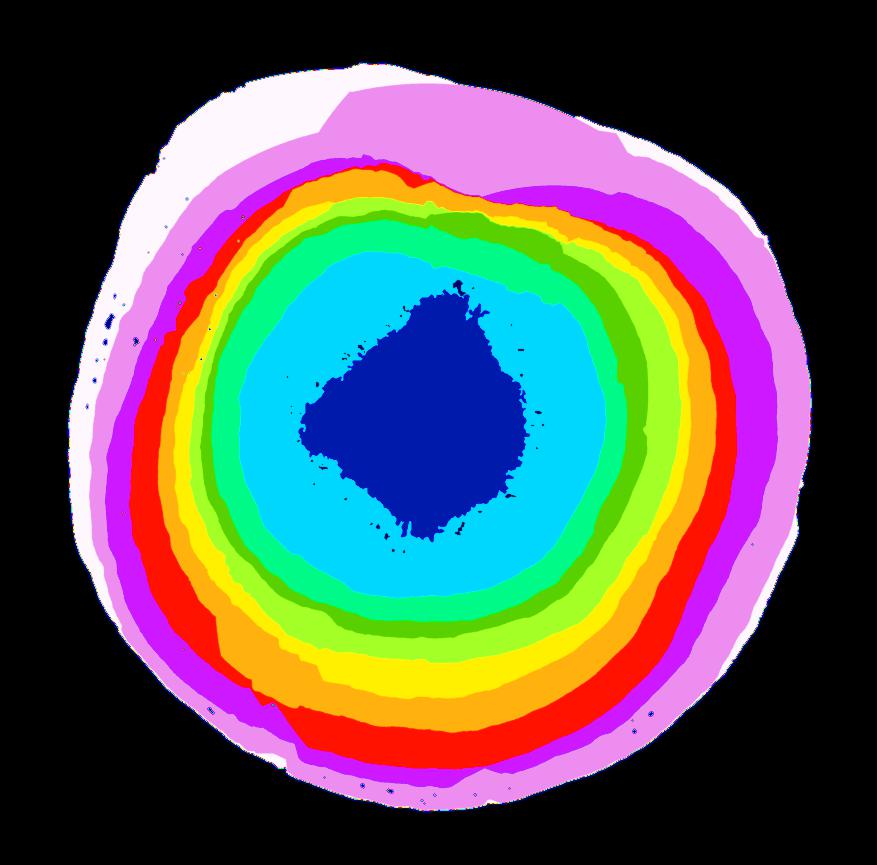}
    \end{subfigure}
    \begin{subfigure}{0.24\linewidth}
        \includegraphics[width=\textwidth, height=40mm]{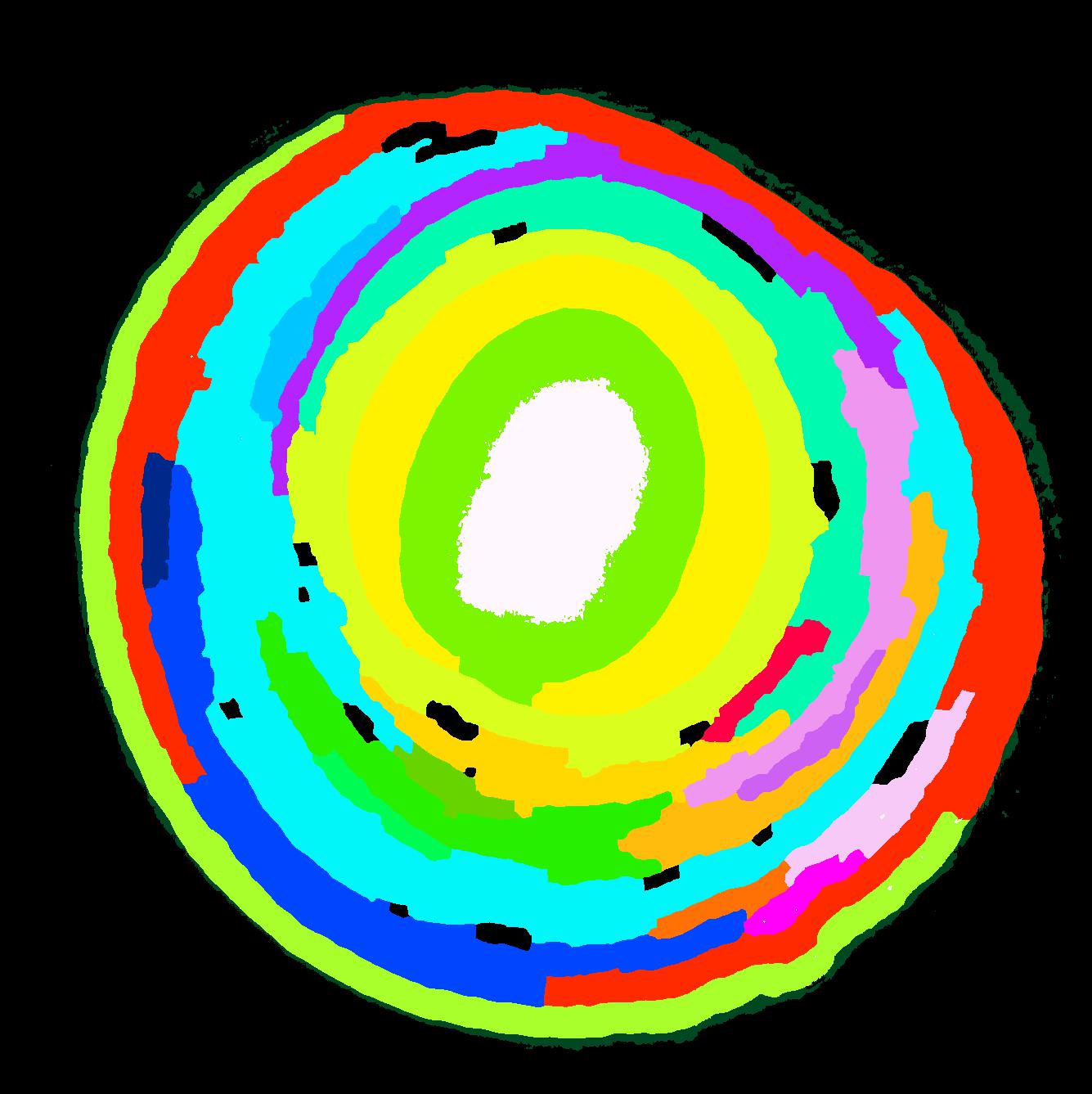}
      \end{subfigure}
      \begin{subfigure}{0.24\linewidth}
          \includegraphics[width=\textwidth, height=40mm]{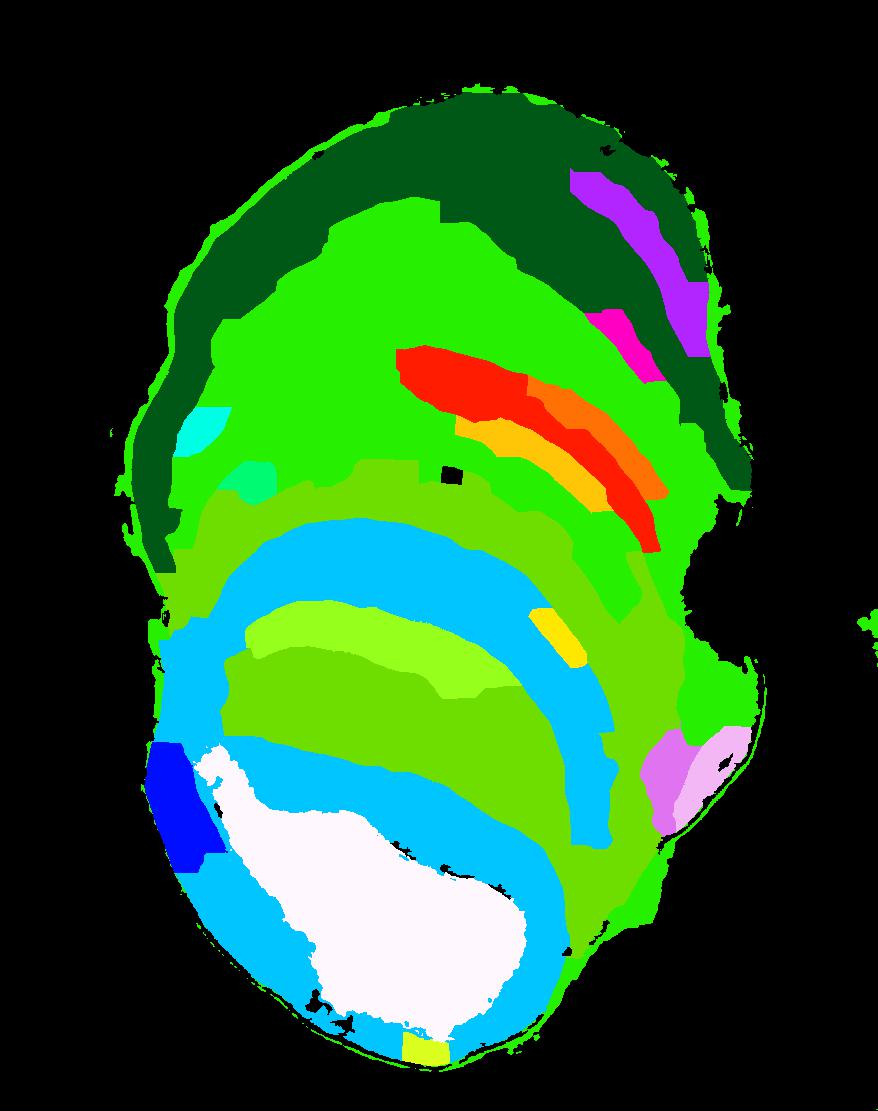}
        \end{subfigure}
      \begin{subfigure}{0.24\linewidth}
          \includegraphics[width=\textwidth, height=40mm]{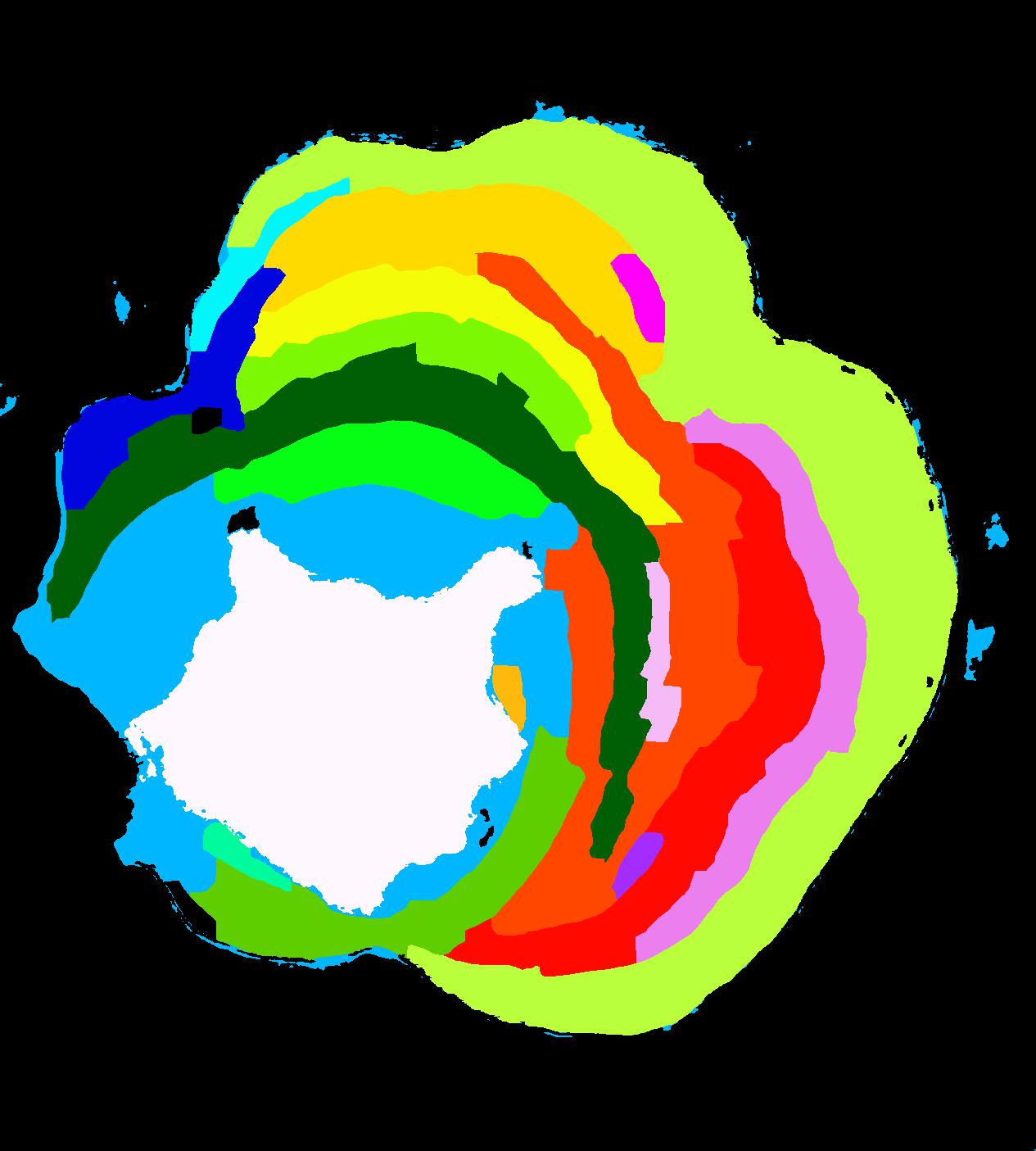}
      \end{subfigure}
      \begin{subfigure}{0.24\linewidth}
          \includegraphics[width=\textwidth, height=40mm]{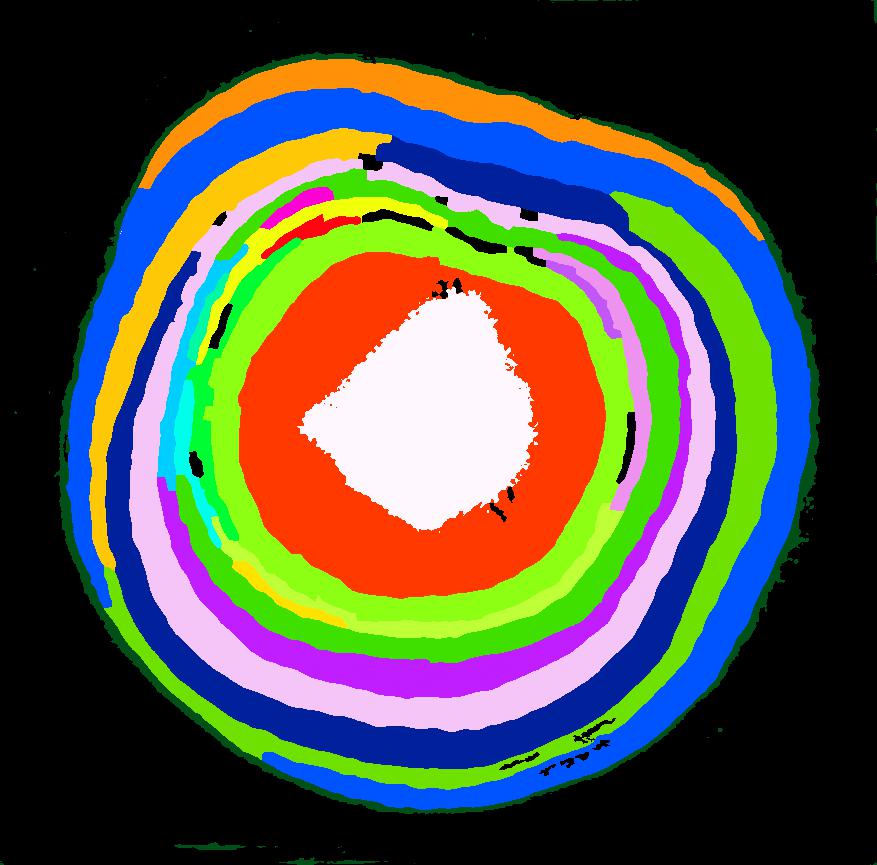}
    \end{subfigure}
    \caption{
        Example images on which none of the compared methods (center: INBD, bottom: Multicut) perform sufficiently well.
        These images are not in our published dataset because annotators were also not able to fully annotate them.
    }
    \label{fig:additional_results}
\end{figure}

\begin{figure}
    \centering
    \begin{subfigure}{0.32\linewidth}
      \includegraphics[width=\textwidth]{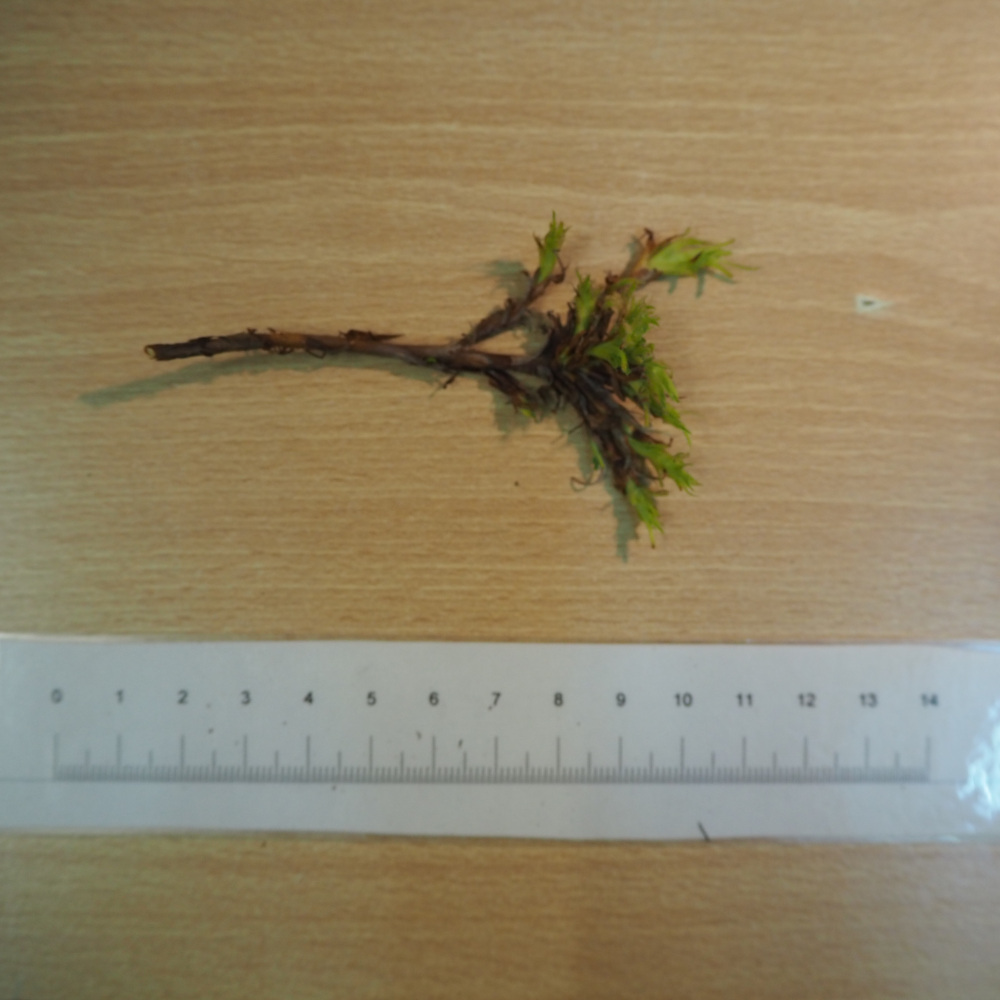}
      \label{fig:challenges-a}
    \end{subfigure}
    \hfill
    \begin{subfigure}{0.32\linewidth}
        \includegraphics[width=\textwidth]{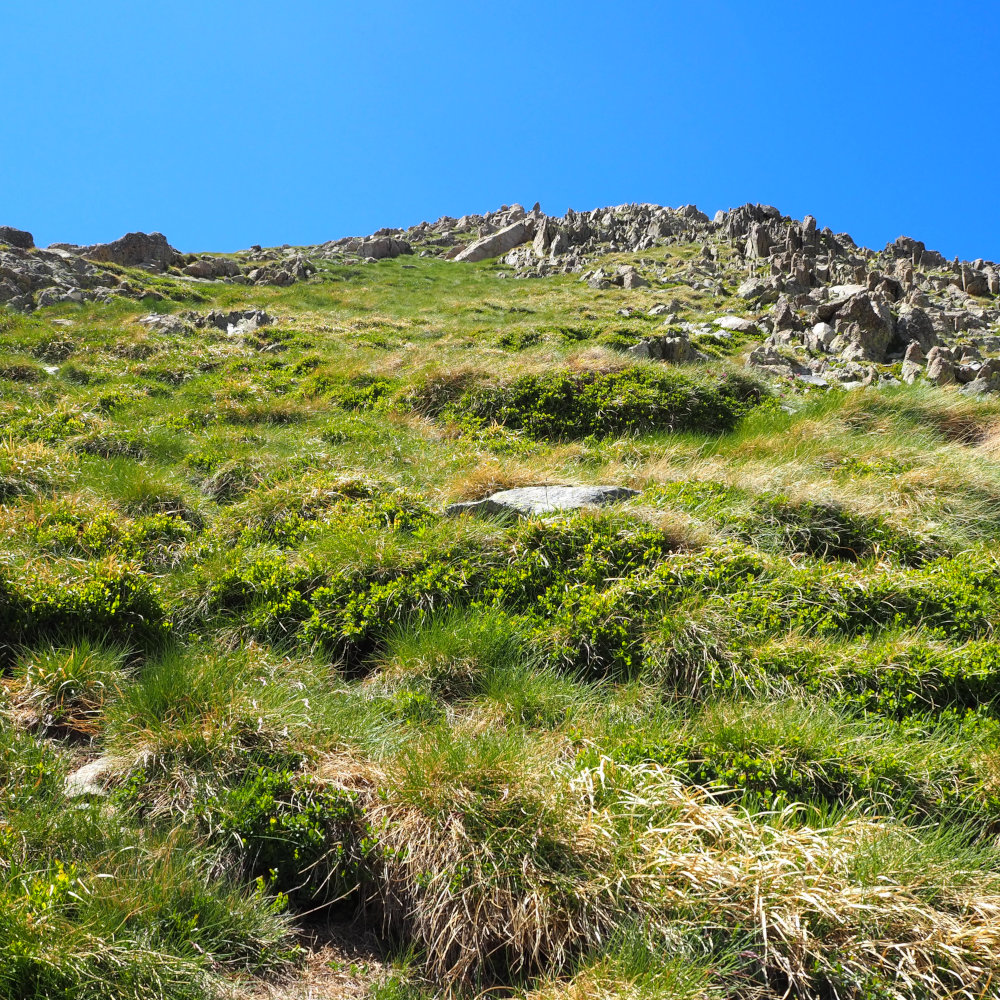}
        \label{fig:challenges-b}
      \end{subfigure}
    \hfill
    \begin{subfigure}{0.32\linewidth}
        \includegraphics[width=\textwidth]{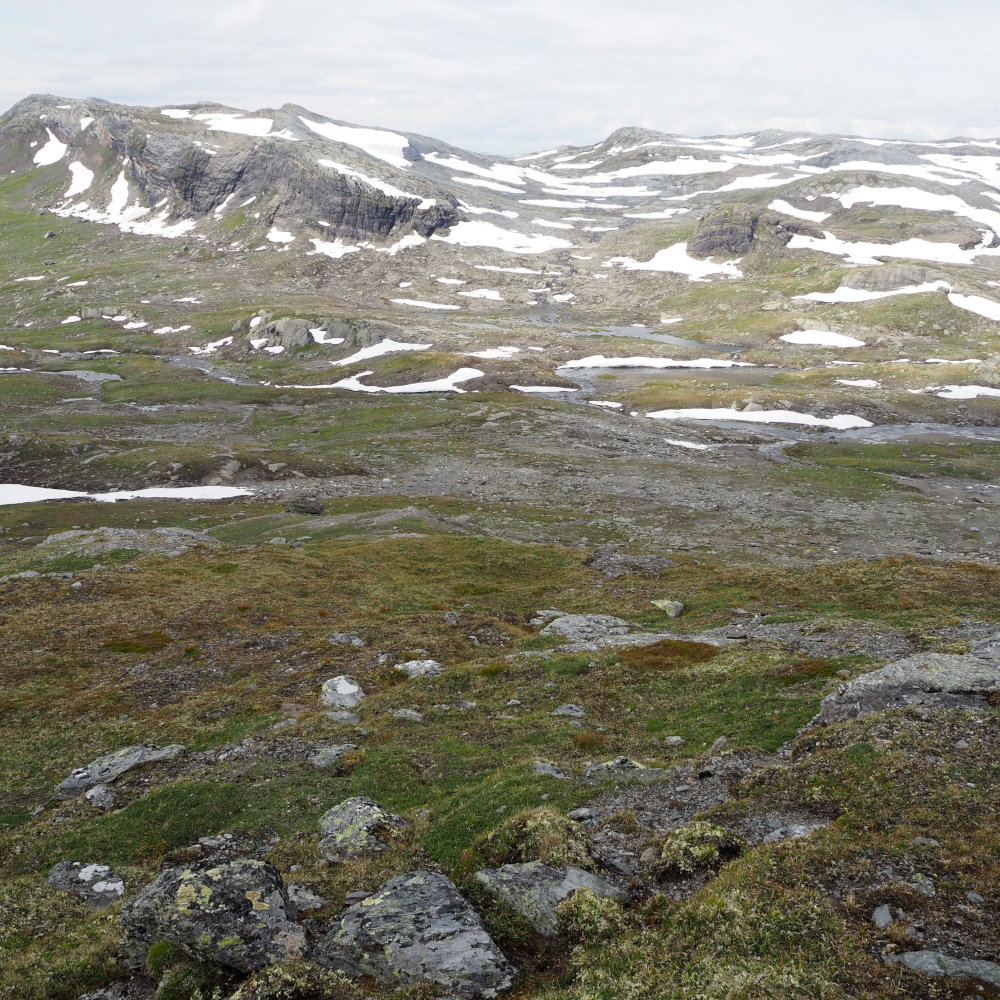}
        \label{fig:challenges-c}
    \end{subfigure}
    \caption{
        Images of branch samples (Dryas octopetala) from our dataset and the landscapes where samples were collected.
    }
    \label{fig:additional_images_bg}
\end{figure}

A comparison with Deep Snake can be seen in Figure \ref{fig:deepsnake}.
More failure cases are shown in Figure \ref{fig:additional_results} which shows the need for more research into this direction.
For better understanding of the application background, Figure \ref{fig:additional_images_bg} shows collected branch samples and the landscape where they were collected.

